\documentclass[12pt,a4paper,twoside,openright]{report}
\let\openright=\cleardoublepage

\usepackage[a-2u]{pdfx} % je potreba aby to generovalo PDFA
\usepackage[utf8]{inputenc}
\usepackage{lmodern}

\usepackage{tgpagella} % nejaky font
\usepackage[T1]{fontenc}
\usepackage{microtype}

\usepackage{amsmath}
\usepackage{graphicx,epstopdf}
\usepackage[round]{natbib}
\usepackage[nottoc]{tocbibind}

\usepackage{amsthm}

\usepackage[usenames]{xcolor}
\definecolor{dgreen}{rgb}{.0,.2,.0}

\usepackage{enumitem}
\usepackage{pdfpages}
\usepackage[ruled]{algorithm2e}
\usepackage{booktabs,tabularx}
\usepackage{pifont}

\usepackage[resetlabels]{multibib}
\newcites{A}{List of Publications}
\def\parciteA#1{\citepA{#1}} % (Smith, 2012)
\def\persciteA#1{\citetA{#1}} % Smith (2012)
\def\inparciteA#1{\citealpA{#1}} % should be Smith, 2012
\def\parcite#1{\citep{#1}} % (Smith, 2012)
\def\perscite#1{\citet{#1}} % Smith (2012)
\def\inparcite#1{\citealp{#1}} % should be Smith, 2012

\usepackage{bibentry}
\nobibliography*

\usepackage{acronym} 
\usepackage{gnuplot-lua-tikz}
\usepackage{amssymb}
\usepackage{multirow}
\usepackage{rotating}
\usepackage{pgfplots}
\pgfplotsset{compat=newest}

\usepackage{hyperref}   % Musí být za všemi ostatními balíčky
\usepackage[capitalise,noabbrev]{cleveref}

\newcommand{\motto}[2]{\begin{flushright}{\it #1}\\{\bf #2}\end{flushright}\hspace{1em}}
\def\EN{\textsc{EN}}
\def\transl#1#2{#1$\rightarrow$#2}
\def\pair#1#2{#1--#2}
\def\circled#1{\textcircled{\raisebox{-.9pt}{#1}}}
\def\significantmark{$\ddagger$}
\def\significant{\rlap{ $\ddagger$}} % to be used in table, zero width

\def\errata#1#2{#2} % for visualizing erratas

\newcommand{\TFA}{$\text{T2T}_4$}
\newcommand{\TFB}{$\text{T2T}_8$}
\newcommand{\TFC}{$\text{T2T}_{11}$}

\newcommand{\p}{\makebox[\widthof{0}][r]{ }}
\newcommand{\pp}{\makebox[\widthof{00}][r]{ }}
\newcommand{\ppp}{\makebox[\widthof{000}][r]{ }}
\def\ps#1{\makebox[\widthof{#1}][r]{ }}

\usepackage{tocloft}
\newcommand{\listapiadd}{List of Observations}
\newlistof{apiadd}{api}{\listapiadd}
\newcounter{observationcounter}[part]

\newenvironment{observationenv}[1][]{%
\refstepcounter{observationcounter}\refstepcounter{apiadd}\par\medskip%
\addcontentsline{api}{apiadd}{\protect\numberline{\theapiadd}\textit{#1}}
   \noindent \textbf{Observation~\theobservationcounter:} \rmfamily\itshape}{\medskip}
   
\Crefname{apiadd}{Observation}{Observations}

\long\def\observation#1#2{
\begin{observationenv}[#1]
#1
\label{#2}
\end{observationenv}
}

\usepackage{tikz}
\usepackage{collcell}
\newcommand*{\MinNumber}{5.0}%
\newcommand*{\MidNumberA}{13.0} %
\newcommand*{\MidNumberB}{15.14} %
\newcommand*{\MaxNumber}{16.7}%
\newcommand{\ApplyGradient}[1]{%
        \ifdim #1 mm > \MidNumberB mm
            \pgfmathsetmacro{\PercentColor}{max(min(100.0*(#1 - \MidNumberB)/(\MaxNumber-\MidNumberB),100.0),0.00)} %
            \hspace{-0.33em}\colorbox{cyan!\PercentColor!green}{#1}
        \else
            \ifdim #1 mm > \MidNumberA mm
                \pgfmathsetmacro{\PercentColor}{max(min(100.0*(#1 - \MidNumberA)/(\MidNumberB-\MidNumberA),100.0),0.00)} %
                \hspace{-0.33em}\colorbox{green!\PercentColor!yellow}{#1}
            \else
                \pgfmathsetmacro{\PercentColor}{max(min(100.0*(\MidNumberA - #1)/(\MidNumberA-\MinNumber),100.0),0.00)} %
                \hspace{-0.33em}\colorbox{red!\PercentColor!yellow}{#1}
            \fi
        \fi
}
\newcolumntype{R}{>{\collectcell\ApplyGradient}r<{\endcollectcell}}
\usepackage{array}
\newcolumntype{H}{>{\setbox0=\hbox\bgroup}c<{\egroup}@{}}

\newcolumntype{Y}{>{\raggedleft\arraybackslash}X} 
\newcolumntype{C}{>{\centering\arraybackslash}X}

\makeatletter
\def\@makechapterhead#1{
  {\parindent \z@ \raggedright \normalfont
   \Huge\bfseries \thechapter. #1
   \par\nobreak
   \vskip 20\p@
}}
\def\@makeschapterhead#1{
  {\parindent \z@ \raggedright \normalfont
   \Huge\bfseries #1
   \par\nobreak
   \vskip 20\p@
}}
\makeatother

\def\chapwithtoc#1{
\chapter*{#1}
\addcontentsline{toc}{chapter}{#1}
}

\usepackage{quotchap}

\hypersetup{pdfdisplaydoctitle, breaklinks, 
colorlinks=True,
linkcolor=dgreen, citecolor=dgreen, filecolor=dgreen, urlcolor=dgreen,
}  

\begin{document}

% Trochu volnější nastavení dělení slov, než je default.
\lefthyphenmin=2
\righthyphenmin=2

%%% Titulní strana práce
\pagenumbering{roman}

\pagestyle{empty}
\begin{center}

\centerline{\mbox{\includegraphics[width=160mm]{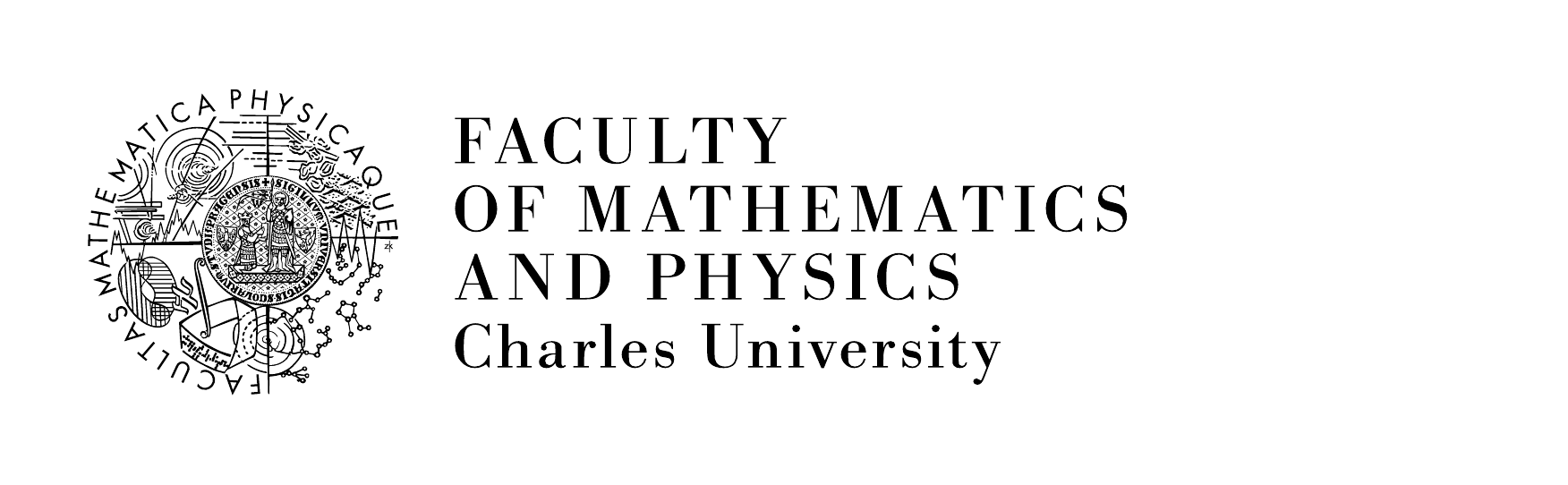}}}

\vspace{-8mm}
\vfill

{\bf\Large DOCTORAL THESIS}

\vfill

{\LARGE Tom Kocmi}

\vspace{15mm}

% Název práce přesně podle zadání
{\LARGE\bfseries Exploring Benefits of Transfer Learning \\ in Neural Machine Translation}

\vfill

% Název katedry nebo ústavu, kde byla práce oficiálně zadána
% (dle Organizační struktury MFF UK)
Institute of Formal and Applied Linguistics

\vfill

\begin{tabular}{rl}

Supervisor of the doctoral thesis: & doc. RNDr. Ond\v{r}ej Bojar, Ph.D. \\
\noalign{\vspace{2mm}}
Study programme: & Computer Science \\
\noalign{\vspace{2mm}}
Specialization: & Mathematical Linguistics \\
\end{tabular}

\vfill

% Zde doplňte rok
Prague 2019

\end{center}

\cleardoublepage
\newpage
\pagestyle{plain}

%%% Strana s čestným prohlášením k disertační práci

\vglue 0pt plus 1fill

\noindent
I declare that I carried out this doctoral thesis independently, and only with the cited
sources, literature and other professional sources.

\medskip\noindent
I understand that my work relates to the rights and obligations under the Act No.
121/2000 Coll., the Copyright Act, as amended, in particular the fact that the Charles
University has the right to conclude a license agreement on the use of this
work as a school work pursuant to Section 60 paragraph 1 of the Copyright Act.

\vspace{10mm}

\hbox{\hbox to 0.5\hsize{%
% az budu vedet den odevzdani tak bych mohl zde napsat primo datum
Prague, September 30, 2019
\hss}\hbox to 0.5\hsize{%
\hfill
Tom Kocmi
\hss}}

\vspace{20mm}

\cleardoublepage
\newpage

%%% Povinná informační strana disertační práce

\noindent
\begin{tabular}{@{}ll@{}}
\textbf{Title:} & Exploring Benefits of Transfer Learning \\ 
 & in Neural Machine Translation \\[0.3cm]
\textbf{Author:} & Tom Kocmi \\[0.3cm]
\textbf{Department:} & Institute of Formal and Applied Linguistics \\[0.3cm]
\textbf{Supervisor:} & doc. RNDr. Ond\v{r}ej Bojar, Ph.D., \\
& Institute of Formal and Applied Linguistics \\[0.3cm]
\textbf{Keywords:} & transfer learning, machine translation, \\
& deep neural networks, low-resource languages\\
\end{tabular}

\vspace{1cm}
\noindent
\textbf{Abstract:}

\noindent
Neural machine translation is known to require large numbers of parallel training sentences, which generally prevent it from excelling on low-resource language pairs. This thesis explores the use of cross-lingual transfer learning on neural networks as a way of solving the problem with the lack of resources. We propose several transfer learning approaches to reuse a model pretrained on a high-resource language pair. We pay particular attention to the simplicity of the techniques.
We study two scenarios: (a) when we reuse the high-resource model without any prior modifications to its training process and (b) when we can prepare the first-stage high-resource model for transfer learning in advance. For the former scenario, we present a proof-of-concept method by reusing a model trained by other researchers. In the latter scenario, we present a method which reaches even larger improvements in translation performance.
Apart from proposed techniques, we focus on an in-depth analysis of transfer learning techniques and try to shed some light on transfer learning improvements. We show how our techniques address specific problems of low-resource languages and are suitable even in high-resource transfer learning. We evaluate the potential drawbacks and behavior by studying transfer learning in various situations, for example, under artificially damaged training corpora, or with fixed various model parts.

% \cleardoublepage

% \noindent
% \begin{tabular}{@{}ll@{}}
% \textbf{N{\'a}zev pr{\'a}ce:} & Zkoumání výhod přenosu znalostí \\
% & v neuronovém strojovém překladu\\ [0.3cm]
% \textbf{Autor:} & Tom Kocmi \\[0.3cm]
% \textbf{{\'U}stav:} & {\'U}stav form{\'a}ln{\'i} a aplikovan{\'e} lingvistiky \\[0.3cm]
% \textbf{Vedouc{\'i} diserta\v{c}n{\'i} pr{\'a}ce:} & doc. RNDr. Ond\v{r}ej Bojar, Ph.D., \\
% & {\'U}stav form{\'a}ln{\'i} a aplikovan{\'e} lingvistiky \\[0.3cm]
% \textbf{Kl{\'i}\v{c}ov{\'a} slova:} & transfer znalost{\'i}, strojov{\'y} p\v{r}eklad, \\
% & hlubok{\'e} neuronov{\'e} s{\'i}t\v{e}, jazyky s m{\'a}lo zdroji\\
% \end{tabular}

% \vspace{1cm}
% \noindent
% \textbf{Abstrakt:}

% \noindent
% \XXlater{Český abstrakt}

\newpage

%%% Na tomto místě mohou být napsána případná poděkování (vedoucímu práce,
%%% konzultantovi, tomu, kdo zapůjčil software, literaturu apod.)

\openright

\noindent
\chapter*{Acknowledgements}

I will always remember my doctoral years as an intensive period of my life, full of both amazing and tough experiences. I could have never finished my thesis without the inspiration and support of many excellent people around me.

First of all, I would like to express my thanks to my supervisor Ondřej Bojar for his excellent guidance, exceptional encouragement, and his willingness to help. Getting to know him is one of my biggest fortunes and the most precious part of my PhD study. I cannot express my gratitude enough for his support and extraordinary care over the years.

I am very grateful to all my friends and colleagues at the Institute of Formal and Applied Linguistics, for being such friendly people with whom it is such a pleasure to work. I cannot forget all the invaluable time we spent discussing research and life, especially during our after-lunch coffee expeditions.

Special thanks go to all the inhabitants of office 423. I wouldn't have made it without your encouragement and support.

With special regards, I would like to thank Martin Popel, Jindřich Libovický, prof. Martin Holeňa, Shadi Saleh, Barbora Chattová, Miklós Danka, Tomáš Musil, and others who read and commented on earlier versions of this thesis.

Many thanks to my dear friends, and to people from StopTime for keeping me sane during my studies. Especially, I thank them for teaching me not to forget about fun in my life. Special thanks to Quido for his never-ending supplies of coffee and much more.

Finally, I would like to express my wholehearted thanks to my mother and sister for the endless patience, love, and encouragement they provided me throughout my entire life. Special thanks to my grandfather František, who taught me to be always curious, which was the essence that inspired me to become a scientist.

\vfill

{\footnotesize The work on this thesis was supported by the Charles University Grant Agency (GAUK 8502/2016, SVV~260~333, SVV~260~453), the Czech Science Foundation (18-24210S), and projects 825303 (Bergamot), H2020-ICT-2014-1-645442 (QT21) of the European Union.
This work has been using language resources and tools developed, stored, and/or distributed by the LINDAT/CLARIN project of the Ministry of Education, Youth and Sports of the Czech Republic (project LM2015071).}

\newpage

\vspace*{\fill}

\begin{center}
{\itshape\large  To my amazing mom.}
\end{center}

\vspace*{\fill}

\newpage

%%% Strana s automaticky generovaným obsahem disertační práce. U matematických
%%% prací je přípustné, aby seznam tabulek a zkratek, existují-li, byl umístěn
%%% na začátku práce, místo na jejím konci.

\openright
\tableofcontents

\chapter{Introduction}
\pagenumbering{arabic}
\setcounter{page}{1}

\motto{
The Babel fish is small, yellow and leech-like, and probably the oddest thing in the Universe.
If you stick a Babel fish in your ear, you can instantly understand anything said to you in any form of language.}{--The Hitchhiker’s Guide to the Galaxy. Douglas Adams}

With the spread of technology, people around the world are becoming more connected than ever before, and the need for seamless communication and understanding becomes crucial.
According to \perscite{simons2018ethnologue}, there are 7097 living languages in the world. However, most of the language pairs have at most hundreds to thousands of parallel sentences with a limited set of paired languages. This lack of data is a severe problem for the training of suitable \ac{MT} systems because both \ac{SMT} and \ac{NMT} are data demanding machine learning approaches.

Before my doctoral study, the primary approach for \ac{MT} was \ac{PBMT} \parcite{koehn2003statistical, bojar2015wmt}. However, a complete paradigm shift with the rise of \ac{NMT} approaches a few years ago \parcite{bojar2016wmt, bojar2017wmt}.

Currently, \ac{NMT} is a common approach to automatic translation, and according to \perscite{humanparityNMT2018, findings2018wmt} it starts to reach human parity in some language pairs. However, it is only valid for high-resource language pairs where we have plenty of available data -- usually dozens of millions of parallel sentences. The performance with low amounts of data can be dramatically reduced up to the point where \ac{PBMT} systems outperform \ac{NMT} \parcite{koehn2017six}. To some extent, the problem can be mitigated if the \ac{NMT} model is scaled down accordingly \parcite{sennrich2019revisiting}, but it does not resolve the data demands.

The goal of this thesis is to study transfer learning techniques to improve the performance of \ac{NMT}, especially for translating low-resource language pairs. In general, transfer learning refers to the use of vaguely related training data to improve the performance at the desired task. For instance, in \ac{NMT}, transfer learning reuses the model, or its parts, trained for one language pair to improve the performance in a different language pair.

\section{Contributions}

The main contributions of this work are the proposed transfer learning techniques with a broad analysis. We show several approaches on how to improve the performance of (not only) low-resource language pairs with a model trained for a different high-resource language pair. During our analysis, we show that:

\begin{itemize}
\item Transfer learning works for both low and high-resource language pairs and achieves better performance than training from random initialization.
\item Transfer learning in \ac{NMT} does not have negative effects known in other fields and can be used as an initialization method for \ac{NMT} experiments.
\item We show that the quantity of parallel corpus plays a more important role in transfer learning than relatedness of language pairs.
\item We observe that transfer learning works as a better initialization technique and improves performance even when no language is shared between both models.
\end{itemize}

Apart from the main contributions, we also describe several other research ideas, starting with our contribution to the development of \pair{Czech}{English} parallel corpora \parciteA{bojar2016czeng}, experiments with pretrained word embeddings \parciteA{kocmi2017icon}, word embeddings with subword information \parciteA{kocmi2016subgram}, a neural language identification tool \parciteA{kocmi2017lanidenn}. We also contributed to the implementation of a research sequence-to-sequence framework Neural Monkey \parciteA{neuralmonkey}. 

During my doctoral study, we investigated the use of curriculum learning \parciteA{kocmi2017curriculum}, helped to prepare a Neural Training Shared Task at WMT 2017 \parciteA{cuni_training_wmt_2017}, and developed a neural abstractive summarization tool \parciteA{straka2018sumeczech}. Furthermore, we participated in several shared tasks \parciteA{iwslt_bojar2016ufal, cuni_translation_wmt_2017, wat2017, kocmi2018wat, kocmi2018wmt, kocmi2018iwslt, kocmi2019wmt}.

The complete list of publications that I co-authored during my doctoral study can be found in the \emph{List of Publications} on page \pageref{list:publications}. All publications went through the peer-review process. The only exception is \persciteA{kocmi2019ecofriendly} that is not yet published.

\section{Structure of the Thesis}

In the following chapters, we describe our approach to low-resource transfer learning in \ac{NMT}. This thesis consists of two main parts: the description and evaluation of our transfer learning approaches and a broad analysis of the approach. The thesis includes results and text snippets from our published works. Short textual parts are reused only from works without other co-authors except for my supervisor. Furthermore, the works used in this thesis are stated at the beginning of individual sections.

\begin{itemize}
  \item In \cref{chap:background}, we introduce the definition of low-resource languages and describe language resource categories for \ac{MT}. We describe \pair{Czech}{English} corpus \parciteA{bojar2016czeng}, show the problem with noisy parallel sentences, and present our tool for neural language identification \parciteA{kocmi2017lanidenn}. Lastly, we describe the standard approaches to the evaluation of machine translation.
  \item In \cref{chap:nmt}, we describe \ac{NMT} architectures with a focus on word embeddings and segmentation of words. We present our approach to word embeddings that includes subword information in word representation \parciteA{kocmi2016subgram}.
  \item In \cref{chap:transfer_learning}, we present one of the central parts of this thesis: transfer learning for \ac{NMT}. We describe two scenarios, which differ in assumed conditions on the transferred model. We propose several approaches for both scenarios and evaluate them. This chapter is based mostly on our two papers: \persciteA{kocmi2018trivial} and \persciteA{kocmi2019wmt}.
  \item Equally important is \cref{chap:analysis}, where we analyze the gains by transfer learning and shed some light on the understanding of the underlying mechanisms.
  \item We conclude the thesis in \cref{chap:conclusion}.
\end{itemize}

\chapter{Background}
\label{chap:background}
% \addcontentsline{toc}{chapter}{Background}

In this chapter, we discuss the definition of low-resource languages, sources, and quality of parallel corpora in \cref{sec:language_resource}. In \cref{sec:lanidenn}, we describe our language identification tool that is going to be used later in transfer learning analysis. Then we describe the training data used throughout our work in \cref{sec:training_data}. Lastly, we explain how the \ac{MT} is evaluated in \cref{sec:machine_translation_evaluation}.

\section{Language Resources}
\label{sec:language_resource}

Understanding and collecting the available resources is a crucial step towards training \ac{NMT} systems. In general, we can classify resources based on their domain, their size, and their quality. Another criterion is whether they are monolingual or parallel.

The availability of parallel sentences is crucial for \ac{NMT}. It is expensive and hard to obtain a large number of parallel sentences. On the other hand, it is easier to obtain a monolingual data by crawling the Internet or from various online sources. The amount of available monolingual data in the target language typically far exceeds the number of parallel sentences. Thus researchers have been trying to utilize monolingual data for \ac{MT} and we investigate it in \cref{sec:backtranslation}.

A comparably important criterion to the size of the parallel corpus is its domain and quality. It is well-known that domain-specific training data are better for the final performance than some general data. The same holds for the quality where a large, noisy training set can notably hinder the performance of an \ac{NMT} system (e.g. survey by \inparcite{chu2018survey}).

Throughout the thesis, we use standard shortcuts ``k'' for thousand and ``M'' for million.
In the case of a parallel corpus, we often use ``sentences'' as a shortcut for ``sentence pairs''. 

Whenever we talk about a language pair, we use a dash between the languages, e.g. \pair{Czech}{English}. On the other hand, whenever we talk about actual translation direction, we use an arrow to specify a direction from which language to which we translate, e.g. \transl{English}{Czech}.

In this section, we introduce a definition of an \ac{MT} domain, followed by an examination of the quality of parallel corpus and approaches for dataset cleaning.

\subsection{Definition of Domain in Machine Translation}
\label{sec:domain_definition}

The definition of a domain varies among papers. In general, it is considered any set of instances from a dataset containing a common feature. In most of the papers concerning domain adaptation, the authors define the domain as the source of the dataset. This domain is closely related to the topic or genre of the documents \parcite{hildebrand2005adaptation, chu2017empirical, servan2016domain}. Examples of such domains are subtitles, literature, news, medical reports, patents, IT, and many more. All of them vary in the used vocabulary, style of writing, and content. 
%We demonstrate this domain with the following example:

% \begin{description}[noitemsep]
% \item[Source EN:] The trial ended in March.
% \item[News CS:] Soudn\'{i} proces skon\v{c}il v b\v{r}eznu. (The \emph{court case} ended in March)
% \item[Scientific CS:] Studie byla ukon\v{c}ena v b\v{r}eznu. (The \emph{study} ended in March)
% \end{description}

Another feature is the formality or informality of the document, which is closely related to honorifics in languages like Czech, German, or Japanese \parcite{sennrich2016controlling}. 
It is a way of encoding the relative social status of speakers to the readers, and for many styles, like official documents, it is an important feature determining the quality of the translation.

Further, we can distinguish documents based on sentiment. The sentiment tone of a text can change with machine translation \parcite{glorot2011domain, mohammad2016translation} mainly because of language differences and ambiguity. Another issue with sentiment is that the same information can be written from a positive, neutral, or negative stance. 
We can go even further and distinguish documents based on the writing style of the author or expected style preference of the reader, as of formality of a speech, specialized vocabulary, or dialects \parcite{jeblee2014domain}.

\subsection{Definition of Low-Resource Languages}
\label{sec:lowresource_definition}

Recently, rapid development of \ac{NMT} systems led to the claims that the human parity has been reached on high-resource language pairs like \pair{Chinese}{English} \parcite{humanparityNMT2018} or \pair{Czech}{English} \parcite{findings2018wmt}. However, \ac{NMT} systems tend to be very data-hungry. \perscite{koehn2017six} have shown that in the low-resource scenarios, \ac{NMT} lag behind \ac{PBMT} approaches. This problem led to the rise of attention in low-resource \ac{NMT} in recent years.

A precise definition of language pairs that count as low-resource is a research question itself. One must consider all aspects of available language resources as well as the language itself.

One of the aspects is the domain of the parallel corpus. Having a large amount of domain-specific parallel sentences can be considered high-resource in the given domain, but low-resource in the general domain, where the performance can be terrible.
For example, one common source of parallel sentences for low-resource languages is the Bible, which is translated into hundreds of languages \parcite{bibleCorpora}. However, it is a highly domain and style specific text.

Highly-inflected languages further complicate the definition of low-resource by presenting a notable sparsity problem with the various forms of inflected words and therefore requiring more parallel sentences to reach comparable performance as the translation of less inflected languages \parcite{denkowski2017stronger}.

\perscite{gu-etal-2018-universal} define the \textit{extremely low-resource scenario} by the minimal amount of data needed for \ac{NMT} to obtain a reasonable translation quality. They showed that extremely low-resource scenario could be considered up to the 13-28k parallel sentences for \transl{English}{Romanian} translation.

In recent years, researchers have organized several machine translation shared tasks in the low-resource scenario. \perscite{niehues2018iwslt} introduced a task on \transl{Basque}{English} low-resource translation with an available in-domain corpus of 6k sentence pairs and 940k out-of-domain sentence pairs. The low-resource translation tasks in \ac{WMT} 2018 \parcite{findings2018wmt} have been \pair{Estonian}{English} with 880k and \pair{Turkish}{English} with 208k parallel sentences. This year in \ac{WMT} \parcite{findings2019wmt}, the low-resource language was \pair{Gujarati}{English} with 170k parallel sentences and \pair{Kazakh}{English} with 220k parallel sentences.

Notably, \perscite{koehn2017six} found out that \ac{NMT} outperforms \ac{SMT} when more than 24.1M words are available, i.e., approximately 1M \pair{Spanish}{English} parallel sentences. However, we need to add that \perscite{sennrich2019revisiting} recently revisited the training condition for low-resource and showed that current systems outperform \ac{SMT} even in the low-resource scenario if careful training is performed.
 
Thus a usual definition is that language pairs with less than a million training pairs are deemed low-resource.

\begin{figure}
\caption{Learning curves for various language pairs with various sizes of parallel corpus. We can see that language pairs with less than 1M data quickly flatten out or even start overfitting as in the case of \transl{Basque}{English}.}
\begin{center}
\input{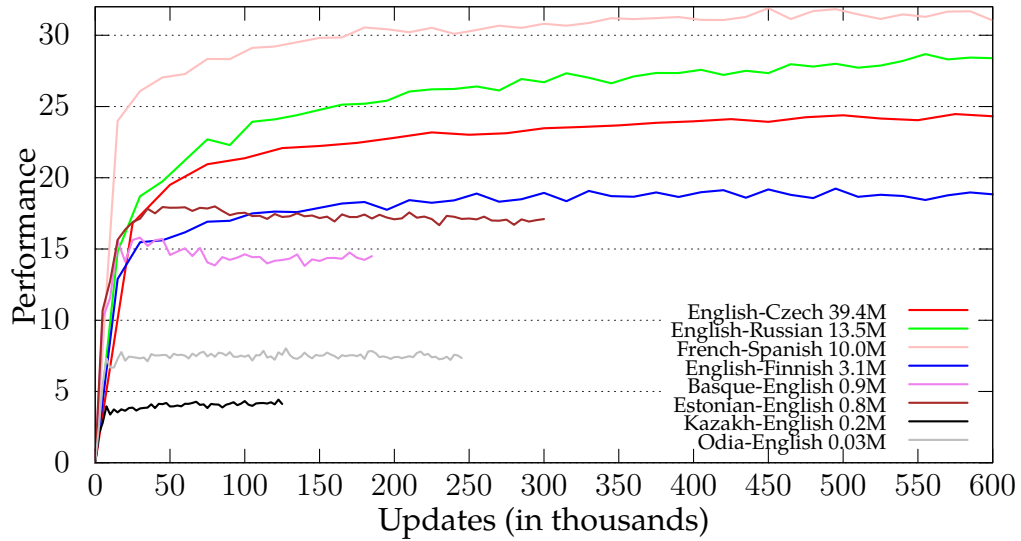}
\end{center}
\label{fig:low_resource_overfitting_non}
\end{figure}

Another point of view, which can be used to delineate low-resources is behavior during training. \cref{fig:low_resource_overfitting_non} shows so-called learning curves, i.e., the performance of a given system on a development set (also called held-out set) throughout the training. When the curves bend down (e.g. the performance starts decreasing), it is an indication that the model is overfitted, usually by memorizing training sentences.

During our experiments, we noticed that low-resource languages often overfit or flatten out within first 50-100k of weight updates, which in our setting is roughly half a day of training on a single GPU. In contrast, the higher-resource language pairs rarely overfit, usually flatten out after several hundred thousand steps, see \cref{fig:low_resource_overfitting_non} for an illustration.

% With this in mind, we propose an alternative definition of low-resource language pair based on its behavior during the training of \ac{NMT}:

% \textit{A language pair is considered low-resource for a given domain if the neural machine translation can overfit or flatten out with the available data within the first 50k updates.}

% This definition is still not exact because the batch size can vary even across various architectures and systems. However, based on our experiments, this definition correlates with the size of the parallel corpus, when \ac{NMT} quickly overfit or flatten out when trained on less than 1M parallel sentences.

We should also note that a language pair considered low-resource today might not be considered low-resource in the future: either due to the newly available data but also due to the improvements in \ac{NMT} training techniques.

\subsection{Resource Quality}

Language resources for machine translation come in varying degrees of quality. On the one extreme, there are proper translations made by professional translators, which result in parallel corpora with more aligned translations and adherence to sentence-to-sentence alignment. On the other extreme sentences automatically extracted from noisy crawled web pages, which are often miss-aligned, contain non-word parts, and even sentences in a wrong language.

Another issue with parallel corpora is the phenomenon called translationese \parcite{gellerstam1986translationese}. A text translated from one language to the second one has different linguistic properties than text written in the second language originally. According to \perscite{baker1993corpus}, translated texts are often:

\begin{itemize}
  \item Simplified -- when translators subconsciously simplify the message.
  \item Normalized -- to conform the typical features of target languages up until a point of a slight change in meaning.
  \item Explicitated -- the notion that structures of text are explained in more detail due to the rarity of the phenomenon in the target language, for example, explaining abbreviations. It is, to some extent, an inverse to the simplification.
\end{itemize}

\perscite{stymne2017effect} evaluated the effect of translationese on the \ac{MT} systems and found out that the translation direction indeed influences the final quality. However, as the authors mention, their study would need to be evaluated over a larger sample of language pairs and \ac{MT} systems to be more reliable.

\subsection{Corpus Cleaning}

Collecting training data is the first step needed in order to start training machine translation models. For low-resource languages, crawled data are often one of the largest sources of parallel sentences. Unfortunately, the resources based on crawled data are usually noisy. In order to use the crawled data, we need to clean them first.  

There is a whole field for filtering parallel corpora. \perscite{koehn2018filtering} prepared a shared task intending to study various techniques of filtering parallel corpus for \ac{NMT} to improve its performance by reducing noisy data. Filtering usually consists of pre-filtering rules, like removing non-word tokens, various tags or sentences with mismatched lengths. More advanced techniques rely on various scoring functions. 

Furthermore, we can remove short segments of up to a few tokens. These sentences are often relics of misaligned pairs. A standard is to remove sentences with less than five tokens \parcite{koehn2018filtering}. 

In contrast, we can also think about removing long segments or breaking them to shorter ones. For practical reasons of batch training on GPU, sentences within one batch are padded to a fixed length according to the longest sentence in the batch. Thus in order to increase the batch size, we can remove very long sentences from the corpus, because they result into large padding of other sentences in the same batch. However, removing long sentences should be done only in the case if there are very few of them compared to the size of the whole corpus so that their removal will not affect the overall number of parallel sentences. In our papers, we followed the recommendation of \perscite{popel2018training} and set the threshold to 100 or 150 tokens as we have not witnessed any change in the performance, but it allows us to increase the batch size, which is beneficial for the training \parcite{popel2018training}.

Lastly, whenever we are dealing with a corpus that was automatically collected, we may want to check all sentences by automatic language identification tool in order to remove sentences that are in a different language. Although this step can remove correct sentences due to errors done by the automatic language identification, it is usually beneficial since it removes a part of noisy sentences.

We have also contributed to the field of language identification by developing a neural language identification tool called \ac{LanideNN}. We describe the tool in the following section.

\section{LanideNN: Language Identification Tool}
\label{sec:lanidenn}

In language identification, we want to determine the language of some input text automatically. Monolingual language identification assumes that the given document is written in a single language. In multilingual language identification, the document is usually in two or three languages, and we only want their names. In the task of code-switching identification, where the speaker of one language uses words from a different second language, we need to tag individual words or phrases by intended language.

Techniques of language identification can rely on handcrafted rules, usually of high precision but low coverage, or data-driven methods that learn to identify languages based on sample texts of sufficient quantity \parcite{cavnar1994ngram,yang2010ngram,carter2013microblog}.

In \persciteA{kocmi2017lanidenn}, we have developed a neural language identification tool. It focuses on multilingual identification as well as language identification from short segments. This section includes results and textual parts of our paper.
We use this tool for analysis of transfer learning behavior in \cref{sec:traces_of_parent_language_pair}.

%moved
\begin{figure}[t]
\caption{Illustration of our LanideNN model architecture. The input on top is processed through the network to get assigned language label at bottom based on argument maximum.}
\begin{center}
\def\svgwidth{\columnwidth}
\includegraphics[width=\textwidth]{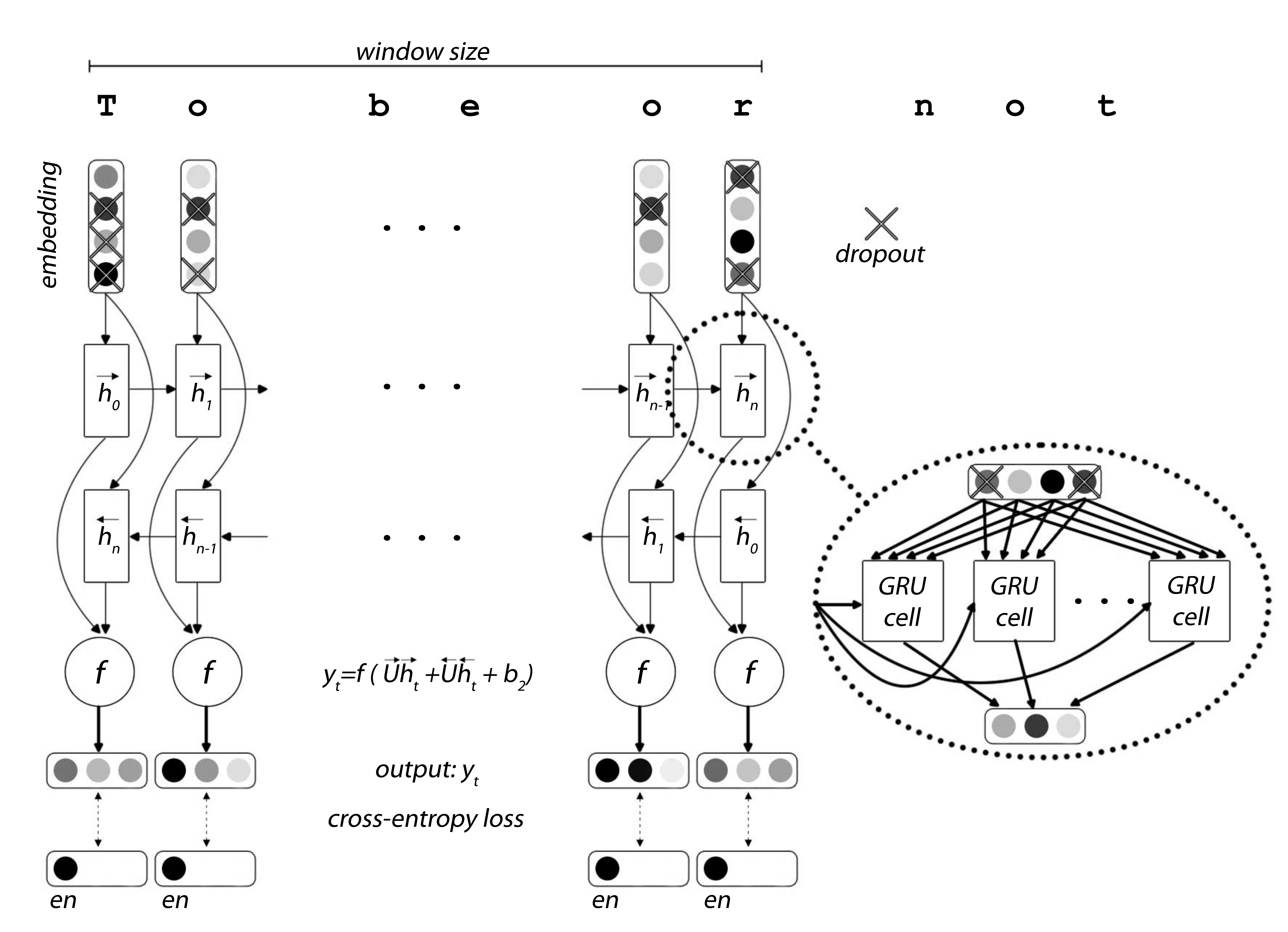}
\end{center}
\label{fig:lanidenn_model}
\end{figure}

To our knowledge, and also according to the survey by \perscite{garg2014survey}, we have been the first\footnote{One exception in using \ac{NN} for language identification task before our work is \perscite{dubaee2010language}, who combine a feed-forward network classifier with wavelet transforms of feature vectors to identify English and Arabic from the Unicode representation of words, sentences or whole documents.
The benefit of \ac{NN} in this setting is not very clear to us because the writing scripts of studied languages can alone distinguish English from Arabic.} who approached the language identification with neural networks, and further, we reached the state-of-the-art in multilingual language identification in the year 2017 \parciteA{kocmi2017lanidenn}.

The method we propose is designed for short texts without relying on sentence or document boundaries. If documents are known and if they can be assumed to be monolingual, this additional knowledge should not be neglected. However, for the long term, we aim at streamlined processing of noisy data genuinely appearing in multilingual environments. For instance, our method could support the study of code-switching in e-mails or other forms of conversation, or to analyze various online media such as Twitter, for example, see \perscite{montes2007blogging}.

The model is trained on a character sliding window of input texts. It takes individual source characters as input and provides a language label for each of them. Whenever we need to recognize the language of a document, we take the language assigned by our model to the majority of characters. Our model operates on a window of 200 characters of input text, i.e. individual characters, encoded in Unicode. The model classifies each character separately but quickly learns to classify neighboring characters with the same label.

The architecture consists of a character level embedding layer that is connected to \acf{BiRNN}. Each unit of \ac{BiRNN} is the Gated Recurrent Unit \parcite{cho2014learning}. The model outputs a probability distribution over all language tags. In order to determine the language tag of a character, we take the index of tag in the output layer with the maximum value. The architecture is illustrated in \cref{fig:lanidenn_model}.

To prevent overfitting, we use dropout 0.5 \parcite{dropout2014} during model training on the character embeddings. The key idea is to drop connections randomly. It prevents neurons from co-adapting too much, i.e. starting to depend on outputs of other neurons too much, which is a typical symptom of overfitting to training data.

We have collected, and processed training data containing 131 languages, which is more than other tools can recognize. We also added support to identify non-linguistic structures such as HTML codes, which are often present in noisy crawled data. 
For the final training set, we mixed all sources for a given language at the line level and removed the end of lines. Thus we artificially created a large code-switching training dataset.

\subsection{Monolingual Language Identification}

Most of the related research is focused on monolingual language identification, i.e. recognizing the single language of an input document. We compare our approach in this setting with several other algorithms on the dataset presented by \perscite{baldwin2010language}. The dataset consists of 3 different testsets, each containing a different number of languages, styles, and document lengths collected from different sources.

\begin{table}[t]
\caption{
Results of monolingual language identification. Entries marked with * are accuracies reported by \perscite{lui2012langidpy}, the rest are our measurements.}
\begin{center}
\begin{tabularx}{\textwidth}{@{}lcCCC@{}}
\toprule
System  & Supported languages & EuroGov & TCL & Wikipedia \\
\midrule
* LangDetect  & \p{}53 & -- & .818 & .867\\
* TextCat     & \p{}75 & .941 & .605 & .706\\
* CLD         & \p{}64 & .983 & .732 & .831\\
Langid.py     & \p{}97 & \textbf{.987} & .931 & \textbf{.913}\\
CLD2          & \p{}83 & .979 & .837 & .854 \\
Our model & \textbf{132} &  .977 & \textbf{.954} &  .893\\
\bottomrule
\end{tabularx}
\end{center}
\label{tab:lanidenn_monolingual}
\end{table}

\cref{tab:lanidenn_monolingual} summarizes the accuracy of several most popular algorithms on the three testsets \parcite{baldwin2010language}. For some algorithms, we report values as presented by \perscite{lui2012langidpy} without re-running. 

Despite the considerably higher number of languages covered, our model performs reasonably close to the competitors on EuroGov (testset of 10 languages) and Wikipedia (testset of 67 languages). We reached the best score on the TCL (testset of 60 languages).

We compare our method with two top language recognizers, Langid.py \perscite{lui2012langidpy} and CLD2.\footnote{\url{https://github.com/CLD2Owners/cld2}} We train our model on more languages, and we do not restrict it to only the languages included in the testset. We did not restrict LanideNN to recognize only a subset of languages. Thus we may be losing by recognizing detailed dialect labels. Furthermore, our approach evaluates the examples on a short span of 200 characters at a time. The final prediction is based on the average predictions across the document. A different strategy of breaking the input could improve our results.

\subsection{Short-Text Language Identification}

In order to demonstrate the ability of our method to identify the language of short texts such as tweets, search queries or user messages, we wanted to use an existing corpus, such as the one released by Twitter.\footnote{\url{http://blog.twitter.com/2015/evaluating-language-identification-performance}}
Unfortunately, the corpus contains only references to the actual tweets, and most of them are no longer available. We thus have to rely on our testset, as described in \persciteA{kocmi2017lanidenn}, where the average line length of example is 142.3 characters.

\begin{table}[t]
\caption{Results on our testset for short texts. The first column shows an accuracy over all 131 languages. The second column shows an accuracy over languages that all systems have in common.}
\begin{center}
\begin{tabularx}{\textwidth}{@{}l@{\hskip 3cm}CC@{}}
\toprule
System & All languages & Common languages \\
\midrule
Langid.py & .567 & .912 \\
CLD2 & .545 & .891 \\
Our model & \textbf{.950} & \textbf{.955}\\
\bottomrule
\end{tabularx}
\end{center}
\label{tab:short-text-results}
\end{table}

Results on short texts are reported in \cref{tab:short-text-results}. The two other systems, Langid.py and CLD2, are trained on texts unrelated to our collection of data and cover fewer languages. It is therefore not surprising that they perform much worse when averaged over all languages.

For a fairer comparison, we also report accuracy on a restricted version of the testset that included only languages supported by all three tested tools. Both our competitors are meant to be generally applicable, so they should (and do) perform quite well. Our system nevertheless outperforms them, reaching the accuracy of 95.5. Arguably, we can be benefiting from having trained on different texts and different distribution but the same sources as this testset.

\subsection{Multilingual Language Identification}
\label{sec:lanidenn_multilingual_experiments}

In multilingual language identification, systems are expected to report the set of languages used in each input document. The evaluation criterion is thus macro- (M) or micro- ($\mu$) averaged precision (P), recall (R) or F-measure (F).\footnote{Note that for comparability with results reported in other works, macro-averaged F-score is calculated as average over individual F-scores instead of the harmonic mean of $P_M$ and $R_M$.} The main criterion of the ALTW2010 shared task \parcite{baldwin2010language} was to maximize the micro-averaged F-score ($\mathbf{F_\mu}$).

To interpret the character-level predictions by our model for multilingual identification, we used the ALTW2010 development data to set the threshold empirically: if a language is predicted for more than 3\,\% of characters in the document,  we consider the language as one of the document's languages.

We evaluate our model on two existing testsets for multilingual identification, ALTW2010 shared task, and WikipediaMulti. Both testsets come with training data. Thus we retrain our model to test its in-domain performance.

ALTW2010 shared task \parcite{baldwin2010language} provided 10000 bilingual documents divided as follows: 8000 training, 1000 development, and 1000 test documents. The task is to recognize which two languages are present in the document.

WikipediaMulti \parcite{lui2014automatic} is a dataset of artificially prepared multilingual documents, mixed from monolingual Wikipedia articles from 44 languages. Each of the artificial documents contains texts in $1\leq k \leq 5$ randomly selected languages. The average document length is 5500 bytes. The training set consists of 5000 monolingual documents, the development set consists of 5000 multilingual documents, and testset consists of 1000 documents for each value of $k$. 

\begin{table}[t]
\caption{Results of multilingual language identification. All models uses the same training set, either ALTW2010 or WikipediaMulti. The * identifies results as reported by \perscite{lui2014automatic}.
}
\begin{center}
\begin{tabularx}{\textwidth}{@{}l@{\hskip 1cm}CC@{\hskip 1cm}CC@{}}
\toprule
 & \multicolumn{2}{c@{\hskip 1cm}}{ALTW2010} & \multicolumn{2}{c}{WikipediaMulti}\\
\cmidrule(r{1cm}){2-3} \cmidrule(l){4-5}
System &  $F_M$ & $\mathbf{F_\mu}$ &  $F_M$ & $\mathbf{F_\mu}$ \\
\midrule
* \perscite{baldwin2010language} &  .464 &  .829 & -& -\\
* ALTW2010 winner &  .699 &  .932 & -& -\\
* SEGLANG  & \textbf{.784} & .905 & .875 & .861\\
* LINGUINI  & .513 & .700 & .802 & .805\\
* \perscite{lui2014automatic} &  .748 & .933  & .961& .959\\
\midrule
\perscite{lui2014automatic} &  .724 & .931  & .961& .963\\
Our model  & .779 & \textbf{.965} & \textbf{.966} & \textbf{.964}\\
\bottomrule
\end{tabularx}
\end{center}
\label{tab:lanidenn_multilang}
\end{table}

The results are in \cref{tab:lanidenn_multilang}. For algorithms SEGLANG and LINGUINI, we only reproduce the results reported by \perscite{lui2014automatic}. We use the system by \perscite{lui2014automatic} as a proxy for the comparison: we retrain their system and obtain results similar to those reported by the original authors. The differences are probably due to the Gibbs sampling used in their approach. 

We see that our model outperforms all other models in the task's main criterion $\mathbf{F_\mu}$. More details and results are in our paper \persciteA{kocmi2017lanidenn}.

\subsection{Code Switching Analysis}

\begin{figure}[t]
\caption{Illustration of text partitioning. The black triangles indicate true boundaries of languages. The black part shows probability of detecting the language labeled in gray color, and the gray part shows complement for the second language since in this setup we restricted our model to use only the two languages in question.
The misclassification of Italian and German as English in the last two examples may reflect increased noise in our English training data.}
\begin{center}
\includegraphics[width=\textwidth]{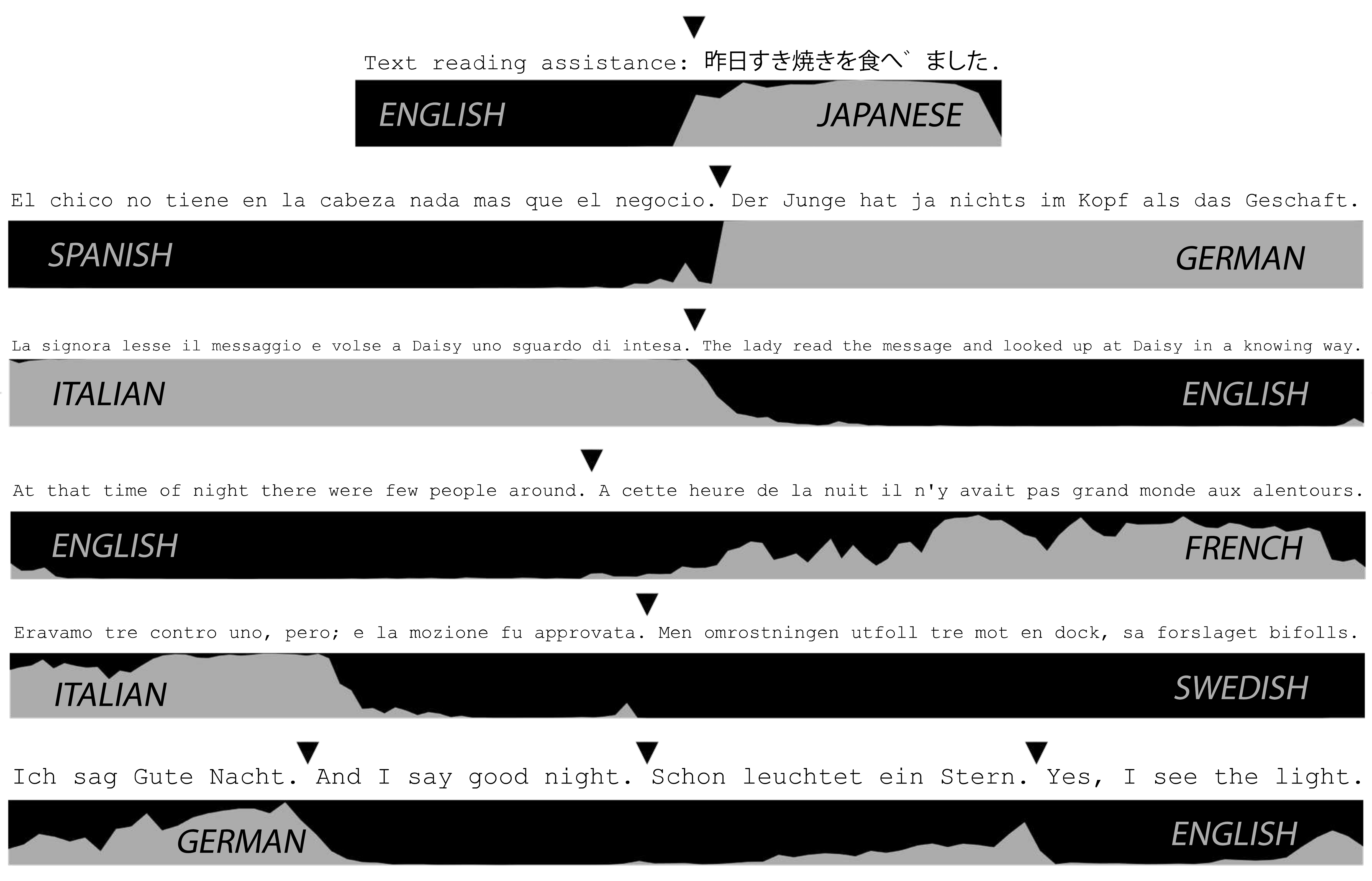}
\end{center}
\label{fig:codeswitching}
\end{figure}

Lastly, we illustrate the ability of our model to recognize borders between languages, even on short sentences. \cref{fig:codeswitching} presents the behavior of our model on text with mixed languages. The graph represents the probability of the model to recognize the first language in one color and (1--probability) for the second languages in the other color.

We have selected very short (50--130 characters) and challenging segments where the languages mostly share the same script. Finding the boundary between languages written in different scripts is not difficult, as illustrated by the first example. Moreover, it can recognize borders even on other examples, however failing on others. The more robust examination would be needed to evaluate this performance. Thus we leave it just as an illustration.

\subsection{Conclusion}

We trained language identification tool with a focus on identifying even short segments such as sentences as they are usually the smallest units gathered for \ac{MT}. Furthermore, we focused on the multilingual setting.
We were one of the first to tackle the problem with neural networks. We reached the best result on one of the three testset in monolingual language recognition. We outperformed other tools in the task of short segment identification as well as multilingual language identification.

We collected a dataset and trained our model to recognize considerably more languages than other state-of-the-art tools. We have developed a language identification algorithm based on bidirectional recurrent neural networks and made it publicly available.\footnote{\url{https://github.com/tomkocmi/LanideNN}}

We use this tool for analysis of transfer learning behavior in \cref{sec:traces_of_parent_language_pair}.

\section{Training Data}
\label{sec:training_data}

This section summarizes all the datasets we use for training, development, and test evaluation throughout the whole thesis. We first introduce our high-resource \pair{Czech}{English} parallel corpora in \cref{sec:czeng}, a core dataset used in most parts of our work. Then we describe the remaining datasets used in this thesis.

\subsection{\pair{Czech}{English} Parallel Corpus}
\label{sec:czeng}

For our work, we have contributed to the development of a high-resource parallel corpus of \pair{Czech}{English}. The size of the parallel corpus is essential for training a neural machine translation model, where more data from various domains generally leads to a better general translation. Therefore, we have extended the original corpus \parcite{bojar2012czeng10} by collecting over four times more data from more domains.
The corpus contains 62.5 million sentences from various sources: Subtitles, EU Legislation, Fiction, Parallel Web Pages, Technical Documentation, Medical, PDFs from Web, News, Project Navajo, and Tweets.

CzEng 1.6 dataset is important for our work for several reasons:

\begin{itemize}
  \item Czech language has rich morphology, which makes machine translation hard \parcite{baerman2015oxford}.
  \item It allows \pair{Czech}{English} to serve as a high-resource language pair with 62.5 million data.
  \item It contains data from various domains.
  \item We understand fluently both languages, which is useful for the manual error analysis.
\end{itemize}

After the release, we have noticed that CzEng 1.6 contains a considerable number of sentence pairs with English sentences in the Czech side of the corpus or vice versa. We identified the wrong sentence pairs by automatic language identification. Initially, we wanted to use our LannideNN (see \cref{sec:lanidenn}); however, the implementation is considerably slower than other available tools. Thus we decided to use Langid.py \parcite{lui2012langidpy} instead to automatically check the languages.

%moved
\begin{table}[t]
\caption{Datasets sizes overview. Word counts are from the original corpora, tokenizing only at whitespace and preserving the case.}
\begin{center}
\begin{tabularx}{\textwidth}{@{}lCCC@{}} 
\toprule
 &   & \multicolumn{2}{c}{Words} \\
\cmidrule(l){3-4} 
Language pair  & Sentence pairs  &  First language & Second language \\
\midrule
\pair{Odia}{\EN{}}        & \pp{}27k & \p{}604k & \p{}706k \\    
\pair{Gujarati}{\EN{}}    & \p{}173k & \p{}1.4M & \p{}1.4M\\    
\pair{Estonian}{\EN{}}    & \p{}0.8M & \pp{}14M & \p{}20M \\
\pair{Basque}{\EN{}}      & \p{}0.9M & \ppp{}5M & \pp{}7M  \\
\pair{Finnish}{\EN{}}     & \p{}2.8M & \pp{}44M  & \p{}64M \\
\pair{German}{\EN{}}      & \p{}3.5M & \pp{}73M & \p{}77M\\
\pair{Slovak}{\EN{}}      & \p{}4.3M & \pp{}82M & \p{}95M  \\
\pair{Russian}{\EN{}}     & 12.6M & \p{}297M & 321M  \\
\pair{French}{\EN{}}      & 34.3M & 1044M & 912M \\     
\pair{Czech}{\EN{}}       & 40.1M & \p{}491M & 563M \\     
\pair{Arabic}{Russian}    & 10.2M & \p{}243M & 252M \\
\pair{French}{Russian}    & 10.0M & \p{}295M & 238M \\
\pair{Spanish}{French}    & 10.0M & \p{}297M & 288M \\
\pair{Spanish}{Russian}   & 10.0M & \p{}300M & 235M \\
\bottomrule
\end{tabularx}
\end{center}
\label{tab:corpora_sizes}
\end{table}

Automatic language identification has better results for longer segments. Fortunately, the CzEng contains information about the original paragraph. We checked the language of each of the paragraphs separately and removed all paragraphs that were identified as a different language than it should be. We removed 4M sentence pairs. 
This way, we filtered mostly the noisy sentences and avoided removing a large part of clean ones. Therefore the corpus still contains some noisy sentences.

We published the updated version as CzEng 1.7.\footnote{\url{http://ufal.mff.cuni.cz/czeng/czeng17}} 

Furthermore, for our experiments with transfer learning, we filtered the CzEng even more by removing short sentence pairs of 3 or fewer words, because these sentences are mostly fragments from subtitles. We also removed the longest sentences with more than 75 words as they were often only lists of items or countries\footnote{Only 0.5M parallel sentences has been removed due to having the length longer than 75 words.}. The final size of the corpus we use in our work is 40.1M sentence pairs.
% \XX{Martin: Hodila by se tu tabulka s CzEng1.6 a 1.7 (a možná i 1.0) a s tím, co nakonec používáš, kde by krom počtu vět byly i počty en a cs slov (případně i subwords). Třeba by mě mohlo zajímat, o kolik slov jsi přišel vyfiltrováním těch cca 18M vět, co byly moc dlouhé či krátké. Předpokládám, že většina byla těch dlouhých, takže poměrně jsi na slova toho odfiltroval víc než na věty.}

\subsection{Other Datasets}
\label{sec:other_datasets}

We experiment with various languages across this thesis to show the generality of proposed methods. For various experiments, we select a representative subset of languages having various sizes of corpora, relatedness, translation performance, and writing scripts. Furthermore, the choice of languages was influenced by various shared tasks, where we participated (see \inparciteA{kocmi2018wmt, kocmi2018iwslt, kocmi2019wmt}).

The main criterion is the size of training corpora. We compare low-resource and high-resource language pairs spanning several orders of magnitude of training data sizes. The smallest dataset is the \pair{Odia}{English} with the size of 27k sentence pairs, and the biggest is the \pair{Czech}{English} with 40.1 million sentences. The sizes of the training datasets are in \cref{tab:corpora_sizes}.

Although the \pair{Basque}{English} has a comparable number of sentence pairs as \pair{Estonian}{English}, it has only a third of the total number of words. This is due to many segments not containing a complete sentence.

The second criterion behind the selection of languages is to include language pairs reaching various levels of translation quality. This is indicated by automatic scores of the baseline setups ranging from 3.54 BLEU (\transl{English}{Odia}) to 36.72 BLEU (\transl{English}{German}),\footnote{The systems submitted to WMT 2018 for \transl{English}{German} translation have better performance than our baseline because we decided not to use Commoncrawl. Thus we made \pair{German}{English} parallel corpus artificially less resourceful.} see \cref{tab:coldstart_results_direct} on page \pageref{tab:coldstart_results_direct}.
% \XX{jindra: predevsim u nemciny a francouzstiny je potreba jak se to ma k WMT datum pro jednotlive roky a jestli je mozne to porovnat s Vaswani nebo ne}

The third criterion is language relatedness. In particular, Estonian and Finnish (paired with English) are linguistically related. Another pair of languages is Czech and Slovak, which are closely related languages with more parallel sentences. 

The fourth criterion is the writing script because our methods do not need transliteration as it was a case of previous approaches. Additionally to Latin, we use languages written in Cyrillic, Brahmic, and Arabic. We present language type, number of speakers, and the writing script in \cref{tab:language_categories}.

\begin{table}[t]
\caption{Language family, branch, number of speakers, and the writing script according to \perscite{simons2018ethnologue}.}
\begin{center}
\begin{tabularx}{\textwidth}{@{}l@{\hskip 1cm}rYYY@{}}
\toprule
Language    & Lang. family & Lang. branch & Speakers & Script \\
\midrule
English  & Indo-European & Germanic & 1132M & Latin \\
German   & Indo-European & Germanic & 132M  & Latin \\
French   & Indo-European & Romance  & 280M  & Latin\\
Spanish  & Indo-European & Romance  & 534M  & Latin \\
Czech    & Indo-European & Slavic   & 11M   & Latin \\
Russian  & Indo-European & Slavic   & 258M  & Cyrillic \\
Slovak   & Indo-European & Slavic   & 5M    & Latin\\
Gujarati & Indo-European & Indic    & 61M   & Brahmic \\
Odia     & Indo-European & Indic    & 34M   & Brahmic \\
Arabic   & Afro-Asiatic  & Semitic  & 274M  & Arabic\\
Estonian & Uralic        & Finnic   & 1M    & Latin\\
Finnish  & Uralic        & Finnic   & 5M    & Latin\\
Basque   & Basque        & Basque   & 1M    & Latin \\
\bottomrule
\end{tabularx}
\end{center}
\label{tab:language_categories}
\end{table}

For most of the language pairs, we use training data from WMT \parcite{findings2018wmt}.\footnote{\url{http://www.statmt.org/wmt18/}} We use the training data without any preprocessing, not even tokenization.\footnote{While the recommended best practice in past \ac{WMT} evaluations was to use Moses tokenizer, it is not recommended anymore for Tensor2Tensor with own build-in tokenizer.}

We use most of the training, development, and testsets from \ac{WMT}.\footnote{\url{http://www.statmt.org/wmt18/}} The complete list of corpora is in \cref{tab:dataset_sources}.

For \pair{Basque}{English}, we use all the available data allowed by the organizers of IWSLT 2018 \parcite{niehues2018iwslt}. In addition to the resources suggested by the organizers, we used the allowed data from OPUS and \ac{WMT}, specifically, corpora PaCo2 \pair{Basque}{English} \parcite{san2012paco2} and QTLeap Batches 1-3 from \ac{WMT} IT Translation.\footnote{\url{http://www.statmt.org/wmt16/it-translation-task.html}}

Our \pair{Slovak}{English} experiments use the corpus from \perscite{galuvscakova2012improving}, detokenized by Moses.\footnote{\url{https://github.com/moses-smt/mosesdecoder}}

The language pairs \pair{Arabic}{Russian}, \pair{French}{Russian}, \pair{Spanish}{French}, and \pair{Spanish}{Russian}, use UN corpus \parcite{ziemski2016united}, which provides over 10 million multi-parallel sentences in 6 languages.

\begin{table}[t]
\caption{Corpora used for each language pair in training set, development set, and the test set. The names specify the corpora from WMT News Task data except of languages from various papers.}
\begin{center}
\scriptsize
\begin{tabularx}{\textwidth}{@{}lrYY@{}}
\toprule
Language pair    & Trainset    & Devset    & Testset \\
\midrule
\pair{English}{Basque} & IWSLT 2018 & IWSLT dev 2018 & IWSLT 2018 \\  
\pair{English}{Estonian} & Europarl, Rapid & WMT dev 2018 & WMT 2018 \\    
\pair{English}{Finnish} & Europarl, Paracrawl, Rapid & WMT 2015 & WMT 2018\\   
\pair{English}{German} & Europarl, News commentary, Rapid & WMT 2017 & WMT 2018\\   
\pair{English}{Odia} & \perscite{parida2018odiacorpus} & \perscite{parida2018odiacorpus} & \perscite{parida2018odiacorpus} \\   
\pair{English}{Russian} & News Commentary, Yandex, and UN Corpus & WMT 2012 & WMT 2018\\
\pair{English}{Slovak} & \perscite{galuvscakova2012improving} & WMT 2011 & WMT 2011 \\  
\pair{English}{French} & Commoncrawl, Europarl, Giga FREN, & WMT 2013 & WMT dis. 2015\\      
                       & News commentary,  UN corpus & & \\
\pair{English}{Gujarati} & Bible, Dictionary, Govincrawl, & WMT dev 2019 & WMT 2019 \\    
                         & Software, Wiki texts, and Wiki titles  &  &   \\    
\bottomrule 
\end{tabularx}

\end{center}
\label{tab:dataset_sources}
\end{table}

\ac{NMT} suffers when the training data is not clean \parcite{koehn2018filtering}. Based on our previous experiments, we exclude the noisiest corpus, i.e. web crawled ParaCrawl or Commoncrawl. Furthermore, language pairs with training sentences shorter than four words or longer than 75 words on either the source or the target side are removed to allow for a speedup of Transformer by capping the maximal length and allowing a bigger batch size. The reduction of training data is small, and it does not change the performance of the translation model. 

In contrast, for \pair{French}{English} we keep all \ac{WMT} 2018 corpora and perform a quick cleaning using language detection by Langid.py \parcite{lui2012langidpy}.
We drop all sentences that are not recognized as the correct language. This cleaning removes 6.5M sentence pairs from the \pair{French}{English} training corpus.

We often abbreviate English as \EN{} to visibly differentiate it from the second language.

\section{Machine Translation Evaluation}
\label{sec:machine_translation_evaluation}

In order to evaluate how successful the machine is in translating, we need to define what is considered a good translation. It inherently leads to defining when two texts, in a different language, constitute the equivalent meaning. It is a difficult task, and we would need to delve into complex theoretical questions, which is out of the scope of this thesis. Thus \ac{MT} researchers usually evaluate \ac{MT} translations by comparing it to the expert human translations. 

\ac{MT} evaluation is usually focusing on two main aspects called fluency and adequacy. Whenever the system produces syntactically well-formed sentences (i.e. high fluency) and does not change the semantics, the meaning of the source sentence (i.e. high adequacy), it is considered as a good translation \parcite{hovy2002introduction}. Various ways of measuring the fluency have been proposed, and new metrics are annually evaluated in the \ac{WMT} shared task \parcite{ma2018metrics}. As for the adequacy, it is more complicated since multiple correct translations are possible, and therefore, it is mostly evaluated by conducting the manual evaluation by humans, which is time-consuming and costly. 

\perscite{bojar2013scratching} created a method for the generation of millions of possible references that could solve the problem with having only a limited number of references (usually only one). However, their testset is restricted only to 50 prototype sentences.

\subsection{Manual Evaluation}

Manual evaluation campaigns are run each year at \acf{WMT} to assess translation quality of both academic and commercial systems. This evaluation is considered a benchmark for identifying state-of-the-art systems of a given year.

Manual evaluation utilizes human ability to judge what is a good translation without a rigorous definition. Throughout the years, the \ac{WMT} manual evaluation has changed several times based on findings from previous years. For example, comparing multiple systems together, binary yes/no decision about the translation, or by directly rating one translation at a time as is the case of last three years \parcite{graham2017can}. The main idea across the approaches remains similar: Crowd-sourced judges are asked to rank presented outputs from various systems based on their intuition. 

However, due to the high demand for cost and time, we are not using the manual score in this thesis. Instead, we rely on automatic metrics that try to replicate human behavior as closely as possible.

\subsection{Automatic Metrics}
\label{sec:automatic_metrics}

Automatic metrics for \ac{MT} evaluations are often based on the estimation of similarity between the system output, and a human-produced reference translation. 
The most often used metric is the BLEU score \parcite{papineni2002bleu}. It is based on comparing $n$-grams of sentence units, typically words, between the system output and one or more reference sentences. 
It is computed as the geometric mean of $n$-gram precisions for $n=1...N$ with penalization for short translations by brevity penalty, according to the following formula:

\begin{equation}
  BP = 
  \begin{cases}
    1       & \quad \text{if } L_{sys} > L_{ref}\\
    e^{\left(1-L_{ref}/L_{sys}\right)}  & \quad \text{if } L_{sys} \leq L_{ref}
  \end{cases}
\end{equation}

\begin{equation}
  BLEU = BP \cdot \exp \left(\sum_{n=1}^{N} w_n \cdot \log{p_n}\right)
\end{equation}

where $w_n$ is a positive weight summing to one, usually $\frac{1}{N}$. $L_{ref}$ denotes the length of the reference text that is closest in length to the system output, $L_{sys}$ is the length of the system output. 
The standard value of $N$ for BLEU is 4, different $n$-gram lengths are rarely used. 
The $n$-gram precision $p_n$ is computed by dividing the number of matching $n$-gram in the system output by the number of considered $n$-grams. The number of $n$-gram matches are clipped to the frequency in the reference when $n$-grams occur multiple times. BLEU is a document-level metric. Thus the counts of confirmed $n$-grams are collected for all sentences in the document (or testset) and then the geometric mean of n-gram precision is computed from the accumulated counts.

% \begin{equation}
%   p_n = \frac{
%   \sum_{sentences} \text{number of matching n-grams in sentence}
%   }{
%   \sum_{sentences} \text{number of n-grams in sentence}
%   }
% \end{equation}

It is more informative to compare system output against several references, but it is expensive to obtain multiple references. Thus only one reference is used in most of the testsets.

There is a numerous criticism that has been observed of BLEU. For instance:

\begin{itemize}
  \item use of a geometric mean, which makes the score 0, when there is no match at any of the $n$-gram levels (usually a problem of short testsets);
  \item gives no credit for synonyms or different inflected forms of the same word;
  \item does not consider the importance of various $n$-grams;
  \item it is too sensitive to tokenization.
\end{itemize}

There are several studies on the reliability of BLEU \parcite{callison2006reevaluation, bojar2010tackling}, which inspired the development of other metrics \parcite{lavie2007meteor, ma2018metrics}. Despite the criticism and other possible metrics, BLEU remains the standard metric for automatic evaluation of \ac{MT} systems.

There are several implementations of BLEU that differ in various tokenization details, case-sensitivity, and other details that lead to different resulting scores. \perscite{post2018sacrebleu} made a call for clarity in reporting BLEU score and implemented SacreBLEU\footnote{\url{https://github.com/mjpost/sacreBLEU}} tool for more comparable results. The evaluation script automatically downloads reference testset and computes performance using various metrics. In this work, we use SacreBLEU with the same setting whenever we are reporting numerical results.\footnote{SacreBLEU signature is: \texttt{BLEU+case.mixed+numrefs.1+smooth.exp+tok.13a+ version.1.2.1}}

We report BLEU score multiplied by 100 as is it usual in most papers instead of values on the interval 0 to 1 as originally described by \perscite{papineni2002bleu}.

\subsection{Statistical Significance}
\label{sec:statistical_significance}

If two translation systems differ in BLEU performance, it does not necessarily mean one is significantly better than the other. 
% Especially in \ac{NMT}, the fluctuations in performance after each training step are high (see any of the learning curves in the thesis).
% The improvement in performance can also be attributed to factors contained in the testset. For example, one system can generate near-perfect translations of only a subset of factors, such as a domain or named entities and fail on others. Then it may look like it is superior to the system that translates comparatively all sentences in general.
In order to indicate the actual quality, \perscite{koehn2004bootstrap} proposed to use the paired bootstrap resampling method to compute the statistical significance and validate the superiority of one of the systems. The method repetitively creates testsets by drawing sentences from the original testset randomly with repetition and evaluating the automatic score. Then it computes statistical confidence overall scores comparing various \ac{MT} systems.

Whenever we talk about statistical significance in this work, we tested the compared systems by paired bootstrap resampling with 1000 samples and the confidence level of 0.05. In the results, we label it with a \significantmark{} symbol and comment it in the text. We perform significance tests, usually comparing the baseline and an examined system (if not specified otherwise in the text).

We do the pair-wise statistical testing as is customary in \ac{NMT}.
\chapter{Neural Machine Translation}
\label{chap:nmt}

In this section, we describe \ac{NMT}. In \cref{sec:word_embedding}, we explain in detail word embeddings, \ac{NMT} part that is crucial for our work. We describe subword representation and how \ac{NMT} handles \ac{OOV} words in \cref{sec:subword_representation}. Then we describe the whole \ac{NMT} architecture in \cref{sec:nmt_architectures}. We describe the toolkit we used and detailed settings for our model in \cref{sec:used_architecture}. Lastly, we describe how the progress of \ac{NMT} training is measured and propose stopping criterion in \cref{cref:measuring_progress}.

\section{Word Embeddings}
\label{sec:word_embedding}

Neural networks work in continuous space. When used for \ac{NLP} tasks, we need to bridge the gap between the world of discrete units of words and the continuous, differentiable world of neural networks.

Originally, the first step was to use a finite vocabulary, where each word had a different index, which was then represented as a one-hot vector of the size equal to the number of items in vocabulary. Sizes of 10--90k words were used in \ac{NLP}. However, the one-hot representation, a vector containing only zeros except at a single position with one, is not continuous and differentiable. Furthermore, it does not generalize such that similar words are closer together within the representation.

One way of compressing the one-hot vectors is to assign each word a specific dense vector through an \ac{NN} layer can be called \emph{lookup tables} or \emph{word embeddings}.

Embeddings \parcite{bengio2003neural} are dense vector representations of words commonly of 100-1000 dimensions. They are trained jointly with the whole network and learn word-specific features and cluster the words in the space so that similar words have vectors that are close to each other. 

\begin{figure}
\caption{Thirty nearest neighbors in cosine similarity for the word ``woman'' visualized in 2D by principal component analysis. The large color clusters were added manually for better presentation. The representation is from the encoder \ac{BPE} subword embeddings of our \transl{Czech}{English} model. This figure shows that the 30 nearest neighbors are variants of the word ``woman''. Interestingly there are two separate clusters for Czech and English words (blue and pink), which suggests that \ac{NN} understands equivalence across languages. Furthermore, there are clusters dividing words for adult women and young women (green and orange). Worth of mentioning is the subword ``kyn\v{e}'', which is a Czech ending indicating the feminine variant of several classes of nouns, e.g. professions. It appears in the ``young women'' cluster probably because of the common word ``p\v{r}{\'i}telkyn\v{e}'' (girlfriend).}
\begin{center}
\def\svgwidth{\columnwidth}
\scalebox{1.0}{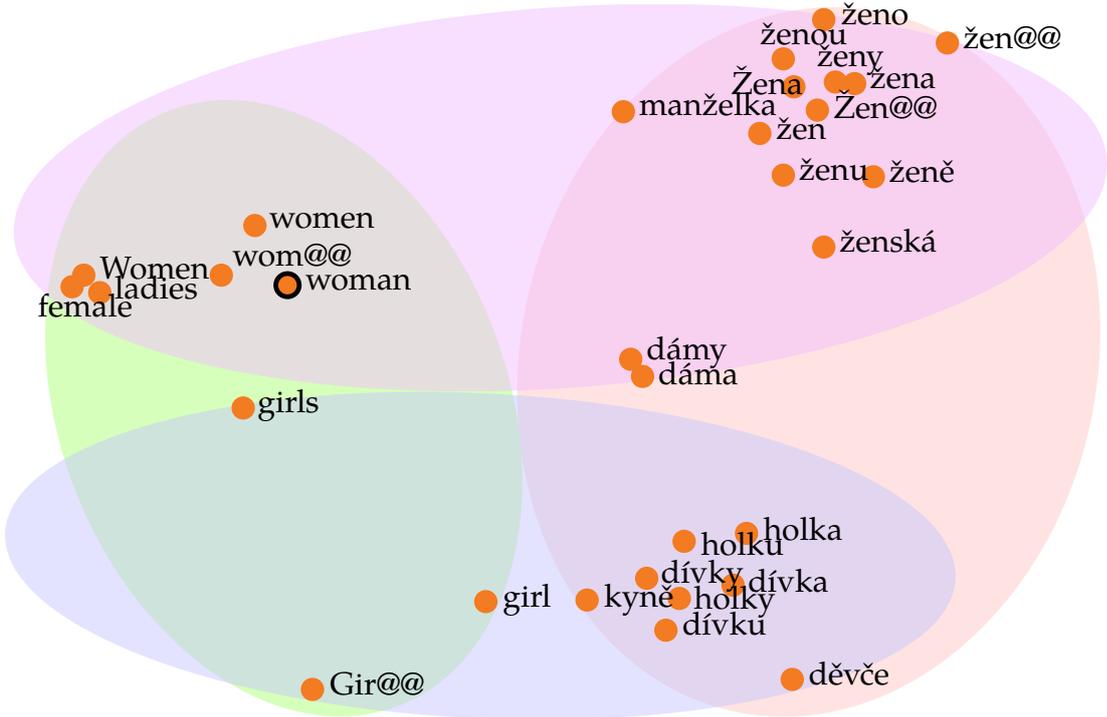}
\end{center}
\label{fig:word_embeddings}
\end{figure}

\perscite{mikolov2013efficient} found that word embeddings in language model \ac{NN} contain semantic and syntactic information without being trained to do so. An example of embedding clustering the space of words is in \cref{fig:word_embeddings}.

Word embeddings can exhibit an interesting correspondence between lexical relations and arithmetic operations in the vector space. The most famous example is the following: 
\def\wove#1{v(\textit{#1})}
\def\samp#1{`\textit{#1}'}
\[
\wove{king} - \wove{man} + \wove{woman} \approx \wove{queen}
\]
In other words, adding the vectors associated with the words \samp{king} and \samp{woman} while subtracting \samp{man} should be close to the vector associated with the word \samp{queen}. We can also say that the difference vectors $\wove{king}-\wove{queen}$ and $\wove{man}-\wove{woman}$ are almost identical and describe the gender relationship.

\perscite{mikolov2013efficient} noticed that such relations emerge without specific training criteria naturally from training the language model with unannotated monolingual data.

\subsection{Lexical Relations Testset}
\label{sec:lexical_relation_testset}

In order to test lexical relations learned by word embeddings, \perscite{mikolov2013efficient} proposed a testset of question pairs. Each question contains two pairs of words ($x_1, x_2, y_1, y_2$) and captures relations like ``What is to `Paris' ($y_1$) as `Czechia' ($x_2$) is to `Prague' ($x_1$)?'', together with the expected answer `France' ($y_2$). The model is evaluated by finding the word representation that has the nearest cosine similarity to the vector $vec(\text{Czechia}) - vec(\text{Prague}) + vec(\text{Paris})$. If the nearest neighbor is $vec(\text{France})$, we consider the question answered correctly.

\begin{table}[t]
\caption{Examples from \perscite{mikolov2013efficient} testset question types, the upper part are semantic questions, the lower part is considered syntactic by \perscite{mikolov2013efficient}.}
\begin{center}
\begin{tabularx}{\textwidth}{@{}l@{\hskip 3cm}Y@{~--~}X@{}}
\toprule
Question Type & \multicolumn{2}{c}{Sample Pair} \\
\midrule
capital-countries &   Athens & Greece \\
capital-world &  Abuja & Nigeria \\
currency &  Algeria & dinar\\
city-in-state &  Houston & Texas \\
man-woman & boy & girl  \\
\midrule
adjective-to-adverb & calm & calmly \\
opposite &  aware & unaware \\
comparative &  bad & worse\\
superlative &  bad & worst\\
present-participle &  code & coding\\
nationality-adjective &  Albania & Albanian\\
past-tense & dancing & danced \\
plural &  banana & bananas\\
plural-verbs &  decrease & decreases\\
\bottomrule
\end{tabularx}
\end{center}
\label{tab:wordvec_samples}
\end{table}

The \perscite{mikolov2013efficient} testset consists of 19544 questions, of which 8869 are called semantic, and 10675 are called syntactic, and further divided into 14 types, see \cref{tab:wordvec_samples}.

After a closer examination of the dataset, we found out that it does not test what the broad terms syntactic and semantic relations suggest. Questions of only three types cover semantics: predict a city based on a country or a state, currency name from the country and the feminine variant of nouns denoting family relations. \perscite{vylomova2016take} showed that many other semantic relationships could be tested, e.g. walk-run, dog-puppy, bark-dog, cook-eat, and others. For the ``syntactic'' relations, the testset contains mostly frequent words that often regularly form morphological variants (e.g. by adding the suffix `ly' to change an adjective into the corresponding adverb), which decreases the generality of the testset.

We have decided to extend the morphosyntactic relations in the testset by adding a substantial number of morphological variants. We extended the syntactic questions except for nationality adjectives, which are already completely covered in the original testset.

We constructed the pairs taking inspiration in the Czech side of the CzEng corpus (see \cref{sec:czeng}), where explicit morphological annotation allows identifying various pairs of Czech words (different grades of adjectives, words, and their negations among others). Word-aligned English words often shared the same properties. As further sources of pairs, we used various webpages usually written for learners of English. For example, for verb tense, we relied on a freely available list of English verbs and their morphological variations. We have included 100--1000 different pairs for each question set. The questions were constructed from the pairs similarly as by \perscite{mikolov2013efficient}: generating all possible pairs of relation pairs. All combinations lead to millions of questions, so we randomly down-sampled to 1000 instances per question set, to keep the testset in the same order of magnitude as in the original one. Additionally, we decided to extend the set of questions on opposites to cover not only opposites of adjectives but also of nouns and verbs.

\begin{table}[t]
\caption{The statistics of \perscite{mikolov2013efficient}'s and our testset. The size represents the total number of questions in the testset.}
\begin{center}
\begin{tabularx}{\textwidth}{@{}l@{\hskip 2cm}ccc@{}}
\toprule
Testset & Size & Categories & Unique words \\
\midrule
\perscite{mikolov2013efficient} (syntactic) & 10675 & 9 & \p{}537\\
Our testset                                 & \p{}8000 & 8 & 5424\\
\bottomrule
\end{tabularx}
\end{center}
\label{tab:statistics_wordvec_testset}
\end{table}

The comparison of the testsets is provided in \cref{tab:statistics_wordvec_testset}. Our testset contains a similar number of questions per category as the original. However, the number of unique words is more than ten times higher, which is useful in diagnosing rare word relations.

Our testset is publicly available. Further details can be found in our paper \persciteA{kocmi2016subgram}. 

\subsection{SubGram Representation}
\label{sec:subgram_representation}

The Skip-gram model \parcite{mikolov2013efficient} uses one-hot representation of a word in vocabulary as the input vector $x$. The embedding of a word then corresponds to the multiplication of the one-hot vector with the trained weight matrix (the lookup table). Therefore weights $w_i$ of the input word $i$ can be directly used
as word embeddings $E$:

\begin{equation}
E_j = \sum_{i=1}^{|x|}x_i*w_{ij} = w_j
\end{equation}

In \persciteA{kocmi2016subgram}, we propose a substring-oriented extension of Skip-gram model that induces vector embeddings from the character-level structure of individual words. Our approach gives the \ac{NN} more information about the examined word reducing the issue of data sparsity and introducing the morphological information about the word to \ac{NN}.

Our approach provides the neural network with a ``multi-hot'' vector representing the substrings contained in the word instead of the one-hot vector representing the whole word. 

We use a vocabulary of substrings, instead of words, created in the following fashion: first, we take all character bigrams, trigrams, tetragrams, and so on up to the length of the word. This way, even the word itself is represented as one of the substrings. As an indication of the beginning and the end of words, we appended the characters \^{} and \$  to each word. 
Here we provide an example of the segmentation:

\begin{center}
`cat' = \{`\^{}c', `ca', `at', `t\$', `\^{}ca', `cat', `at\$', `\^{}cat', `cat\$', `\^{}cat\$'\}
\end{center}

Using all possible substrings would increase the size of vocabulary beyond a reasonable size. Thus we select only the most frequent substrings based on the frequency in the training data.

In order to generate the vector of substrings, we segment the word and create a multi-hot vector, where ``ones'' indicate word's substrings indices in the vocabulary.
In other words, each word is represented as a multi-hot vector indicating which substrings appear in the word. 

The word embedding is created in the same fashion as in the one-hot representation: by multiplication of the input vector with the weight matrix. We have to keep in mind that each word has a different number of substrings. Thus the embeddings need to be normalized either by sigmoid function or by averaging over the number of substrings. We decided to use the mean value as it is computationally simpler than sigmoid:

\begin{equation}
E_j = \frac{\sum_{i=1}^{|x|}x_i*w_{ij}}{\sum_{i=1}^{|x|}x}
\end{equation}

\noindent
where $x$ is the multi-hot vector, and the summation in denominator represents the number of found substrings of the word. 

\subsubsection{SubGram Results}

We conclude this section with the evaluation of the SubGram on the Lexical relation testsets as described in \cref{sec:lexical_relation_testset}.

\begin{table}
\caption{The accuracy (in \%) of word embeddings on the word similarity testsets. The original testset \parcite{mikolov2013efficient} does not contain many \ac{OOV} words, thus the score cannot be computed. Our testset is constructed in a similar way as the original, although it is more challenging and contains many \ac{OOV} question pairs.}
\begin{center}
\begin{tabularx}{\textwidth}{@{}lCCCC@{}}
\toprule
 & \multicolumn{2}{c}{Skip-gram} & \multicolumn{2}{c}{SubGram} \\
\cmidrule(r){2-3} \cmidrule(l){4-5}
Testset        & All  & Only \ac{OOV} & All & Only \ac{OOV} \\
\midrule
Original semantic  & 47.7 & --   &  \p{}0.0   & -- \\ 
Original syntactic & 42.5 & --   & 42.3 & --  \\
Our testset        &  \p{}9.7 & 0.0 & 22.4 & 1.6\\
\bottomrule
\end{tabularx}
\end{center}
\label{tab:subgram_results}
\end{table}

\cref{tab:subgram_results} reports the results from our paper \parciteA{kocmi2016subgram}. It compares Skip-gram \parcite{mikolov2013efficient} with our SubGram. We trained both with the same framework \parcite{rehurek2010gensim} on the same training data. By comparing both approaches on the original testset, we see that both algorithms reach overall a similar performance in the syntactic pairs. On the other hand, SubGram does not capture the tested semantic relations at all.

When comparing models on our testset (see \cref{sec:lexical_relation_testset}), we see that given the same training set, SubGram significantly outperforms Skip-gram model.
Furthermore, our testset contains many questions with rare words to test the capability to encode \ac{OOV} by our SubGram model. The results show that our model can capture a small fraction (1.6\%) of relations on the \ac{OOV} part of testset compared to flat zero for Skip-gram. The performance on \ac{OOV}s is expected to be lower since the model has no knowledge of syntactic exceptions and can only benefit from regularities in substrings.

To conclude, our model, compared to Skip-gram, can encode even unseen words, has a comparable or better performance on syntactic tasks and shows some performance on the \ac{OOV} part of the testset. It could be useful for \ac{NLP} tasks that do not produce any textual output, for example, sentiment analysis, language identification, or \ac{POS} tagging. However, the approach is not reversible, and there is no simple way to transform embeddings back to word forms, which would be needed in word generation, such as the target side of machine translation. However, the inability to decode the embedding back to the word form and the not so high performance on \ac{OOV}s were the main reasons why we decided not to test the SubGram in \ac{NMT}. In the following section, we describe two approaches for solving the \ac{OOV} problem that later became widely used in the \ac{NMT}.

\subsection{Word Embedding Initialization}
\label{sec:embedding_initialization}

Initialization of weights for various \ac{NN} layers is known to be critical for the final performance of the model. And bad initialization can doom the training altogether \parcite{glorot2010understanding, mishkin2015all}. The rest of the thesis is devoted to this topic, and here we start with the earliest part, word embeddings.

Various studies have provided valuable information on initialization of weights for various parts of the \ac{NN} \parcite{glorot2010understanding, he2015delving, kumar2017weight}. Up until our study \persciteA{kocmi2017icon}, there has been a lack of research examining the initialization of word embeddings, which has distinctive properties compared to inner \ac{NN} layers, e.g. it is the first layer of the network, and its input is a discrete one-hot vector.

Traditionally, word embeddings were initialized either randomly with uniform or normal distribution with small variance and a zero mean. Another option is to take word embeddings from the model trained on the same task's training data called ``self-pretrain'' or on a different task, usually language modeling, which can be trained on abundant monolingual data.

%moved
\begin{table}[t]%[!ht] tohle mozna bude resit jeste neco s tabulkou contentu
\caption{Task performance with various embedding initializations.
The higher the score, the better, except for the \ac{LM} perplexity. The best results for random (upper part) and pretrained (lower part) embedding initializations are in bold. The * marks comparably performing settings in each category (random/pretrained).}
\begin{center}
\begin{tabularx}{\textwidth}{@{}lCCCC@{}}
\toprule
 & \ac{NMT}  & \ac{LM} & TAG  & LEM \\
Initialization & (BLEU) & (Perplexity) & (\%) &  (\%) \\
\midrule
$\mathcal{N}(0,10)$      & \p{}6.93   & 76.95 & 85.2   & 48.4  \\
$\mathcal{N}(0,1)$       & \p{}9.81   & 61.36 & 87.9  & 94.4  \\
Only ones                & 10.63  & 62.04 & 90.2  & 95.7  \\
* $\mathcal{N}(0,0.1)$    & 11.77  & 56.61 & 90.7  & 95.7 \\
* $\mathcal{N}(0,0.01)$   & 11.77  & 56.37 & \textbf{90.8 } & \textbf{95.9 }\\
* $\mathcal{N}(0,0.001)$  & \textbf{11.88} & \textbf{55.66} & 90.5  & \textbf{95.9 }\\
* Only zeros              & 11.65  & 56.34 & 90.7  & \textbf{95.9 }\\
* \perscite{he2015delving}& 11.74  & 56.40 & 90.7 & 95.7  \\
* \perscite{glorot2010understanding}& 11.67  & 55.95 & \textbf{90.8 } & \textbf{95.9 } \\
\midrule
* Word2Vec                & 12.37   & \textbf{54.43} & 90.9 & 95.7 \\
GloVe                   & 11.90   & 55.56 & 90.6  & 95.5 \\
* Self-pretrain           & \textbf{12.61} & 54.56 & \textbf{91.1 } & \textbf{95.9 } \\
\bottomrule
\end{tabularx}
\end{center}
\label{tab:embedding_initialization_results}
\end{table}

In practice, random initialization of embeddings is still more common than using pretrained embeddings, and it should be noted that pretrained embeddings are not always better than random initialization \parcite{dhingra2017comparative}.

In our study \parciteA{kocmi2017icon}, we investigated various random as well as pretrained initialization of embeddings to determine the best approach on four tasks: \transl{English}{Czech} \ac{NMT}, \ac{LM}, part-of-speech tagging (denoted TAG), and lemmatization (LEM). For further details see \persciteA{kocmi2017icon}

The results are presented in \cref{tab:embedding_initialization_results}. We found out that the embeddings are not prone to bad random initialization and do not need to be set precisely to maximize the performance as it is a case for the inner layers \parcite{mishkin2015all}. As long as the variance is low, up until 0.1 for normal distribution, the \ac{NN} trains ideally and the final performance is comparable. 

The most surprising result of our work is that embeddings initialized with only zeros performed equally well as random initializations. If all the weights in the \ac{NN} are initialized with zero, the derivative with respect to loss function is the same for every weight in each layer. Consequently, the network would train to be symmetric at each layer and would not be better than the linear model. However, it is not the case when only embeddings start with zeros, and the rest of the network is initialized randomly with various distributions.

Our results on the transferring of pretrained word embeddings support previous findings that these embeddings improve the performance over random initialization \parcite{kim2014convolutional, lample2016neural}. However, the pretrained embeddings have a disadvantage that the model to which they are transferred has to have the same vocabulary and embeddings dimensions.

To our knowledge, our work was the first that compared various initialization techniques of the embeddings on multiple tasks. However, our work has been done on word-level \ac{NN}. Since then, the research focus shifted the standard approach from the word-level translation to subword-level translation \parcite{sennrich2016bpe}.

% The more variance the randomly initialized embeddings have, the more effort must the neural network exerts to store information in the embeddings space. Above a certain effort threshold, it becomes easier to store the information in the subsequent hidden layers (at the expense of some capacity loss) and uses the random embeddings more or less as a strange \equo{multi-hot} indexing mechanism. On the other hand, initialization with a small variance or even all zeros leaves the neural network free choice over the utilization of the embedding space.

Transferring of pretrained word embeddings have proven to be invaluable for improving performance of natural language tasks that often suffer from lack of training data \parcite{kim2014convolutional, lample2016neural}, thanks to the utilization of unsupervised pretraining on a large quantity of monolingual data. However, it is less common in \ac{NMT} to utilize the pretrained embeddings, mostly because the number of available parallel sentences for various language pairs tends to be several times larger than available annotated data for other \ac{NLP} tasks such as Penn Treebank \parcite{marcus1993building} for parsing. However, \perscite{qi2018pre} showed that the use of pretrained embeddings dramatically improves the performance of the model in the low-resource \ac{NMT} scenario. Especially for the extremely low-resource languages with less than 20k parallel sentences, pretrained embeddings help to improve the performance of up to 10 BLEU points.

Recently, pretrained embeddings have been used in unsupervised \ac{NMT}, where the goal is to train an \ac{MT} system without any parallel sentences \parcite{artetxe2018unsupervised, lample2018phrase} with only the monolingual data in both languages. They used the capability of embeddings to represent the meaning of words (see \cref{sec:word_embedding}), and with the use of linear transformation, they mapped the embedding spaces of the two languages into the same vector space. Then they use this mapping to roughly translate the monolingual sentences word-by-word, and in following steps, they iteratively train the \ac{NMT} system.

In recent years the research focus has shifted from word-level \ac{NMT} to subword-level \ac{NMT}, which made the use of pretrained embeddings in \ac{NMT} obsolete. Although, several notable improvements in language modeling \parcite{devlin2019bert, yang2019xlnet} have appeared recently, which can lead to restoring a research focus in pretrained embeddings.

\section{Subword Representation}
\label{sec:subword_representation}

Traditionally \ac{NMT} systems relied on a vocabulary to store all words used in the translation. The capacity of this vocabulary was typically 10--90k words. However, this is not enough to cover all words in a language. That is why the first \ac{NMT} systems used a special \ac{OOV} symbol as a replacement for remaining rare words.

The \acf{OOV} words are a substantial problem especially for languages with inflection, agglutination or compounding, where many variants of a frequent word become rare. For example consider German compound word `Abwasser$\vert$behandlungs$\vert$anlange' for `sewage water treatment plant' or Czech `velko$\vert$v{\'y}roba' for `mass-production', for which a segmented representation is more informative than one vector for the whole word.

Increasing the size of the vocabulary in order to reduce the number of \ac{OOV} words proportionally increases the training complexity as well as decoding complexity. Moreover, the \ac{NMT} systems will not be able to learn good encoding for uncommon words or word-forms as they are seen only a few times within the training corpus or not at all.

To overcome the large vocabulary problem and avoid the \ac{OOV} problem, translation models need mechanisms that go below the word level. There are two possible solutions. Either by including more information into the representation of words or splitting uncommon words and translating on the level of subword units.

The former approach tries to encode additional information about the word structure or linguistic classes into the representation of a word. \perscite{tamchyna2017modeling} removed the inflection by morphologically annotating training sentences and let the \ac{NMT} to translate only the lemmas and assigned morphological tags. \perscite{luong2016achieving} proposed to use a hybrid approach where first we get the word embeddings from characters followed by standard \ac{NMT} on the computed embeddings. In \persciteA{kocmi2016subgram}, we proposed  to include the substring structure of a word into the word embedding (see \cref{sec:subgram_representation}).

The latter approach breaks uncommon words into subword units that are handled by the \ac{NN} as standalone tokens. The trivial approach is to break the sentence into individual characters, but it needs much longer training times as the number of tokens per training example is several times higher than the number of words, and it creates a problem with long-range dependencies in characters making the character-level translation sub-optimal \parcite{tiedemann2009character,ling2015character}. Thus we need to split the words into the least number of subwords but avoid bloating the size of the subword vocabulary. 

In recent years, several segmentation algorithms have been proposed, however, only the byte pair encoding \parcite{sennrich2016bpe} and wordpieces \parcite{wu2016google} became widely used. We describe both of the methods in the next sections. For the completeness, \perscite{kudo2018sentencepiece} recently developed a segmentation method called SentencePiece, which shows promising improvements in performance over them.

\subsection{Byte Pair Encoding}

Using a word-based vocabulary in \ac{NMT} leads to problems with \ac{OOV}. \perscite{sennrich2016bpe} tackled this problem by segmenting the words into more frequent subword tokens with the use of byte pair encoding \parcite{gage1994new}.

\ac{BPE} is a simple data compression algorithm, which iteratively merges the most frequent pairs of consecutive characters or character sequences. A table of the merges, together with the vocabulary, is then required to segment a given input text.

The table of merges is generated in the following way. First, all characters from the training data are added into the vocabulary plus a special symbol for the word ending `$\langle{}/w\rangle{}$', which is used to restore original segmentation after the translation. Then we add the ending symbol to all words in the training set and separate them to individual characters.
We iteratively find the most frequent symbol pairs and replace them with a new single symbol representing their concatenation. Each merge thus produces a new symbol that represents a character n-gram. We continue until we have the same number of initial characters and merges as is our desired size of the vocabulary. By this process, frequent words become directly included in the vocabulary. A toy example is in \cref{fig:bpe_example}.

\begin{figure}
\caption{\ac{BPE} merges learned from a vocabulary \{`old',`older','wider`\}.}
\begin{center}
\begin{tabularx}{\textwidth}{@{}@{\hskip 5cm}llcl@{}}
\toprule
  r & </w> & $\rightarrow$ & r</w>\\
  o & l &$\rightarrow$ & ol\\
  e & r</w> &$\rightarrow$ & er</w> \\
  d & er</w> &$\rightarrow$ & der</w> \\
  w & i &$\rightarrow$ & wi\\
\bottomrule
\end{tabularx}
\end{center}
\label{fig:bpe_example}
\end{figure}

The merges are applied in advance on the training corpus by merging characters based on learned merges. The \ac{BPE} segments the words into subword tokens, which can be used by \ac{NMT} without any need for architecture modification. In other words, the \ac{NMT} model handles subwords as regular words.

In practice, the symbol for the end of the word is not produced during segmentation. Instead, a `@@' is added to all subword tokens that end in the middle of a word. For example the word `older' would be segmented into `ol@@ der', see \cref{fig:bpe_example}. 

\perscite{sennrich2016bpe} also showed that using joint merges, generated from concatenated trainsets for both the source and the target language, is beneficial for the overall performance of \ac{NMT}. This improved consistency between the source and target segmentation is especially useful for the encoding of named entities, which helps \ac{NMT} in learning the mapping between subword units.

\ac{BPE} implementation has several disadvantages. It cannot address well languages that do not use a space as a separator between words, for example, Chinese. It fails when encoding characters that are not contained in the vocabulary, for example, foreign words written in a different alphabet. Lastly, \ac{BPE} algorithm relies on a tokenizer. Without its use, the punctuation attached directly to words would have different word segmentation than when separated. The wordpiece method \parcite{wu2016google} solves all these problems. We describe it in the next section.

\subsection{Wordpieces}
\label{sec:wordpiece_description}

Wordpiece is another word segmentation algorithm. It is similar to \ac{BPE} and is based on an algorithm developed by \perscite{schuster2012japanese}. \perscite{wu2016google} adopted the algorithm for \ac{NMT} purposes. We describe the algorithm in comparison to \ac{BPE}.

The wordpiece segmentation differs mainly by using language model likelihood instead of highest frequency pair during the selection of candidates for new vocabulary units. Secondly, it does not employ any tokenization leaving it for the wordpiece algorithm to learn.
% \XX{Martin: Máš nějaký toy example, kde BPE a wordpiece dají různý výsledek, právě kvůli této vlastnosti? Do dizertace to není nutné, spíš by mě to zajímalo osobně. Rozdíly kvůli zacházení s mezerou a dalším implementačním detailům jsou mi jasné, ale nad tímto jsem se nikdy nezamyslel - a toy exampl by mi pomohl, protože jsem líný myslet:-).}

% \begin{itemize}
%   \item uses `\_' symbol to mark the start of a word instead of `$\langle{}/w\rangle{}$' used for a word ending;
%   \item does not store the merge table;
%   \item forms new subwords through language model likelihood instead of highest frequency pair.
% \end{itemize}

The algorithm works by starting with the vocabulary containing units of characters and building a language model on the segmented training data. Then it adds a combination of two units from the current vocabulary by selecting the pair of units that increases the likelihood on the training data the most, continuing until the vocabulary contains the predefined number of subword units.

The iterative process would be computationally expensive if done by brute-force. Therefore the algorithm uses several improvements, e.g. adding several new units at once per step or testing only pairs that have a high chance to be good candidates.

The segmentation works in a greedy way when applying. It finds the longest unit in the vocabulary from the beginning of the sentence, separates it and continues with the rest of the sentence. This way, it does not need to remember the ordering of merges; it remembers just the vocabulary. This makes it simpler than \ac{BPE}.

The Tensor2tensor \parcite{tensor2tensor} framework slightly improves the wordpiece algorithm by byte-encoding \ac{OOV} characters, which makes any Unicode character encodable. It uses an underscore instead of `$\langle{}/w\rangle{}$' as an indication of the word endings. 

Furthermore, the implementation by \perscite{tensor2tensor} optimizes the generation by counting frequencies for only a small part of the corpus. We extended it and created vocabularies in this thesis from the first twenty million sentences. Additionally, \parcite{tensor2tensor} introduce a 1\%\footnote{The implementation in \ac{T2T} tries to create vocabulary several times, and if it fails to create a vocabulary within this tolerance, it uses the generated vocabulary with the closest size.} tolerance for the final size of the vocabulary. Therefore instead of having 32000 subwords,\footnote{We use exactly 32000 as a vocabulary size instead of $2^{15}=32768$.} the vocabulary has between 31680 and 32320 items, see \url{https://github.com/tensorflow/tensor2tensor/blob/v1.8.0/tensor2tensor/data_generators/text_encoder.py\#L723}.

Whenever we talk about the wordpiece segmentation in this work, we mean the \ac{T2T} implementation described in this section.

\section{Neural Machine Translation Architectures}
\label{sec:nmt_architectures}

There were early attempts to use neural networks in machine translation \parcite{waibel1991janus,forcada1997recursive}. However, the rise of neural networks in \ac{MT} came two decades later, when the hardware resources became able to handle large models with millions of parameters.

A shift in paradigm in \ac{MT} happened when \ac{NMT} end-to-end model scored higher that previous \ac{PBMT} models \parcite{jean2015using,sennrich2016edinburgh}.

One of the first end-to-end \ac{NMT} systems was \perscite{sutskever2014sequence}. They used \ac{LSTM} \ac{RNN} model that processes one word at a time until it reads the whole input sentence. Then a special symbol ``<start>'' is provided and the network produces the first word based on its inner state and the previous word. This generated word is then fed into the network, and the second word is generated. The process continues until the model generates the ``<end>'' symbol.

The main disadvantage of the work of \perscite{sutskever2014sequence} is that the network has to fit the whole sentence into a vector of 300--1000 dimension before it starts generating the output. Therefore \perscite{bahdanau2015neural} proposed the so-called attention mechanism. The attention mechanism gives the network the ability to reconsider all input words at any stage and use this information when generating a new word. 

\perscite{gehring2017convolutional} redesigned the previous architecture with \ac{CNN}, which handles all input words together, therefore making the training and inference process faster.

\perscite{vaswani2017attention} completely redesigned the \ac{NMT} architecture and introduced Transformer model, which uses feed-forward layers in contrast to previous architectures that use \ac{RNN} or \ac{CNN} structures. We describe this architecture in detail in the next section.

\subsection{Transformer Model}
\label{sec:transformer}

In our work, we use the Transformer architecture \parcite{vaswani2017attention}. Transformer architecture consists of an encoder and decoder, similarly as the previous approaches. The encoder takes the input sentence and maps it into a high-dimensional state space. Its output is then fed into the decoder, which produces output sentence. However, instead of going one word at a time from left to right of a sentence, encoder sees the entire input sequence at once.
This makes it faster in terms of training and inference speed in comparison to previous neural architectures because it allows better usage of parallelism. The decoder remains ``autoregressive'', i.e. always producing the output symbol with the knowledge of the previously produced output symbol. Non-autoregressive models \parcite{libovicky2019nonautoregressive} are still an open research question.

\subsubsection{Transformer Attention}

The idea of attention mechanism \parcite{bahdanau2015neural} is to look at the input sequence and decide which words are important for the generation of a particular output word. The novel idea of self-attention is to extend the mechanism to the processing of input sequences and output sentences as well. In other words, it helps the model to understand the word it is currently processing with the use of relevant words from its context.

For example, when processing the sentence ``The kitten crawled over the room because it was hungry.'', \ac{NMT} needs to know the antecedent of the word ``it''. The self-attention mechanism solves this problem by incorporating the information into the representation of the word ``it'' at deeper layers of the encoder.

In general form, the Transformer attention function uses three vectors: queries (Q), keys (K) and values (V). The output is a weighted sum of values, where weights are computed from queries and keys. The attention is defined as follows:

$$\text{Attention}(Q,K,V)=\text{softmax}(\frac{QK^T}{\sqrt{d_k}})V$$

where $d_k$ is the square root of the dimension of the key vectors, which is normalization necessary to stabilize gradients. 

The intuition behind the attention is that we get a distribution over the whole sequence using the dot product of queries (which are a hidden state of all positions in the sequence) and keys followed by softmax. This distribution is then used to weight the values (which encode a hidden state similarly like queries). It results in a vector, where relevant words or their features are stressed.

The attention is used separately in encoder and decoder as a self-attention where all queries, keys, and values come from the previous layer. It is also used in encoder-decoder attention, where queries and keys come from encoder and values from the decoder.

\subsubsection{Multi-Head Attention}

Having only one attention, \ac{NMT} would focus solely on some positions in the previous layer, leaving other relevant words ignored or conflating mutually irrelevant aspects into one overused attention. Transformer model solves this by using several heads within each layer, each with its own linear transformation, which leads to the concurrent observation of different parts of the input.
% \XX{I one-head attention funguje, je to jen o pár BLEU horší. Attention má typicky trochu plošší distribuci než je medián z multi-head (tipuji). Fór je v tom, že encoder má několik layers, a pokud se v první vrstvě dostane kontext do nejbližších slov, v další vrstvě se už může propagovat dál, i kdyby byla attention velmi ostře zaměřená. Krom toho některé layers mohou mít tu plošší attention a dívat se "všude". Transformační matice Q, K a V a skip connections umožní, že toto taky funguje. Tedy ten problém one-head attention není tak zásadní, jak se z té tvé věty může zdát. Ale v zásadě máš pravdu, tak to asi nech.}

Formally, the multi-head attention is defined as follows:

$$\text{MultiHead}(Q,K,V)=\text{Concatenate}(\text{head}_1, ... , \text{head}_h)W^O$$
$$\text{where head}_i=\text{Attention}(QW_i^Q,KW_i^K,VW_i^V)$$

where the projection matrices $W^{Q/K/V}$ are trainable matrices different for each attention head and $h$ is the number of heads, sixteen in the ``Transformer-big'' model. The concatenation in multi-head attention is then linearly projected by a matrix $W^O$.

\subsubsection{Positional Encoding}

With the use of attention mechanism, one thing is missing from the model. It is the information about the position of each word. This is solved by adding a special positional encoding to all input words, which helps \ac{NMT} to identify the word order.

The absolute position encoding of a word $pos$ is defined as:

$$PE_{(\text{pos},2i)}=\text{sin}(\text{pos}/10000^{2i/d_{model}})$$
$$PE_{(\text{pos},2i+1)}=\text{cos}(\text{pos}/10000^{2i/d_{model}})$$

where $i$ is the dimension in the positional encoding $PE$, in other words, each dimension of $PE$ corresponds to a sigmoid.

In \cref{sec:word_embedding}, we explained how discrete words are mapped to embedding vectors. The positional encoding is added to the word embeddings and used as the input to the first layer of Transformer.

\begin{figure}
\caption{Transformer architecture. The image is taken from \perscite{vaswani2017attention}.}
\begin{center}
\includegraphics[scale=0.5]{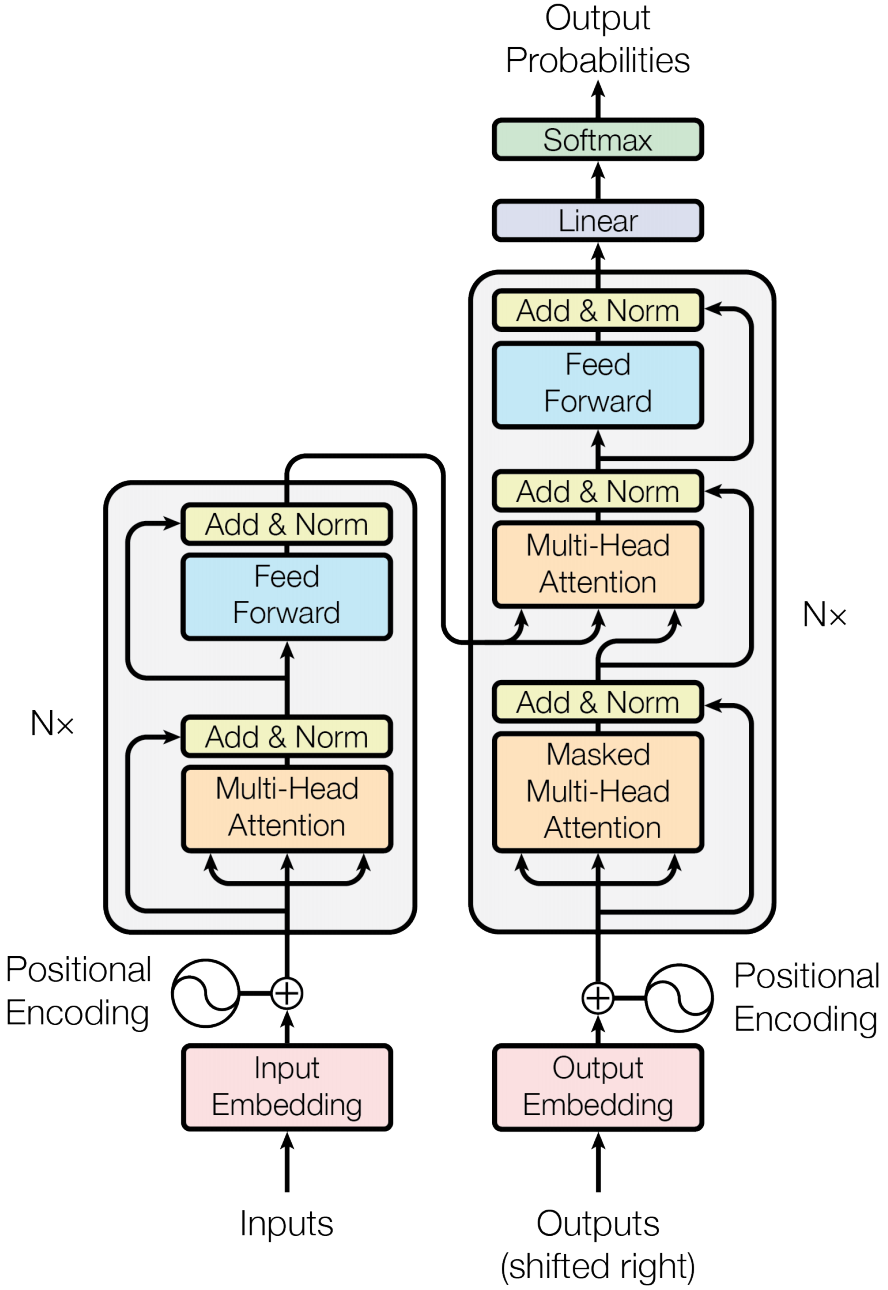}
\end{center}
\label{fig:transformer}
\hrulefill
\end{figure}

\subsubsection{Complete Architecture}

The complete architecture is illustrated in \cref{fig:transformer}. Except parts mentioned above, there is a residual connection after each multi-head attention, which sums input of multi-head attention with its output followed by layer normalization \parcite{ba2016layer} (it is labeled as ``add \& norm'' in \cref{fig:transformer}). The model stacks several layers of multi-head attention on top of each other, with position-wise feed-forward networks. In the original model, six layers are used.

The output of the decoder is finally modified by linear transformation followed by the softmax function that produces probabilities of words over the model vocabulary.

For further reading, see the original paper \parcite{vaswani2017attention} or various blog posts describing the model.\footnote{\url{http://jalammar.github.io/illustrated-transformer/}}

\section{Neural Machine Translation Model Setting}
\label{sec:used_architecture}

In our earlier works on \ac{NMT}, we used the Neural Monkey toolkit and the \ac{RNN} model with an attention mechanism. The Neural Monkey has been developed mainly for quick prototyping, see further details in \persciteA{neuralmonkey}. 

\perscite{vaswani2017attention} published their work with the Transformer model, which showed significant gains in the training time and performance. Furthermore, they published the \ac{T2T} framework \parcite{tensor2tensor}. For the majority of this thesis, we use the \ac{T2T} framework.

During our research, \ac{T2T} has been rapidly developed and improved. Therefore, we changed to the newer version twice throughout our research. This means that the results should not be directly compared as any change in the network architecture or training regime may lead to large changes in the performance.

In this thesis, each section uses one of three different settings that we explain in this section. Each of the settings has been used in our papers, and results are comparable within each setting. However, the results for same language pair should not be compared across settings. We use the following setups:

\begin{itemize}
  \item \textbf{\TFA{}} -- this setup uses \ac{T2T} in version 1.4.2. This setting is used in the ``warm-start'' experiments in \persciteA{kocmi2018trivial, kocmi2018wmt, kocmi2018iwslt, kocmi2018wat}.
  \item \textbf{\TFB{}} -- this setup uses \ac{T2T} in version 1.8.0 and concerns around the ``cold-start'' experiments in \persciteA{kocmi2019ecofriendly}.
  \item \textbf{\TFC{}} -- our latest setup uses \ac{T2T} in version 1.11.0 and is involved in \persciteA{kocmi2019wmt}. Furthermore, we use this setup for experiments that are not yet published.
\end{itemize}

We use a model based on the ``Big single GPU Transformer'' setup as defined by \perscite{vaswani2017attention} with a few modifications. We set maximal length of a sentence to 100 wordpieces. We set a length normalization penalty to 1. The individual versions use additional modifications:

\begin{itemize}
  \item \textbf{\TFA{}} uses Adam optimizer \parcite{adam_optimizer} with 32000 linear warm-up steps, and batch size of 2300 tokens.
  \item \textbf{\TFB{}} uses Adafactor optimizer \parcite{shazeer2018adafactor} with 8000 linear warm-up steps, batch size of 2900 tokens, and disabled layer dropout.
  \item \textbf{\TFC{}} uses Adafactor optimizer with 16000 linear warm-up steps, batch size of 4500 tokens, and disabled ``layer dropout''.
\end{itemize}

For details about individual parameters, see \perscite{tensor2tensor}.

\section{Measuring Training Progress}
\label{cref:measuring_progress}

The progress of \ac{NN} training can be measured in various ways. We can measure, for example, the wall-clock time passed, the number of processed training examples or the financial cost of the training. Each method is more relevant for various applications.

The most common approach is to report the number of processed training steps. It can be reported either by the number of individual seen examples, number of training epochs or as a number of batches.
The individual examples are preferred in tasks, where each example has the same informative value. In \ac{MT}, one training example is usually a sentence pair, which can be of varying length. 
Reporting the number of epochs does not rely on the ordering of sentences in corpus because it is reported after each complete pass over the training corpus. Lastly, we can report the number of batches, also called training steps, which specifies the number of updates (error backpropagation) through the network.

The batch size can be either fixed to a given number of sentence pairs (e.g. in Neural Monkey \inparciteA{neuralmonkey}) or to a number of subword tokens in all sentences (e.g. \inparcite{tensor2tensor}). The former definition correlates with measuring the number of training examples. However, the latter is optimized for training as it can better use the available GPU memory.

Another option is to measure the time passed to achieve a given result. \perscite{popel2018training} use this reporting of wall-clock time and justify it by stating that time computed as steps per second, fluctuates at most by 2\% during the training. Wall-clock time can be more informative than reporting a number of seen examples since the training step can contain a variable number of sentences or tokens. However, the training time is heavily influenced by the hardware and the other load on a given machine.\footnote{The \ac{NN} research is usually carried out on clusters, where several people run various processes. Even when the GPU is allocated only for a given job, other shared resources, like CPU or network disks, can slow down the training process.}

Lastly, practitioners are primarily concerned with the error rate and the cost they need to pay to achieve that error level. It is referred to as a hardware cost \parcite{shallue2018measuring}. This cost can be measured by multiplying the number of training steps by the average price per one step. It heavily depends on the respective hardware, but the number of training steps is hardware-agnostic and can be computed for any hardware given the average cost per step.

We run our experiments on various types of hardware. Our department's cluster contains three types of GPU cards: NVidia GeForce GTX 1080, the Titan version 1080Ti and Quadro P5000. We use the \ac{T2T} framework, which uses batches of a varying number of sentences but approximately the same number of tokens. Thus reporting the training time is not comparable due to various machine setups and the number of examples or epochs is not exact due to the \ac{T2T} batching behavior. We report the number of training steps, the actual updates of the \ac{NN}.

\subsection{Convergence and Stopping Criterion}
\label{sec:stopping_criterion}

Training of \ac{NMT} models is complicated by the fact that the learning curves, showing the performance of the model over the learning period on a fixed development set, usually never fully flatten or start overfitting on reasonably large datasets. 
Signs of overfitting\footnote{By overfitting we mean the situation, when the performance on the training set is improving, but the score on the development set is worsening.} are noticeable in low-resource settings only, as discussed in \cref{sec:lowresource_definition}. Especially with the recent models trained on a large parallel corpus, we can get some improvements, usually around tenths of a BLEU point, even after several weeks of training and the learning curve will not fully flatten out.

The common practice in machine learning is to use a stopping criterion. One option is to set the maximum number of training steps or epochs. The second option is to evaluate the model every X steps (or minutes) on the development set and stop the training whenever the last N updates do not improve the performance by at least some small delta.

The former approach relies on an intuition of how long approximately is enough to train the model. Usually, more complex models need more time to reach the maximal score, and an incorrectly set number of steps could stop the training prematurely. The latter approach is sensitive to the number of steps (and their duration) between individual evaluations: if the evaluations are too close to each other, the training can stop too early due to the training fluctuations, and when they are too far apart the training would not stop in a reasonable time on big datasets.

Many papers do not specify the stopping criteria or only mention an approximate time or the number of steps for how long the model was trained \parcite{bahdanau2015neural, vaswani2017attention}. Presumably, the models are trained until no apparent improvement is visible on the development set. However, this stopping criterion is not perfect since the models could be stopped at various stages of training, and the comparison could be unfair. 

\begin{figure}
\caption{Examples of real learning curves. Full dots represent the best performance, squares represent our stopping criteria.}
\begin{center}
\input{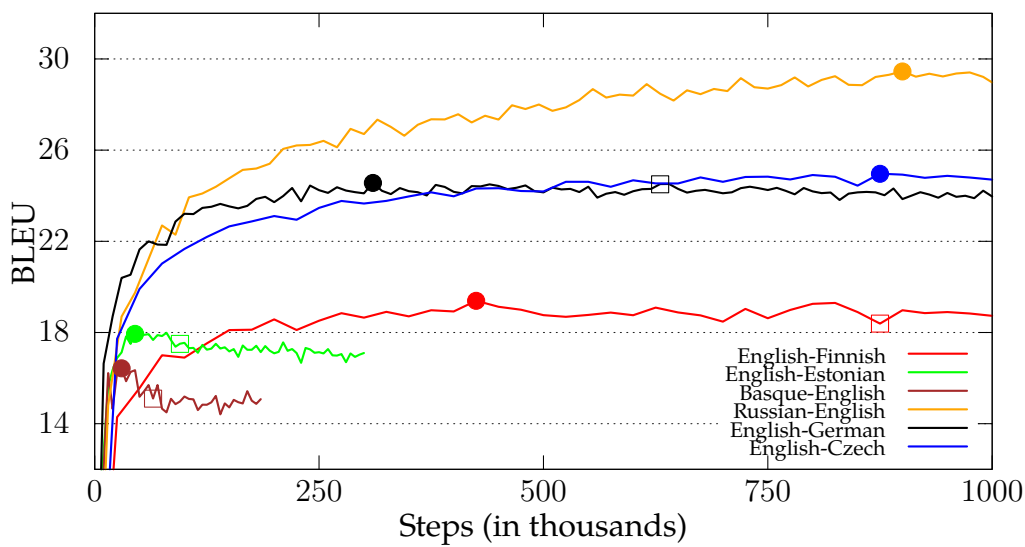}
\end{center}
\label{fig:example_learning_curves}
\end{figure}

In this thesis, we compare low-resource language pairs that converge within 50k steps and high-resource pairs that improve even after 1000k steps. Hence, to avoid premature stopping, we set a high upper level of maximal steps for each language pair. The low-resource languages, with less than 500k training examples, are trained for at most 200k steps. We also evaluate low-resource languages more often. The rest of the languages are trained at most for the 1000k steps. 

However, this stopping criterion usually trains for a much longer time than necessary. Thus later in our work, we defined a more general convergence criterion. We stop the training earlier whenever there was no improvement bigger than 0.5\% of maximal reached BLEU within the past 50\% of evaluations. This criterion is comparable to stopping after X batch updates without any improvement, and it is less sensitive to the number of steps between evaluations as the low-resource languages are evaluated up to ten times more often. Importantly, we set mild conditions for stopping criteria on purpose to prevent the risk of premature judgments.

Our stopping criterion is especially useful for low and middle resource languages. \cref{fig:example_learning_curves} presents with square a point where our training stops. Without any doubt, the model already passed its best performance, and thus the stopping is valid. We can notice that for high-resource languages, the stopping criterion does not trigger within 1000k steps.

With the stopping criteria in mind, once the training stops, we take the best performing model on the development set and report the number of training steps instead of the last step where the training stopped. Additionally, we regularly report full learning curves. Thus the readers can judge by themselves if they would expect any sudden change in the final performance.

\chapter{Transfer Learning}
\label{chap:transfer_learning}
% \addcontentsline{toc}{chapter}{Transfer Learning}

Humans have the ability to utilize knowledge from previous experience when learning a new task. It helps us to learn new skills in a shorter time and with less effort. In fact, the more related a new task is, the faster we learn. In contrast, machine learning algorithms usually learn the task from random initialization on isolated data without any prior knowledge. Transfer learning attempts to change this approach by improving the performance on a new task with the usage of knowledge obtained for solving other tasks \parcite{bahadori2014general,farajidavar2015transductive,moon2017completely}.

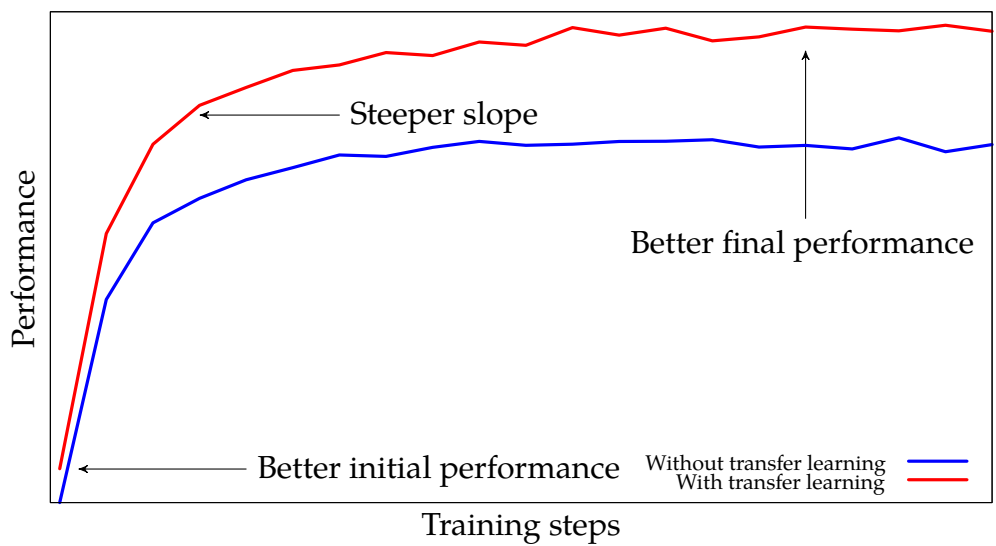
\begin{figure}
\caption{Three impacts where transfer learning improves the training process. These are real learning curves for warm-start transfer learning on \transl{English}{Estonian} (see \cref{sec:warm_start_transfer}).}
\begin{center}
\begin{tikzpicture}[gnuplot]
%% generated with GNUPLOT 5.0p3 (Lua 5.1; terminal rev. 99, script rev. 100)
%% Sun 22 Sep 2019 11:02:32 AM DST
\path (0.000,0.000) rectangle (4.500,7.400);
\gpcolor{color=gp lt color border}
\gpsetlinetype{gp lt border}
\gpsetdashtype{gp dt solid}
\gpsetlinewidth{1.00}
\draw[gp path] (0.554,7.031)--(0.554,0.523)--(12.945,0.523)--(12.945,7.031)--cycle;
\node[gp node left] at (3.130,0.968) {Better initial performance};
\node[gp node left] at (4.357,5.661) {Steeper slope};
\node[gp node left] at (8.038,3.948) {Better final performance};
\draw[gp path,->](3.130,0.968)--(0.922,0.968);
\draw[gp path,->](4.357,5.661)--(2.517,5.661);
\draw[gp path,->](10.491,4.291)--(10.491,6.517);
\node[gp node center,rotate=-270] at (0.246,3.777) {Performance};
\node[gp node center] at (6.749,0.215) {Training steps};
\node[gp node right,font={\fontsize{8pt}{9.6pt}\selectfont}] at (11.699,1.072) {Without transfer learning};
\gpcolor{rgb color={0.000,0.000,1.000}}
\gpsetlinewidth{3.00}
\draw[gp path] (11.846,1.072)--(12.614,1.072);
\draw[gp path] (0.677,0.523)--(1.290,3.216)--(1.904,4.232)--(2.517,4.557)--(3.130,4.803)%
  --(3.744,4.964)--(4.357,5.132)--(4.971,5.113)--(5.584,5.233)--(6.197,5.312)--(6.811,5.260)%
  --(7.424,5.275)--(8.038,5.311)--(8.651,5.314)--(9.265,5.334)--(9.878,5.237)--(10.491,5.259)%
  --(11.105,5.212)--(11.718,5.359)--(12.332,5.175)--(12.945,5.270);
\gpcolor{color=gp lt color border}
\node[gp node right,font={\fontsize{8pt}{9.6pt}\selectfont}] at (11.699,0.826) {With transfer learning};
\gpcolor{rgb color={1.000,0.000,0.000}}
\draw[gp path] (11.846,0.826)--(12.614,0.826);
\draw[gp path] (0.677,0.972)--(1.290,4.090)--(1.904,5.274)--(2.517,5.790)--(3.130,6.027)%
  --(3.744,6.253)--(4.357,6.326)--(4.971,6.490)--(5.584,6.450)--(6.197,6.630)--(6.811,6.586)%
  --(7.424,6.821)--(8.038,6.721)--(8.651,6.813)--(9.265,6.645)--(9.878,6.698)--(10.491,6.828)%
  --(11.105,6.800)--(11.718,6.778)--(12.332,6.852)--(12.945,6.772);
\gpcolor{color=gp lt color border}
\gpsetlinewidth{1.00}
\draw[gp path] (0.554,7.031)--(0.554,0.523)--(12.945,0.523)--(12.945,7.031)--cycle;
%% coordinates of the plot area
\gpdefrectangularnode{gp plot 1}{\pgfpoint{0.554cm}{0.523cm}}{\pgfpoint{12.945cm}{7.031cm}}
\end{tikzpicture}
%% gnuplot variables
\end{center}
\label{fig:comparison_transfer_non}
\end{figure}

\perscite{torrey2010transfer} describe three ways of how transfer learning can improve performance. Specifically:
\begin{itemize}
  \item improving the initial performance at the beginning of training compared to a randomly initialized model when the tasks are similar;
  \item shortening the time needed to reach the maximal performance;
  \item improving the final performance level compared to training the model without the transfer.
\end{itemize}
These three potential improvements are illustrated on learning curves in \cref{fig:comparison_transfer_non}.

The success of transfer learning is not always guaranteed. For example, when transferring from a weakly related task, it may hinder the final performance in the target task. A phenomenon known as the negative transfer has been well recognized by the transfer learning community \parcite{pan2010transfersurvey, wang2019characterizing}. However, there is a lack of research on this phenomenon in the \ac{NMT} field, which we investigate in \cref{sec:negative_transfer}.

In this thesis, we are highlighting various observations expressed in general way. However, we can only claim that these observations are general under the specific settings of the individual experiments. We are not claiming their decisive generality, because proving it across most language pairs, various experiment settings, and other conditions is out of the scope of this thesis. However, we are basing our observations on as broad a set of experiments as is reasonably possible. The list of all observations is at the end of the thesis in \emph{List of Observations} on page \pageref{list:observations}.

\observation{Observations in this thesis can be generalized only to the extent of our experiment settings.}{obs:not_general}

This chapter is organized as follows. First, we specify the terminology used across this thesis. In \cref{sec:domain_adaptation}, we briefly describe the domain adaptation approach, which is a stepping stone for transfer learning. Then in \cref{sec:transfer_learning} we dive into transfer learning, which is categorized into two groups. We investigate one of the groups in \cref{sec:cold_start_transfer} and introduce two techniques discussed separately in \cref{sec:direct_transfer} and \ref{sec:vocabulary_transformation}. The second group is described in \cref{sec:warm_start_transfer}. We wrap up the chapter with a comparison of the proposed techniques in \cref{sec:warm_cold_comparison}, followed by related work in \cref{sec:transfer_relatedwork} and the conclusion in \cref{sec:transfer_conclusion}.

\section{Thesis Terminology}

The main idea behind transfer learning is to pass on learned knowledge from one model to another. We denote the first model from which parameters are transferred as a \emph{parent} and the designated model as a \emph{child}.

In literature, we can often find a naming convention of \emph{teacher} and \emph{student}, however, it is more related to the knowledge distillation \parcite{hinton2014distillation}, where the parent model (the teacher) is used to generate examples instead of directly sharing parameters. Another terminology is \emph{source} and \emph{target} tasks \parcite{torrey2010transfer}, which is unsuitable for this thesis as we use these terms when referring to the source and target language of the language pair.

Notably, in the \ac{NMT} transfer learning it is customary to use the term ``parent-child'' \parcite{zoph2016transferLowResource, nguyen2017transfer}, thus we are going to use it throughout this thesis.

Nonetheless, we use the naming convention of ``parent-child'' in this thesis for all scenarios, where there is a precedent model needed for training the second model regardless of the transferring technique. For example, during the backtranslation (see \cref{sec:backtranslation}), the parent model generates training data for the child model, but no learned parameters are shared between models. 
Thus the child always specifies our designated task or the language pair for which we are trying to get the best performance. In contrast, we are not interested in the performance of the parent.

Both parent and child can use different language pair. For the parent model that translates from language XX to language YY (\transl{XX}{YY}), we recognize three scenarios:

\begin{itemize}
  \item Shared-source language -- a scenario where the source language is equal for parent and child. In other words, the child model translates \transl{XX}{AA}.
  \item Shared-target language -- a scenario where the target language is equal, therefore the child model translates \transl{AA}{YY}.
  \item No-shared language -- a scenario with no language shared, i.e. the child model translates \transl{AA}{BB}.
\end{itemize}

\subsection{Multitask Learning}

\begin{figure}
\caption{The information flows only in one direction from parent to the child task in transfer learning compared to the multi-task learning, where the information flows freely among all tasks improving them altogether.}
\begin{center}
    \def\svgwidth{\columnwidth}
    \scalebox{.85}{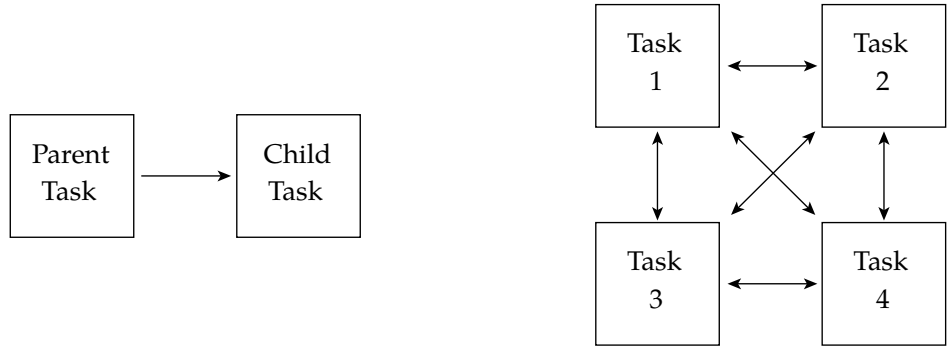}
    \label{fig:information_flow}
\end{center}
\end{figure}

Multitask learning \parcite{caruana1997multitask} is closely related to transfer learning, where the goal is to solve several tasks simultaneously, as illustrated in \cref{fig:information_flow}. 
Multitask learning differs from transfer learning as it needs to keep the performance of all tasks on a high level. In contrast, transfer learning can forget the knowledge of the parent task and only focuses on the performance of the child task. 

It is possible to extend transfer learning to multiple parent tasks and therefore solve multi-task learning with techniques from transfer learning. However, multitask learning is beyond the scope of this thesis, and we are not investigating it.
% Transfer learning uses a parent task as a helping hand during the training, but we are not interested in the performance of the parent task as it
% in contrast to the multitask learning, where the goal is to solve several tasks
% We have to make a difference between transfer learning and multitask learning \parcite{caruana1997multitask}, where the goal is to solve several tasks simultaneously, as illustrated in \cref{fig:information_flow}.
% Although multitask learning is closely related to transfer learning, it needs to keep the performance of all tasks on a high level. In contrast, transfer learning can forget knowledge about the parent task and only focuses on the child task. It is possible to extend a transfer learning to multiple parent tasks and therefore solve multi-task learning with techniques from transfer learning. 

Transfer learning is a broad term used for various approaches not only in deep learning \parcite{pan2010transfersurvey}. In the domain of neural networks, and especially in the \ac{NLP}, we can transfer knowledge either by transferring only a subset of trained parameters or the whole model. 

When only the subset of model parameters are transferred, they are often the embeddings. We discussed transferring of pretrained embeddings in \cref{sec:embedding_initialization}. Transferring all training parameter can be further split into transferring within the same task only by specializing the parent model or by transferring knowledge across different tasks. In \ac{MT}, the former would correspond to the parent using the same language pair as the child, which is called \emph{domain adaptation} described in \cref{sec:domain_adaptation}. The latter case corresponds to different language pairs.

Despite transfer learning being such a broad term, it most often denotes transferring knowledge across different tasks. Thus we are going to use the term of transfer learning only in the scenario when we transfer all model's trainable parameters, and the parent and child use different language pair.

\section{Domain Adaptation}
\label{sec:domain_adaptation}

Domain-specific \ac{MT} systems are in high demand while general \ac{MT} systems have limited applications. The general systems usually perform poorly, and thus it is important to develop \ac{MT} for specific domains \parcite{koehn2017six}.

Domain adaptation is one of the critical issues in \ac{MT}, where the goal is to specialize the model for a more specific domain. It is well known that an optimized model on a specific genre (news, speech, medical, literature, and other) obtains higher accuracy results than a generic system on the given domain \parcite{gao2002improving, hildebrand2005adaptation}. Specifically, when the training data have an unbiased distribution over the target domain, the final model will perform comparably on testing data as it performed during training on the development set. However, the performance will decrease if the training data comes from a different domain than the target domain (see \cref{sec:domain_definition}). For example, when the training data are from news articles and the test domain is more specific as medical. Domain adaptation generally encompasses terminology, domain, and style adaptation.

We often have a large amount of out-of-domain parallel sentences. The challenge of training a domain-specific model is to improve the translation performance in the target domain, given only a small amount of additional in-domain data. 
This can be treated by \emph{fine-tuning} (also called \emph{continued training}) of the generic model with domain-specific data.

Domain adaptation has been successfully used in both statistical and neural \ac{MT} \parcite{gao2002improving, hildebrand2005adaptation, luong2015stanford, pecina2017adaptation}. In-domain gains have been shown with as few as dozens of in-domain training sentences \parcite{barone2017regularization}. 

In a typical \ac{NMT} domain adaptation setup \parcite{luong2015stanford, servan2016domain, chu2018survey}, we first train a parent model on a resource-rich out-of-domain parallel corpus. After the model is trained on the general model, we exchange the training corpus for the in-domain one and follow by fine-tuning the parent model. We can view the domain adaptation as transfer learning from the out-of-domain parent model into the domain-specific child model.

\perscite{freitag2016fast} stated two problems of domain adaptation. First, it is prone to over-fitting due to the limited amount of in-domain data. Second, the process of fine-tuning the parent model on the in-domain data leads to deteriorating the performance on the general domain. 
\perscite{freitag2016fast} propose to tackle the problem by ensembling the parent model with the child model. They showed that the ensembled model improves on the in-domain testset and preserves its performance on the general domain.
\perscite{chu2017empirical} address these problems by mixed fine-tuning, where they combine the out-of-domain corpus with oversampled in-domain data before adapting the general model.

Most researchers investigating domain adaptation assume a scenario where the domain of the data is given, and the in-domain data exists. However, in real-life scenarios such as an online translation engine, the domain of the sentences is not given, and guessing the domains of the input sentences is crucial for proper translation. A method to address the lack of in-domain data can be tackled by classifying the domains of individual sentences in the training data followed by search and selection of training sentences close to the target domain \parcite{farajian2017multi, li2018one}.

This is only one of the approaches for domain adaptation, one which is closest to transfer learning. However, there are many more, especially in different tasks than \ac{MT} \parcite{bruzzone2009domain, pan2010domain}.

\section{Transfer Learning}
\label{sec:transfer_learning}

In the context of \ac{NMT}, transfer learning as well as domain adaptation (see \cref{sec:domain_adaptation}) refer to the situation with a mismatch between the train and test data distribution. In the case of domain adaptation, the parent and child operate on the same language pair but differ in the domain of data. In contrast, transfer learning uses a different parent language pair than the child model. 

Since the domain adaptation works by fine-tuning the parent model trained on the same language pair, it does not require any modification in architecture nor the vocabulary and relies on the continued training process.
In comparison, when we want to retrain the parent with a different language pair than the child, we need to tackle the problem with vocabulary mismatch between parent and child languages. There are two groups of approaches to solving the vocabulary problem based on their application either before training the parent model or after the training. We discuss these two categories in the next section.

\subsection{Transfer Learning Categories}

We distinguish two categories of transfer learning based on whether we have training data for the child language pair at the time of training the parent model. \perscite{neubig2018rapid} call the approaches warm-start and cold-start. In the warm-start approach, we have the child training data available at the time of training the parent, and we can take steps to prepare the parent model for transfer learning. In contrast, the cold-start approaches use a general parent that has not been modified in advance in any way for the child language pair and apply different modifications before training the child model.

Intuitively, we expect the warm-start to perform better because it can better handle the child language pair to some extent. However, the additional training time of the child's specific parent model can be costly. For example, whenever researchers compare various language pairs or hyperparameter setting, they would need to train an individual parent for each such scenario and the time and hardware cost would considerably increase. Therefore, they can consider a scenario to train a strong general \ac{NMT} system only once and use it repetitively in the cold-start transfer learning.
Furthermore, there are situations where the ability to quickly learn the \ac{MT} model given only a small amount of training data can be crucial. For example, when a crisis occurs in a region where people speak an under-resourced language, quick deployment of the \ac{MT} system translating from or to that language can make a massive difference in the impact of the provided support \parcite{lewis2011crisismt}. In such a case, having a strong general \ac{NMT} system available for a quick transfer learning of any low-resource language pair is a great advantage. 

We use the same vocabulary whenever the parent and child have the same set of languages, i.e. disregarding the translation direction and model stage (parent or child). For example, we use the same vocabulary for \transl{English}{Estonian} model as well as \transl{Estonian}{English} although new vocabulary could be generated for each model separately.

Techniques and results in this chapter have been published mainly in two papers. The cold-start scenario is described in the paper \persciteA{kocmi2019ecofriendly}, and warm-start scenario is investigated in the paper \persciteA{kocmi2018trivial}. This chapter includes results as well as short textual parts from those two papers.

\section{Cold-Start Transfer Learning}
\label{sec:cold_start_transfer}

The main problem of transfer learning is the mismatch of parent and child vocabulary. The cold-start transfer learning tackles the problem after the training of the parent by modifying the parent model right before transferring parameters to the child model.

Whenever the parent model uses vocabulary with a high overlap with the child's vocabulary, we can ignore the differences and train the child with the parent's vocabulary. We call this approach ``Direct Transfer'', which we discuss in \cref{sec:direct_transfer}. The second option is to transform the parent vocabulary right before the child's training in various ways to accommodate the needs of the child language pair. Approaches from the second group are discussed in \cref{sec:vocabulary_transformation}.

All cold-start approaches rely on the ability of neural networks to quickly adapt parent parameters to new conditions, i.e. segmenting words usually to more tokens than in parent model or remapping parent subwords embeddings to unrelated child subwords. We show that \ac{NMT} can quickly adapt and obtain better performance on a given child language pair than by training from random initialization.

% The cold-start approach can use any parent model because there is no need to special preparation before parent training. 

The experiments and short textual parts in this section are from our paper \persciteA{kocmi2019ecofriendly} and we use \TFB{}.

\subsection{Cold-Start Evaluation}
\label{sec:cold_start_evaluation}

The cold-start approach can be used with any parent model, even one trained by different researchers.
In order to demonstrate that our cold-start methods are general, not restricted to our laboratory setting and that the results are easily replicable, we do not train the parent model by ourselves, but for all experiments regarding the cold-start transfer learning we reuse the winning model from the \ac{WMT} 2018 \pair{Czech}{English} translation shared task trained by \perscite{popel2018wmt}. The model was trained with CzEng 1.7 and several other smaller corpora. 
 
We decided to use this model for several reasons. It is trained to translate between English and Czech, a high-resource language pair where Czech is not similar to any of the languages we use.\footnote{From our target language selection (see \cref{sec:other_datasets}), the linguistically closest language is Russian, but we do not transliterate Cyrillic into Latin script. Thus the system cannot associate similar Russian and Czech words based on subword appearance.} It is trained using the state-of-the-art Transformer architecture as implemented in the popular Tensor2Tensor framework\footnote{\url{https://github.com/tensorflow/tensor2tensor}} \parcite{tensor2tensor}.

We use two parent models: the \transl{English}{Czech} and \transl{Czech}{English}. Where the parent and child model always use English on the same side. For example, the \transl{English}{Russian} child has \transl{English}{Czech} as the parent. 

To the best of our knowledge, we are the first to use transfer learning on a model trained by someone else, showing the proof-of-concept for recycling models in \ac{NMT}, which reduces the overall training time of \ac{NMT}. Note that reusing neural models is more common in other tasks (see \cref{sec:word_embedding}).

\section{Cold-Start Direct Transfer}
\label{sec:direct_transfer}

Subword-based vocabulary can represent any text from any language by breaking the words down to characters or bytes.\footnote{The standard implementation of \ac{BPE} segmentation by \perscite{sennrich2016bpe} cannot represent unknown characters by breaking them to bytes. However, the implementation could be extended to support encoding of bytes by escaping the byte representation in the same way as in the wordpieces.} We exploit this behavior that a parent vocabulary can encode any child's words and use the parent model as is. We call this approach the ``Direct Transfer''.

In the Direct Transfer approach, we ignore the specifics of the child vocabulary and train the child model using the same vocabulary as the parent model. We take an already trained parent model and use it to initialize parameters for a child language pair. In other words, we continue the training process of the parent model on the child trainset without any change to the vocabulary or hyper-parameters. This applies even to the training meta-parameters, such as the learning rate or moments.

This method of continued training on different data while preserving hyper-parameters is essentially a domain adaptation technique, where the domain is a different language pair.

The intuition behind the Direct Transfer is that \ac{NN} is robust enough that the vocabulary mismatch can be disregarded altogether as long as there is some overlap between the child and parent vocabulary. This is mainly due to the usage of subword tokens, which segment any text into a sequence of allowed subwords (see \cref{sec:subword_representation}). However, Direct Transfer suffers from a deficiency in the segmentation of child words, which can lead to splitting words to individual characters or even bytes that are hard for \ac{NMT} to correctly translate.
%Additionally, the text has to be segmented into a reasonable number of subwords.

Thanks to the simplicity of the Direct transfer, it can be easily applied to existing training procedures as it does not need any modification to the \ac{NN} frameworks. 

In the following sections, we discuss automatically assessed translation quality. We show the results of Direct Transfer in \cref{sec:direct_transfer_results}, followed by an analysis of the usage of parent vocabulary in \cref{sec:parent_vocab_effect} and introduction of a problem with vocabulary mismatch in \cref{sec:vocabulary_overlap}. We conclude by the analysis of Direct Transfer drawbacks in \cref{sec:direct_transform_drawbacks}.

\subsection{Direct Transfer Results}
\label{sec:direct_transfer_results}

%moved
\begin{table}[t]
\caption{``Baseline'' is a model trained from random initialization with own specific vocabulary. ``Direct Transfer'' is using the parent vocabulary. Models in the top part of the table use the \transl{English}{Czech} parent model, models in the bottom part use \transl{Czech}{English}. The scores and difference are in BLEU. The \significantmark{} represents significantly better results based on significance test described in \cref{sec:statistical_significance}.
}
\begin{center}
\begin{tabularx}{\textwidth}{@{}lCCC@{}}
\toprule
Translation direction & Baseline & Direct Transfer & Difference \\
\midrule               
\transl{\EN{}}{Odia}     &  \bf \p{}3.54\significant{} &   \p{}0.04 &  -3.50 \\
\transl{\EN{}}{Estonian} & 16.03 &   \bf 20.75\significant{} &  \ps{-}4.72\\
\transl{\EN{}}{Finnish}  & 14.42 &   \bf 16.12\significant{} &  \ps{-}1.70\\
\transl{\EN{}}{German}   & 36.72 &   \bf 38.58\significant{} & \ps{-}1.86\\
\transl{\EN{}}{Russian}  & \bf 27.81\significant{} & 25.50 &  -2.31\\
\transl{\EN{}}{French}   & 33.72 & \bf 34.41\significant{} &  \ps{-}0.69\\
\transl{French}{Spanish}   & 31.10 & \bf 31.55\significant{} &  \ps{-}0.45\\
\midrule
\transl{Estonian}{\EN{}} & 21.07 & \bf 24.36\significant{} &  \ps{-}3.29\\
\transl{Russian}{\EN{}}  & \bf 30.31\significant{} &  23.41 &  -6.90\\ 
\bottomrule
\end{tabularx}
\end{center}
\label{tab:coldstart_results_direct}
\end{table}

We start with the results of Direct Transfer method, which uses parent vocabulary without any change. The results of our evaluation are tabulated in \cref{tab:coldstart_results_direct}. In comparison to the baseline, the performance of Direct Transfer is significantly better in both translation directions in all cases except for Odia and Russian, which use a different writing script and we discuss it later in \cref{sec:direct_transform_drawbacks}, where we show that it is primarily a problem of filtering long sentences.

Importantly, our baseline, trained only on child data, has an advantage over cold-start transfer learning as it uses child-specific vocabulary. Closer baseline to our transfer learning setup would be to use the parent vocabulary even for baseline, which would lead to an even larger difference in the performance. However, we decided to use the stronger baseline.

The \pair{Estonian}{English} pair confirms that sharing the target language improves performance as previously shown on similar approaches \parcite{zoph2016transferLowResource,nguyen2017transfer}. Moreover, we show that the improvements are significant for the translation direction from English, an area of transfer learning neglected in previous studies.

The largest improvement of 4.72 BLEU is for the low-resource language pair \transl{English}{Estonian}. Furthermore, the improvements are statistically significant even for a high-resource language such as 0.69 BLEU increase for a high-resource \transl{English}{French}. To the best of our knowledge, we are the first to show that transfer learning in \ac{NLP} can be beneficial also for high-resource languages. 

\observation{Direct Transfer can significantly improve the performance of the child model in both directions for both low-resource and high-resource language pairs.}{obs:direct_transfer_helps}

The basic intuition behind the improvements in translation direction into English is that the models reuse the English language model in the decoder and therefore the improvements are due to better ability to generate grammatically correct sentences without the context of the source language. Although better decoder's language model could be one of the reasons behind the improvements, it cannot be the only explanation since we see the improvements also for translation direction where English is the source side, and therefore the decoder has to learn a language model for the second language.

Furthermore, we get improvements even for child language pairs in the no-shared language scenario. In our study, we evaluated \transl{French}{Spanish}, which got a 0.45 BLEU improvement. Although, in this particular case, we need to take into account that it could be partly due to these languages being linguistically closer to the parent's source language English.  

In \cref{sec:shared_decoder_is_easier}, we discuss that the shared-target language is easier for \ac{NMT} than the shared-source language. It is also the main reason why we compare more systems in the direction from English rather than to English. 

The results of such performance boost are even more surprising when we take into account the fact that the model uses the parent vocabulary and thus splits words into considerably more subwords, which we carefully analyze in \cref{sec:vocabulary_overlap}.
It suggests that the Transformer architecture generalizes very well to short subwords and is robust enough to generate longer sentences.

In conclusion, the Direct Transfer learning improves the performance in all cases except \transl{English}{Odia}, \transl{English}{Russian} and \transl{Russian}{English}. In order to shed light on the failure of these languages, we need to analyze the parent vocabulary.

\subsection{Parent Vocabulary Effect}
\label{sec:parent_vocab_effect}

The problem of \ac{OOV} words is solved by using subwords segmentation at the cost of splitting less common words into separate subword units, characters, or even individual bytes, as discussed in \cref{sec:subword_representation}. The segmentation applies deterministic rules on the training corpus to generate the subword segmentation that minimizes the number of splits for the observed word frequencies to fill up the vocabulary of a predefined size (see \cref{sec:subword_representation}).

However, when using a subword segmentation created for a different language pair, the condition of the optimal number of splits is not guaranteed. Especially more linguistically distant languages that contain only a small number of common character n-grams need more splits per word. 

The example in \cref{fig:different_parent_vocab} shows that using a vocabulary for a different language leads to segmenting words to substantially more tokens, especially in the case when the language contains characters not available in the vocabulary. This is most crucial as the unknown character is first transformed into byte representation (``\textbackslash{}237;'' in \cref{fig:different_parent_vocab}) that is later handled as a standard text. 

In \cref{fig:different_parent_vocab}, English vocabulary doubles the number of tokens in the investigated sentence relative to the Czech vocabulary.

\begin{figure}
\caption{A toy example of using English wordpiece segmentation onto Czech sentence. For simplicity, we suppose the vocabularies contain all ASCII characters in addition to the tokens specifically mentioned.
}
\small
% \begin{center}
\begin{tabularx}{\textwidth}{@{}lr@{}}
\\
\toprule
% Czech vocabulary: & \{`bude', 'doma\_`, `end', `me\_', `ví', `vík'\} \\
% English vocabulary: & \{`bud', `dom', `end', 'ho`, `me', `week\_', `will'\} \\
Czech vocabulary: & \{bude, doma\_, end, me\_, v{\'i}, v{\'i}k\} \\
English vocabulary: & \{bud, dom, end, ho, me, week\_, will\} \\
Czech Sentence: & \texttt{O v{\'i}kendu budeme doma.} \\
Segmented by Czech vocab.: &
% \{`O\_',`vík',`end',`u\_',`bude',`me\_',`doma\_',`.\_'\} \\
\texttt{O\_ v{\'i}k end u\_ bude me\_ doma\_ .\_} \\
Segmented by English vocab.: &
% \{`O\_',`v',`\textbackslash\textbackslash',`2',`3',`7',`;',`k',`end',`u\_',`bud',`e',`me\_',`dom',`a\_','.\_'\}
\texttt{O\_ v \textbackslash{} 2 3 7 ; k end u\_ bud e me\_ dom a\_ .\_} \\
\bottomrule
\end{tabularx}
% \end{center}

\label{fig:different_parent_vocab}
\end{figure}

Direct Transfer approach uses the parent vocabulary, which can lead to segmenting the child training set into many individual tokens that could harm the \ac{MT} performance. In order to study this effect, we examine the influence of using different vocabulary on the training dataset of the child.

We consider the parent \pair{Czech}{English} vocabulary \parcite{popel2018wmt} used in our experiments and apply it for segmentation of language pairs and compare the average number of subwords per word. We examine the language pairs and their training sets that are used in experiments regarding Direct Transfer.

\begin{table}[t]
\caption{Average number of tokens per word (tokenized on whitespace) when applied to the training corpora. ``Specific'' represents the vocabulary created specifically for the examined language pair. ``EN-CS'' corresponds to the use of \pair{Czech}{English} vocabulary. The ``First language'' represents English, except of the last row, where it represents French. ``Second language'' represents the other language of a given language pair.}
\begin{center}
\begin{tabularx}{\textwidth}{@{}lHCHCHCHC@{}}
\toprule
      & \multicolumn{4}{c}{First language} & \multicolumn{4}{c}{Second language}  \\
\cmidrule(r){2-5} \cmidrule(l){6-9}

Language pair & \multicolumn{2}{c}{Specific} & \multicolumn{2}{c}{EN-CS}       & \multicolumn{2}{c}{Specific} & \multicolumn{2}{c}{EN-CS}  \\
\midrule               
\pair{\EN{}}{Odia}     & 30.3 & 1.2 & 37.1 & 1.4   & 95.8 & 3.7 & 496.8 & 19.1\\
\pair{\EN{}}{Estonian} & 29.5 & 1.2 & 29.5 & 1.2   & 26.0 & 1.1 &  56.2 &  \p{}2.3\\
\pair{\EN{}}{Finnish}  & 26.2 & 1.2 & 26.6 & 1.2   & 22.9 & 1.1 &  55.9 &  \p{}2.6\\
\pair{\EN{}}{German}   & 26.4 & 1.2 & 27.2 & 1.2   & 27.4 & 1.3 &  55.4 &  \p{}2.5\\
\pair{\EN{}}{Russian}  & 32.2 & 1.3 & 34.1 & 1.4   & 33.3 & 1.3 & 134.9 & \p{}5.3\\
\pair{\EN{}}{French}   & 34.4 & 1.3 & 37.4 & 1.4   & 42.0 & 1.6 &  65.7 &  \p{}2.5\\ 
\pair{French}{Spanish}& 37.3 & 1.3 & 60.8 & 2.1   & 35.1 & 1.2 &  60.9 &  \p{}2.1\\
\bottomrule
\end{tabularx}
\end{center}
\label{tab:different_vocab_tokenized_dataset}
\end{table}

\cref{tab:different_vocab_tokenized_dataset} documents the splitting effect of various vocabularies. When using the language-pair-specific vocabulary (column ``Specific''), the average number of subword tokens per word (denoted ``segmentation rate'') is relatively constant for the English around 1.2 subwords per word as well as other languages except for the Odia language with 3.7 tokens per word, which we explain in \cref{sed:odia_subword_irregularity}. This regularity possibly emerges from the size of vocabulary and the number of words in both languages.

\observation{Using a language-specific wordpiece vocabulary has a consistent segmentation rate around 1.2 subwords per word.}{obs:segmentation_rate_is_consistent}

% \XX{Jindra: tohle mi prijde ze by melo jit odvodit analyticky z vlastnosti toho algoritmu. Tak nevim, jestli je to neco co by se spis melo dokazovat je zahodno uvadet jako empiricky vysledek}

If we use the \pair{Czech}{English} vocabulary on the child dataset (column ``EN-CS''), there is an apparent increase in the average number of subword tokens per word. For example, German has twice as many tokens per word compared to the language-specific vocabulary that has 1.3 tokens per word on average.
Russian has four times more tokens per word due to Cyrillic, similarly for the Odia script.

Russian Cyrillic alphabet happens to be contained in the parent vocabulary together with 59 Cyrillic bigrams and 3 trigrams, which leads to 5.3 tokens per word.
The Odia script is not contained in the \pair{Czech}{English} vocabulary at all, leading to the splitting of each word into individual bytes, which explains the 19.1 tokens per word (see \cref{fig:different_parent_vocab}). 

The first language is not affected by the parent vocabulary much (only slightly for the French-Spanish language pair) because English is shared between both the child and the parent vocabulary. The second language that differs between parent and child approximately doubles the number of splits when using parent vocabulary (see the difference between both columns ``EN-CS'').

\observation{Wordpiece vocabulary roughly doubles the segmentation rate for different child languages that use the same script as parent languages.}{obs:parent_vocabulary_doubles_segmentation_rate}

It is necessary to mention that the datasets have various domains and various sizes and therefore the average number of tokens could be different on various domains even for the same language pair. The size of the vocabulary\footnote{We use vocabulary with 32k subwords in all experiments.} is also crucial as it defines the number of available subwords. Moreover, the length relation between the source and target sentences influence the final vocabulary.

The use of a different segmentation roughly doubles the number of tokens per sentence for languages using the same writing script. Therefore the \ac{NMT} models that use Direct Transfer have to adapt to different sentence length in comparison to the parent. However, as we showed in \cref{sec:direct_transfer_results}, the Direct Transfer significantly improves the performance over the baseline showing the robustness of \ac{NMT}. 

\subsubsection{Odia Subword Irregularity}
\label{sed:odia_subword_irregularity}

We try to shed some light, why Odia has, on average, more tokens per word after subword segmentation even when using a language-specific vocabulary. The Odia script (also called Oriya script) has 52 characters, which lead to more character combinations than in English, which we believe is linked to a higher number of subwords per word.

To confirm our intuition, we investigate the \pair{Odia}{English} vocabulary. The average length of Odia tokens in the vocabulary is 4.2 characters compared to the 6.9 characters for English subwords in the same vocabulary.
The average length of a non-segmented word in the \pair{Odia}{English} training set is 6.4 characters for Odia and 5.2 characters for English. With that in mind, Odia has on average longer words but uses shorter subwords than English, which leads to the substantially higher average number of tokens per word as reported in \cref{tab:different_vocab_tokenized_dataset} in comparison to other languages.

This is mainly due to the size of vocabulary, which is not enough for the \pair{Odia}{English} language pair. Larger vocabulary would contain longer Odia subwords, thus would make the segmentation less frequent. This fact could be one of the reasons why the performance of Direct Transfer is worse than baseline as reported in \cref{sec:direct_transfer_results}. On the other hand, Odia is a low-resource language, and having large vocabulary would result in fewer examples per individual subwords in training data. We investigate the problem in \cref{sec:direct_transform_drawbacks}.

\subsection{Vocabulary Overlap}
\label{sec:vocabulary_overlap}

Direct Transfer uses parent vocabulary, and we showed how it increases the segmentation of the child's training corpus in \cref{sec:parent_vocab_effect}. Now we examine what percentage of the parent vocabulary is used by the child language pair and investigate how large is the part of parent vocabulary that is left unused with the child language pair.

In order to find tokens from the parent vocabulary that are used by the child model, we segment the training corpus of the investigated languages with the \pair{Czech}{English} vocabulary and count how many unique tokens from the vocabulary appear in the segmented child's training corpus.

\begin{table}[t]
\caption{The percentage of the parent vocabulary tokens used in the child's trainset. The vocabulary is shared for both languages. The column ``Both'' represents the number of vocabulary tokens used by both languages.}
\begin{center}
\begin{tabularx}{\textwidth}{@{}l@{\hskip 2cm}CC@{\hskip 1cm}C@{}}
\toprule
Language pair & 1st language & 2nd language & Both  \\
\midrule               
\pair{\EN{}}{Czech}     & 71.1\% & 98.0\% & 98.8\% \\
\midrule
\pair{\EN{}}{Odia}       & 23.3\% & \p{}0.8\% & 23.9\% \\
\pair{\EN{}}{Estonian}   & 54.8\% & 39.7\% & 57.0\% \\
\pair{\EN{}}{Finnish}    & 59.2\% & 54.5\% & 60.6\% \\
\pair{\EN{}}{German}     & 58.4\% & 52.9\% & 59.9\% \\
\pair{\EN{}}{Russian}    & 68.6\% & 67.0\% & 71.4\% \\
\pair{\EN{}}{French}     & 64.6\% & 64.5\% & 65.5\% \\
\pair{French}{Spanish} & 55.1\% & 54.0\% & 57.1\% \\
\bottomrule
\end{tabularx}
\end{center}
\label{tab:parent_vocab_usage_ratio}
\end{table}

The percentage of used tokens are in \cref{tab:parent_vocab_usage_ratio}. Before examining the results, we need to mention that the training sets are usually noisy, and some sentences from other languages can easily appear in them. For example, it is often a case that part or whole sentence is left untranslated in the parallel corpus. For the simplicity of our argument, we do not remove any foreign sentences nor ignore the least used tokens in any way. Therefore, the actual used part of parent vocabulary by a given language is smaller than presented in \cref{tab:parent_vocab_usage_ratio}.

With that in mind, we can notice that English always uses more tokens from the vocabulary than the second language. This is caused by English being the shared language and already present in the vocabulary. Although the reverse is true for the original dataset of \pair{Czech}{English} where English occupies a smaller part of the vocabulary (71.1\%) than Czech (98.0\%). The reason why the total is not 100\% but only 98.8\% (see column ``Both'') is possibly due to a slightly different training set as the vocabulary was prepared by \perscite{popel2018wmt}.

We can see that the Odia does not use many of available subwords, as concluded in the previous section. Interestingly and contrary to our previous findings, Russian utilizes a substantial part of the parent vocabulary. After a closer examination of the training corpus, we noticed that the Russian part contains many Czech and English sentences. When we counted only subword tokens that contain at least one Cyrillic character, the used part of vocabulary dropped to 0.3\% for Russian confirming the previous findings with extremely high segmentation rate than other languages. 

The most important result is that most language pairs use around 60\% of parent vocabulary. This means that remaining tokens are left unused and only slow down the training and inference because the model has to calculate the softmax over the size of the vocabulary. Thus in \cref{sec:direct_transform_drawbacks}, we propose a vocabulary transformation approach that overrides these unused tokens with child-specific ones.

\observation{Child language pair uses roughly 60\% of the available parent vocabulary tokens when sharing one language with the parent in Direct Transfer.}{obs:child_uses_only_sixty_percent_of_parent_vocab}

\subsection{Direct Transfer Drawbacks}
\label{sec:direct_transform_drawbacks}

%\XX{jindra: na to ze je to ve skutecnosti bug tak to podle me moc rozpytvavas}

In this section, we discuss the failure to improve the performance of child models of \transl{English}{Odia}, \transl{English}{Russian} and \transl{Russian}{English} translation pairs. Intuitively, Russian and Odia use a different writing script than English, and therefore the parent vocabulary segments them into many wordpieces as we discussed in \cref{sec:vocabulary_overlap}, which possibly harms the \ac{NMT} and fails to adapt to such long sentences fully. However, we need to take into account another possible explanation. 

The framework \ac{T2T} drops sentences from training corpus that are too long in order to allow bigger batch sizes, which lowers the training time. In our experiments, the threshold is set to 100 subwords, which is based on our previous findings \parciteA{kocmi2018wmt}, where it was enough for most of the languages. \ac{NMT} translates individual sentences with an average length of 20-30 words. Thus the average segmentation of 1.2 leads to an average length of a sentence under 50 tokens. However, the limit was set for a language-pair-specific segmentation, and thus, we investigate the influence of a higher segmentation rate.

We segmented the training data with both \pair{Czech}{English} and language-specific vocabulary and compared the percentage of filtered sentences by counting sentences with less than 100 tokens. The percentage of training corpus that has been filtered out is presented in \cref{tab:filltering_problems}.

\begin{table}[t]
\caption{The amount of training corpus removed by filtering long sentences with more than 100 subwords (lower is better). The column ``\pair{Czech}{English}'' shows results when parent segmentation is used. The ``Child-specific'' exploits vocabulary prepared for each language pair separately, the child-specific vocabulary has been used by the baseline systems.}
\begin{center}
\begin{tabularx}{\textwidth}{@{}l@{\hskip 3cm}CC@{}}
\toprule
Language pair & \pair{Czech}{English} (\%) & Child-specific (\%) \\
\midrule               
\pair{\EN{}}{Odia}       & 98.6  & 12.0 \\
\pair{\EN{}}{Estonian}   & \p{}8.7  & \p{}0.0\\
\pair{\EN{}}{Finnish}    & 12.6  & \p{}0.2\\
\pair{\EN{}}{German}     & \p{}4.0  & \p{}0.3\\
\pair{\EN{}}{Russian}    & 58.1  & \p{}1.7\\
\pair{\EN{}}{French}     & 10.8  & \p{}1.0 \\
\pair{French}{Spanish} & 13.9  & \p{}0.1 \\
\bottomrule
\end{tabularx}
\label{tab:filltering_problems}
\end{center}
\end{table}

As we have feared, the filtration has removed 98.6\% of sentences from the \pair{Odia}{English} corpus and 58.1\% from the \pair{Russian}{English} corpus. Additionally, other language pairs also have a notable drop in the total amount of training data. We note that this happens silently on the fly in the \ac{T2T} framework.

\observation{Corpus filtration based on the length of the sentences can drop a large part of training corpora due to the higher segmentation rate when using a parent vocabulary.}{obs:filtration_drops_lot_of_sentences_due_parent_vocab}

On the defense of the limit, when using the child-specific vocabulary, the filtering does not remove many sentence pairs, and therefore, it is justifiable. However, the limit does not take into consideration the usage of a not-optimized vocabulary.

These findings mean that the Direct Transfer results have been negatively affected in comparison to the baseline, and therefore, the actual result could have been even better. Unfortunately, we had made this analysis after running all the experiments. Since the Direct Transfer had improved the results over the baseline for most cases, we decided to avoid rerunning all the experiments with a different limit. We re-run only the experiments with a higher threshold for filtering long sentences only for the three systems that scored worse than the baseline and used a different script than the parent. 

We increased the threshold from 100 to 500 tokens per sentence. This filtering is already high enough to maintain most of the training corpus. Specifically, it removes only 3.6\% sentences from the \pair{Odia}{English} and 0.8\% from \pair{Russian}{English} corpus. Increasing the number of tokens per sentence increases the nodes in the computation graph. In order to fit the same model on GPU, we run these experiments on our 16GB GPUs Quadro P5000. Results are in \cref{tab:coldstart_results_reruns}.

\begin{table}[t]
\caption{Direct Transfer performance with increasing the filtering limit. The best performance (BLEU) is in bold. The \significantmark{} represents significantly better results when compared to baseline.}
\begin{center}
\begin{tabularx}{\textwidth}{@{}l@{\hskip 1cm}CCC@{}}
\toprule
Language pair & Baseline & Filter 100 & Filter 500 \\
\midrule               
\transl{\EN{}}{Odia}     & \bf \p{}3.54\significant{} &  \p{}0.04 & \p{}0.26 \\
\transl{\EN{}}{Russian}  & \bf 27.81\significant{} & 25.50 & 27.19 \\
\transl{Russian}{\EN{}}  & 30.31 & 23.41 & \bf 30.49\\
\bottomrule
\end{tabularx}
\label{tab:coldstart_results_reruns}
\end{center}
\end{table}

% \XX{Jindra: jako reviever bych po tobe chtel experiment, kde by se pridala transliterace do cestiny, coz je rychly postup pro obe strany. coz by mel tvuj transfer vylepsit}

We see that with a higher filtering limit, the performance indeed improves for \pair{Russian}{English}, but not for \transl{English}{Odia}. However, the baseline is still outperforming the Direct Transfer in \transl{English}{Odia} and \transl{Russian}{English}. It is an indication that the model cannot adapt to sentences segmented to individual bytes. However, it can adapt to short n-grams of characters as in Russian.

Interestingly, \transl{Russian}{English} performs on par with the baseline, neither of the systems being significantly better than the other. This finding suggests that the model can adapt in the situation when the source language is segmented, but the target language is not. In this case, the target language is not affected by a high segmentation rate. Intuitively, it suggests that the shared-target scenario is easier for transfer learning. In other words, the neural network can utilize the language model of the parent and adapt it. We analyze it in more depth in \cref{sec:shared_decoder_is_easier}.

\observation{Direct Transfer does not improve performance when a child uses language with a different writing script.}{obs:direct_transfer_hurt_when_child_use_different_script}

In conclusion, Direct Transfer improves the performance over the baseline as long as the segmentation due to vocabulary mismatch is reasonably low. An interesting question is if there is a threshold for segmentation ratio, which would predict if the model will improve over the baseline or if the Direct Transfer fails, or if the explanation is because of sentence length mismatch between source and target sentences. We leave these questions for future work.

In the following section, we discuss a vocabulary transformation technique, which avoids the segmentation problem and further improves the performance of cold-start transfer learning.

\section{Cold-Start Vocabulary Transformation}
\label{sec:vocabulary_transformation}

Direct Transfer relies on the fact that we use subword units that can be used to encode any textual string. The pre-processing splits unseen words into several subwords, characters or possibly down to individual bytes. This feature ensures that the parent vocabulary can, in principle, serve for any child language pair, but it can be highly suboptimal, segmenting child words into too many subwords, where individual subword units do not contain relevant meaning.
As expected, this is most noticeable for languages using different scripts (see the statistics in \cref{tab:different_vocab_tokenized_dataset}).
Also, the child does not use the whole vocabulary leaving around 40\% unused, which only slows down the training and inference process. In order to avoid both problems, the unused token positions could be reused to represent subwords more suitable for the need of the child language pair.

In this section, we propose a vocabulary transformation approach that changes unused subwords in parent's vocabulary with child-specific ones. We show that a straightforward replacement of subwords leads to significant improvements in performance.

\ac{NMT} models associate each vocabulary item with its embedding. When transferring from the parent to the child, we can remap subwords and their assigned embedding as trained in the parent model without any modification to the architecture.
The remapped subwords become associated with embeddings that initially behave as trained for the original subwords and the \ac{NMT} has first to retrain them.

The algorithm for vocabulary transformation is explained in \cref{alg:vocab_match}.

% Our vocabulary transformation starts by constructing the optimal subword vocabulary for the child language pair with a size smaller than or equal to the parent vocabulary size.
% Then we map all child's subwords not represented in the parent vocabulary to parent's subwords not present in a child's vocabulary. To put it another way, we reorder the child's vocabulary by changing indices of identical subwords to match the position in parent vocabulary and others assigned arbitrarily, see \cref{alg:vocab_match}.

\setlength{\algoheightrule}{0pt}
\setlength{\algotitleheightrule}{0pt}

\begin{algorithm}[t]
\caption{Transforming parent vocabulary to contain child subwords and match positions for subwords common for both of language pairs.}
\begin{center}
\framebox{
\parbox{.9\columnwidth}{
 \SetKwInput{KwData}{Input}
 \KwData{Parent vocabulary (an ordered list of parent subwords) and the training corpus for the child language pair.}
 Generate custom child vocabulary with the maximum number of subwords equal to the parent vocabulary size\;
 \For{subword S in parent vocabulary}{
  \eIf{S in child vocabulary}{
  continue\;
  }{
  Replace position of S in the parent vocabulary with the first unused child subword not contained in the parent\;
  }
 }
 \KwResult{Transformed parent vocabulary}
}}
 \vspace{1em}
 \label{alg:vocab_match}
\end{center}
\end{algorithm}

\subsection{Results with Transformed Vocabulary}

Direct Transfer significantly outperforms the baseline, trained only on the child data, whenever the parent vocabulary does not segment the child training sentences into many tokens. In this section, we evaluate our Transformed Vocabulary approaches, which remaps unused parent subwords to useful child-specific ones.

\begin{table}[t]
\caption{``Transformed Vocab'' has the same setting as Direct Transfer but merges the parent and child vocabulary as described in \cref{sec:vocabulary_transformation}. The structure is the same as in \cref{tab:coldstart_results_direct}. The baseline uses child-specific vocabulary. The statistical significance \significantmark{} is measured between Direct Transfer and Transformed Vocabulary.
}
\begin{center}
\begin{tabularx}{\textwidth}{@{}lCCc@{}}
\toprule
Model & Baseline & Direct Transfer & Transformed Vocab\\
\midrule               
\transl{\EN{}}{Odia}       &  \p{}3.54 &  \p{}0.04 & \bf \p{}6.38\significant{} \\
\transl{\EN{}}{Estonian}   & 16.03 & \bf 20.75 & 20.27 \\
\transl{\EN{}}{Finnish}    & 14.42 & 16.12  & \bf 16.73\significant{} \\
\transl{\EN{}}{German}     & 36.72 & 38.58 & \bf 39.28\significant{} \\
\transl{\EN{}}{Russian}    & 27.81 & 25.50 & \bf 28.65\significant{}\\
\transl{\EN{}}{French}     & 33.72 & 34.41 & \bf 34.46  \\
\transl{French}{Spanish} & 31.10 & 31.55 & \bf 31.67  \\
\midrule
\transl{Estonian}{\EN{}}   & 21.07 & 24.36 & \bf 24.64\\
\transl{Russian}{\EN{}}    & 30.31 & 23.41 & \bf 31.38\significant{}\\ 
\bottomrule
\end{tabularx}
\end{center}
\label{tab:coldstart_results_bleu}
\end{table}

First two columns of \cref{tab:coldstart_results_bleu} are the same as in \cref{tab:coldstart_results_direct}. We added a column with results of Transformed Vocabulary.
We see ``Transformed Vocabulary'' delivering the best performance for all language pairs except \transl{English}{Estonian}, significantly improving over ``Direct Transfer'' in most cases. 
The only exceptions are \transl{Estonian}{English}, \transl{English}{French} and \transl{French}{Spanish}, where neither of the systems is significantly better than the other, however, both of them are significantly better than the baseline.

Furthermore, it confirms that our Transformed Vocabulary successfully tackles the problem of Direct Transfer when the child language uses a different writing script such as \transl{English}{Odia}, \transl{English}{Russian}, and \transl{Russian}{English}.

\observation{Transformed Vocabulary is not negatively affected by parent vocabulary segmentation.}{obs:transformed_vocab_can_do_different_scripts}

We see that cold-start transfer learning is not restricted to the low-resource scenario as it also improves high-resource language pairs: \pair{Finnish}{English}, \pair{German}{English}, \pair{Russian}{English}, and \pair{French}{English}.

\observation{Our cold-start transfer learning improves the performance of both low-resource and high-resource language pairs.}{obs:coldstart_improves_highresource}

Interestingly, the cold-start transfer learning technique improves even the scenario with no-shared language, in this case \transl{French}{Spanish}.
%However, we cannot conclude the general applicability of the approach as it has been shown only on one language pair. 

Furthermore, the positive results imply that the neural network is robust enough to correct the randomly assigned child-specific embeddings and therefore reuse even more capacity of the parent model.

\subsection{Various Vocabulary Transformations}
\label{sec:various_vocabulary_transformations}

Our Transformed Vocabulary technique assigns unmatched subwords mostly at random. However, there are many other variants. We propose several of them and evaluate them in this section.

We noticed that the vocabulary is structured in stages roughly based on the frequency of subwords in the corpora. This is due to the vocabulary creation that adds less frequent subwords in stages until reaching the requested size of the vocabulary. Therefore, we call the technique from the previous section ``Frequency-based''.

In contrast to frequency, we can assign tokens at random. Either all of them or only unmatching tokens. We call the former approach ``Everything random''. It is when all subword tokens are first shuffled and then assigned at random. This approach does not match any tokens. Therefore the \ac{NMT} needs to learn even tokens that have been used by the parent model. The latter approach is called ``Unmatched random''. It first assigns subwords that are in parent and child vocabulary. Then it assigns the remaining child tokens at random.

Last option is the assignment of subwords based on some distance. We select the Levenshtein distance, which measures the number of edits between two strings. The vocabulary created by this technique assigns subwords iteratively by increasing the allowed distance for assignment. In other words, it starts by assigning all matching subwords (distance 0), then subwords that have a distance of one edit, then two edits and so on.

\begin{table}[t]
\caption{Comparison of various approaches to replacing tokens in parent vocabulary.}
\begin{center}
\begin{tabularx}{\textwidth}{@{}l@{\hskip 2cm}CCC@{}}
\toprule
Language pair & \transl{\EN{}}{Estonian} & \transl{Estonian}{\EN{}} \\
\midrule
Frequency-based      & 20.27 & 23.32 \\
Everything random    & 16.41 & 19.84\\
Unmatched random     & \bf 20.28 & 22.45\\
Levenshtein distance & 20.04 & \bf 23.66\\
\bottomrule
\end{tabularx}
\end{center}
\label{tab:transformed_vocabulary_variants}
\end{table}

The results of comparing various approaches to replacing tokens in parent vocabulary are presented in \cref{tab:transformed_vocabulary_variants}. All approaches reach comparable performance except the "Everything random" assignment. Thus we found no significant differences in the performance as long as subwords common to both the parent and child keep their embeddings, i.e. are mapped to the same index in the vocabulary. The subwords unique to the child vocabulary can be assigned randomly to the unused parent embeddings.
A similar result was observed by \perscite{zoph2016transferLowResource}. They show that the random assignment of words in their approach works as well as an assigning based on lexical similarity.

\observation{It is necessary to preserve tokens shared between parent and child in the same place. The order of assignment for the remaining tokens does not play an important role.}{obs:random_is_bad}

There is still plenty of space for experiments with more advanced techniques of vocabulary and embedding mapping, e.g. utilizing multilingual embeddings like Multivec \parcite{berard2016multivec}. We leave this for the future work in order to keep our approach as straightforward as possible and focus more on the analysis.

\subsection{Training Time}

Lastly, we also evaluate the total training time needed for cold-start transferred models to reach the best performance. The times in \cref{tab:coldstart_time_results} represent the number of steps needed to reach the best performing models from \cref{tab:coldstart_results_bleu}. The steps are comparable due to the equal batch size. However, due to the training fluctuations in performance, it is not possible to define the exact step when the model converged. Therefore, the results should be seen only as a rough estimate of the training time. We use the stopping criterion as defined in \cref{sec:stopping_criterion}

\begin{table}[t]
\caption{
The number of steps needed for a model to converge. We present the step where the model has the highest performance on the development step based on the stopping criterion described in \cref{sec:stopping_criterion}.
}
\begin{center}
\begin{tabularx}{\textwidth}{@{}lCCc@{}}
\toprule
Language pair & Baseline & Direct Transfer & Transformed vocab\\
\midrule               
\transl{\EN{}}{Odia}     &  \pp{}45k &   \p{}47k & \bf \p{}38k \\
\transl{\EN{}}{Estonian} &  \pp{}95k & \bf \p{}75k & \bf \p{}75k \\
\transl{\EN{}}{Finnish}  & \p{}420k &  \bf 255k  & 270k \\
\transl{\EN{}}{German}   & \p{}270k & 190k  & \bf 110k \\
\transl{\EN{}}{Russian}  & 1090k & 630k  & \bf 450k \\
\transl{\EN{}}{French}   & \p{}820k & \bf 660k  & 720k \\
\transl{French}{Spanish}   & \p{}390k &  435k & \bf 375k \\
\midrule
\transl{Estonian}{\EN{}} & \pp{}70k & \bf \p{}30k   & \p{}60k \\
\transl{Russian}{\EN{}}   & \p{}980k &  \bf 420k   & 700k \\ 
\bottomrule
\end{tabularx}
\end{center}
\label{tab:coldstart_time_results}
\end{table}

Both transfer learning approaches converged in a comparable or lower number of steps than the baseline as we see in \cref{tab:coldstart_time_results}. The reduction in the number of steps is most visible in \transl{English}{German} and \transl{English}{Russian}, where we got to less than half of the total number of steps.

\observation{Cold-start transfer learning converges faster than training from random initialization.}{obs:coldstart_converges_faster}

As mentioned earlier, the results are only approximate. However, based on the broad range of our experiments where we compared nine language pairs using various scripts and various training corpora sizes we conclude that both approaches of transfer learning are faster than training the model from random initialization.

\subsection{Conclusion on Cold-Start Transfer Learning}

The common practice in \ac{NMT} is to train models from random initialization, which makes \ac{NMT} demanding in terms of training time and hardware resources. Especially in the deployment where multiple language pairs are used, or we do not have resources to train all language pairs.

We proposed and studied two cold-start methods of transfer learning, namely Direct Transfer and Vocabulary Transformation, which improve the child model language pair regardless of the original parent training languages. 

We achieve better translation quality and shorter convergence times than when training from random initialization. We showed the results on several language pairs as well as both translation directions. Furthermore, we showed the improvements even for high-resource language pairs such as \transl{English}{French}.
The improvements are significant even in the scenario with no-shared language, in our case, the \transl{French}{Spanish} pair.

Above all, we showed a proof-of-concept of reusing models trained by others, thus training the parent model is not necessary, and anyone can use a model trained by others for transfer learning.
The usage of models trained by \perscite{popel2018wmt} proves that we could not manipulate the parent model in any way for transfer learning, for example, by hyperparameter search or using modified parent training data. 

We also showed the robustness of neural networks that works despite the adverse conditions of randomly assigned child-specific subwords to embeddings previously trained for parent language pair. \ac{NMT} have the ability to quickly retrain pretrained embeddings and obtain even better performance compared to the Direct Transfer.

The transfer learning technique, as presented in this chapter, is not suitable for scenarios with various architectures like various sizes of matrices or different network layouts. Likewise, the size of the parent vocabulary restricts the size of a child's vocabulary. As we saw in the example with \pair{Odia}{English}, using a language-specific vocabulary of 32k subwords leads to a high segmentation rate of 3.7 (see \cref{tab:different_vocab_tokenized_dataset}), and therefore a vocabulary with more subwords would be necessary to lower the segmentation rate. Although in this case of low-resource language, it would create other problems as it is better to use smaller vocabulary when handling low-resource languages \parcite{sennrich2019revisiting}.

To conclude, the cold-start transfer learning does not need complicated modifications to the framework, only the ability to continue training. It improves performance and shortens the training time. Thus we suggest using transfer learning, especially the Transformed Vocabulary technique, as a new standard for model parameter initialization in scenarios where the architecture has not been changed, e.g. when not experimenting with various sizes of matrix dimensions.

In the following section, we investigate warm-start transfer learning, i.e. an approach that prepares the parent model with the knowledge of the child language pair in advance. In our case, we create a parent vocabulary already prepared for a specific child.

\section{Warm-Start Transfer Learning}
\label{sec:warm_start_transfer}

In the previous chapter, we discussed the cold-start scenario, where the parent model is trained in advance without prior knowledge about the child's language pair. The main disadvantage of that method is the need to reuse the parent vocabulary, which is associated with trained embeddings. The Direct Transfer, therefore, has a problem with a high segmentation rate and leaving roughly 40\% of vocabulary unused. The Vocabulary transformation overcame these problems by randomly assigning child's subwords to unused embeddings, but this method restricted the maximal size of child vocabulary to be equal to the parent's vocabulary. Furthermore, the randomly assigned child subwords embeddings first had to be retrained. The warm-start scenario allows us to overcome such problems by preparing the parent model in advance for the upcoming transfer learning to the child's language pair.

The basic idea of how to avoid problems with randomly assigned embeddings is to directly use child-specific vocabulary at the time of parent model training. The use of a child's vocabulary in the parent model would inevitably undermine the performance of the parent model because we restrict the parent vocabulary. We studied this effect on child models in \cref{sec:parent_vocab_effect}, where we showed that inappropriate vocabulary segments text to an increased number of tokens. However, in the transfer learning task, we do not pay attention to the final performance of the parent language pair. Thus we can ignore this performance drop.

The experiments in this section are mostly from \persciteA{kocmi2018trivial} and use \TFA{} if not specified otherwise.

\subsection{Warm-Start Methods}
\label{sec:warmstart_methods}

In addition to using the child-specific vocabulary, we propose other variants of shared vocabulary in order to find the best approach for transfer learning. So far, we described two options: either to use parent-specific vocabulary (Direct Transfer) or child-specific vocabulary. Each of them has similar problems, but either for the parent or the child model. A solution is to create a vocabulary that is shared between both parent and child language pairs.

\begin{figure}[t]
\caption{A process of generation of shared vocabularies: Merged (left) and Balanced (right) Vocabulary.}
\begin{center}
\def\svgwidth{\columnwidth}
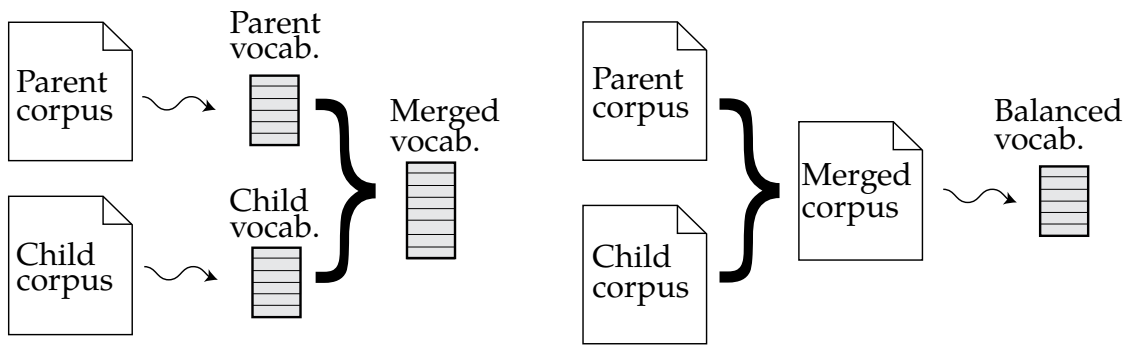
\end{center}
\label{fig:warmstart_vocabulary}
\end{figure}

We propose two approaches of creating a shared vocabulary. One option of a shared vocabulary containing both the parent as well as the child subwords can be created by merging their specific vocabularies. We call this approach Merged Vocabulary. The second option is to utilize the algorithm for subword generation (wordpiece or \ac{BPE}) by applying it to the concatenated corpus of both parent's as well as the child's vocabulary. This way, it prepares a balanced vocabulary based on the subword frequency in both corpora. We call the second approach to Balanced Vocabulary. Both approaches are illustrated in \cref{fig:warmstart_vocabulary}.

Merged Vocabulary is related to the cold-start Transformed Vocabulary as it tries to combine two separate vocabularies into one. The process of vocabulary generation in Merged Vocabulary is as follows: first, we generate vocabulary for both parent and the child language pair separately, and then we merge both vocabularies and remove duplicated subwords. Because the vocabularies are text files where each subword is on an individual line, the merge operation is a simple concatenation of both vocabularies together.
Therefore this approach applies only to wordpiece-type vocabulary, which is not dependent on the ordering of the vocabulary such as the \ac{BPE} algorithm.

The second approach, Balanced Vocabulary, is obtained by concatenating both parent's and child's training corpora into one corpus, which is then used to generate the vocabulary by the wordpiece (possibly also \ac{BPE}) algorithm. Note that this corpus is used only for the vocabulary generation purposes and we are not using it for training \ac{NMT} models.

The Balanced Vocabulary method is sensitive to the total size of the corpus. Whenever one of the corpora is bigger, more subwords of that language pair get into the final vocabulary. Mainly due to our focus on low-resource language pairs, we try to balance the amount of language-specific subwords in the vocabulary by randomly selecting an equal number of sentences from both corpora instead of using all sentences. Thus the mixed corpus contains 25\% sentences from each of the four languages. We need to note that we do not take into account the number of words per sentence as it differs across corpora and also languages. Thus we suppose that corpora are usually already segmented on a sentence level and we disregard the average length of the sentences. We leave for future work an investigation of different ratios between languages and balancing the number of words instead of sentences across corpora.

Whenever we obtain the vocabulary by either of the methods, we follow the transfer learning pipeline that differs from cold-start by the fact that we need to train the parent model for each child. The training is the following: we train the parent model with a given vocabulary until convergence, followed by continued training on the child dataset. We remind the reader that in the case of warm-start transfer learning, we do not change the vocabulary nor any other (hyper)parameters when we exchange the training corpora. Therefore the \ac{NN} is not forced to completely retrain embeddings when we change the training corpora from parent to child. It avoids the problem with the cold-start Transformed Vocabulary approach. Although some subwords have a different meaning in different languages, hence they need to be retrained even in the warm-start scenario.

\subsection{Comparison of Warm-Start Techniques}
\label{sec:warmstart_comparison}

In this section, we compare the two warm-start techniques proposed earlier with parent-specific and child-specific vocabularies to figure out which one leads to the best performance.

We compare the proposed methods using both pairs with both directions (i.e. two low-resource language pairs, namely \pair{Estonian}{English} and \pair{Basque}{English}). In this comparison, we use the parent model that has English on the same side. Thus \transl{English}{Czech} parent is used for \transl{English}{Estonian} and \transl{Czech}{English} parent is used for \transl{Basque}{English}.

We compare four different approaches: parent-specific (Direct Transfer), child-specific, Merged Vocabulary, and Balanced Vocabulary. All setups use the same layout of transfer learning and training conditions. We trained four parent models for \transl{English}{Czech} and four parent models for \transl{Czech}{English}, each with a different vocabulary. The parent training took one million steps before the transfer learning of the child. 

The setups for \transl{Basque}{English} are from \persciteA{kocmi2018iwslt}. We extended the evaluation with \transl{English}{Estonian} translation direction, for which we downsampled the original \pair{Estonian}{English} (see \cref{sec:other_datasets}) corpus to only 100k sentence pairs. Hence, we artificially created a less resourceful language pair. The \transl{English}{Estonian} is based on different version of \ac{T2T} framework than the rest of the section, namely \TFC{}.

Merged Vocabulary approach generates a larger vocabulary because merging two vocabularies of equal size leads to a bigger size of the merged vocabulary. Moreover, larger vocabulary leads to less comparable results as improvements could be attributed to different vocabulary size. We want to evaluate all methods in the closest setting as possible. Thus for Merged Vocabulary, we generate smaller parent and child-specific vocabularies in the first place in a way that the final size after merging and removing duplicates is approximately the same as vocabulary size of other methods, in our experiments 32k subword vocabulary. 
Consequently, we define the size of initial two vocabularies experimentally by iteratively decreasing their size and measuring the size of merged vocabulary until we obtained the size of merged vocabulary within a 1\% tolerance of the vocabulary size (the 32k subwords). The tolerance is in accordance with the \ac{T2T} specific wordpiece implementation, see \cref{sec:wordpiece_description}.

\begin{table}[t]
\caption{Results of various warm-start transfer learning approaches. The first two columns show the performance of the parent model, the third and forth column is the child model based on the corresponding parent in the same row. The baseline does not use transfer learning and uses language-pair-specific vocabulary. Scores are in BLEU and are comparable only within columns.}
\begin{center}
\begin{tabularx}{\textwidth}{@{}lccCC@{}}
\toprule
& \multicolumn{2}{c}{Parent score} & \multicolumn{2}{c}{Child score} \\
\cmidrule(r){2-3} \cmidrule(l){4-5}
Method  & \transl{\EN{}}{CS} & \transl{CS}{\EN{}} & \transl{\EN{}}{Estonian}  & \transl{Basque}{\EN{}}  \\
\midrule
Direct Transfer               & 22.78 & 27.81 & 15.55 & 23.29 \\
Child-specific voc.              & 21.05 & 24.93 & 15.73 & 22.92 \\
Balanced Vocabulary           & 22.58 & \bf 27.93 & \bf 16.41 & \bf 23.63 \\
Merged Vocabulary             & \bf 22.68 &    \ding{54} & 16.05 &    \ding{54} \\
Baseline & -- & -- & \p{}8.70 & 19.09 \\ % baseline je Direct Transfer
\bottomrule
\end{tabularx}
\end{center}
\label{tab:warmstart_transfer_approaches}
\end{table}

We start by investigation of the child model performance (column Child score) in \cref{tab:warmstart_transfer_approaches}. All four transfer learning techniques outperform the baseline trained only on the child language pair, the gains of nearly 8 BLEU points for \transl{English}{Estonian} and 4 BLEU points improvement for \transl{Basque}{English} are significant. 

\observation{Both Balanced and Merged Vocabulary warm-start techniques significantly outperform the baseline.}{obs:balanced_merged_outperformn_baseline}

The missing result of Merged Vocabulary in \transl{Basque}{English} is because we have not conducted a comparison of this technique in \persciteA{kocmi2018iwslt}.

Both Direct Transfer, or child-specific vocabulary setups, performed worse than the Merged and Balanced Vocabulary. We assumed that the parent-specific vocabulary would perform the worst since it introduces segmentation problems into the child language pair, as discussed in \cref{sec:parent_vocab_effect}. Interestingly, the child-specific vocabulary is not the best approach even though the vocabulary is specifically tailored to the child model. Furthermore, the parent score is significantly lower (1.53--1.73 BLEU) for the child-specific vocabulary when compared to other variants, which is due to a sub-optimal vocabulary. This suggests that the translation quality of the parent model plays an essential role in the transfer model. 
This is an exciting result, as we usually disregard the performance of the parent in the transfer learning setup. We study the parent performance and behavior in \cref{sec:parent_performance_dropping}.

\observation{Child-specific transfer learning performs worse than Balanced or Merged Vocabulary even though its vocabulary contains more child-specific subwords.}{obs:child_specific_perform_worse_that_balanced}

Lastly, we want to briefly discuss the ratio of parent vs. child subwords in the variants of the vocabulary. The parent-specific vocabulary (Direct Transfer) in general contains no child language pair subwords. On the other hand, child-specific vocabulary contains only child-specific subwords. In between are both Merged and Balanced Vocabulary that contain roughly half of the parent's subwords and half of the child's subwords. However, it could be the case that the best approach would be different, for example, including only 30\% of parent vocabulary and 70\% of child vocabulary. If that would be a case, it is most likely going to be specific for the parent and child language pairs and not being general for transfer learning. We leave this as an open question for future work.

In conclusion, both Merged and Balanced Vocabulary are better than baseline and the cold-start Direct Transfer. Furthermore, Balanced Vocabulary leads to better quality than Merged Vocabulary. However, we are not claiming that Balanced Vocabulary is strictly better than Merged Vocabulary. Lastly, the Merged Vocabulary technique is less general as it works only for the wordpiece segmentation (see \cref{sec:warmstart_methods}).
Therefore, in the following experiments and analysis regarding the warm-start transfer learning, we are going to use only the Balanced Vocabulary technique as the standard technique. We leave the Merged Vocabulary as an alternative for other work.

\subsection{Broader Evaluation}
\label{sec:broad_evaluation}

In order to prove the generality of the warm-start method, we evaluate it across various parents and children, comparing linguistically related and unrelated languages as well as having shared English on the source or target side. We use language pairs described in \cref{sec:other_datasets}. 
%The results in this section are from our paper \persciteA{kocmi2018trivial}.

As mentioned earlier, we are going to discuss only the Balanced Vocabulary approach ignoring the Merged Vocabulary and the child-specific vocabulary. Thus, whenever we talk about warm-start technique, we consequently mean the Balanced Vocabulary technique. 

\begin{table}[t]
\caption{
Transfer learning with English reused either in source (encoder) or target (decoder).
The baselines correspond to training on one corpus only. BLEU scores are always reported for the child language pair. The scores are comparable within lines or whenever the child language pair is the same. The \significantmark{} represents significantly better results.}
\begin{center}
\begin{tabularx}{\textwidth}{@{}llCCC@{}}
\toprule
      &         & Balanced          & \multicolumn{2}{c}{Baselines: Only} \\
\cmidrule(l){4-5}
Parent & Child  & Vocabulary & Child & Parent\\
\midrule
\transl{\EN{}}{Finnish} & \transl{\EN{}}{Estonian} & \bf 19.74\significant{} & 17.03 & \p{}2.32\\
\transl{\EN{}}{Russian} & \transl{\EN{}}{Estonian} &  \bf 20.09\significant{} & 17.03 & \p{}0.57\\
\transl{\EN{}}{Czech} & \transl{\EN{}}{Estonian} &  \bf 20.41\significant{} & 17.03 & \p{}1.42\\
\midrule
\transl{Finnish}{\EN{}} & \transl{Estonian}{\EN{}} &  \bf 24.18\significant{} & 21.74 & \p{}2.44\\
\transl{Russian}{\EN{}} & \transl{Estonian}{\EN{}} &  \bf 23.54\significant{} & 21.74 & \p{}0.80\\
\midrule
\transl{\EN{}}{Czech} & \transl{\EN{}}{Slovak} & \bf 17.75\significant{} & 16.13 & \p{}6.51\\
\transl{Czech}{\EN{}} & \transl{Slovak}{\EN{}} & \bf 22.42\significant{} & 19.19 & 11.62\\
\bottomrule
\end{tabularx}
\end{center}
\label{tab:warmstart_highresourceparent}
\end{table}

\cref{tab:warmstart_highresourceparent} summarizes our results for various combinations of a high-resource parent and a low-resource child language pairs. All comparisons are with the English on the same translation side for both parent and child. The baselines models are trained exclusively on the child or parent parallel corpus. We do not report parent score on parent testset.

The column with the child baseline is essential as it shows the impact of transfer learning. We see that for all language pairs, the transfer learning significantly outperform the baseline. However, as we evaluate some linguistically related languages, for example, Czech and Slovak, we also evaluate the performance of parent model only on the child's testset to show that without transfer learning the performance is strictly worse. For the Czech and Slovak, the parent alone can roughly translate given sentences and obtain 6.51 BLEU score for direction into Slovak and 11.62 BLEU for \transl{Slovak}{English}. In contrast, Estonian and Finnish are not as related as Czech and Slovak. Thus the parent model does not perform well on the child testset, obtaining only 2.44 BLEU for \transl{Estonian}{English} translation. This highlights that the improvement brought by our method cannot be solely attributed to the relatedness of languages.
%: e.g. Czech and Slovak are much more similar than Czech and Estonian (Parent Only BLEU of translation out of English is 6.51 compared to 1.42) 
%and yet the gain from transfer learning when using \transl{English}{Czech} parent model is larger than the baseline for \transl{English}{Estonian} (+3.38) than from Slovak (+1.62), confirming the need for transfer learning.

Earlier works on \ac{NMT} transfer learning \parcite{dabre2017empirical,nguyen2017transfer} supposed linguistically related languages. We confirm their results also with our warm-start transfer learning on linguistically similar Finnish/Estonian and Czech/Slovak languages.
Furthermore, the improvements are not limited only to related languages as Estonian and Finnish. Unrelated language pairs like Czech/Estonian or Russian/Estonian work comparably well. 

\observation{Warm-start transfer learning improves performance even for unrelated languages.}{obs:warm_start_improves_unrelated}

The most surprising is the comparison of \transl{English}{Estonian} performance across various parents. We see that Finnish, the linguistically related language, improves the performance the least compared to other parents. We reach an improvement of 3.38 BLEU for \transl{English}{Estonian} when the parent model was \transl{English}{Czech}, compared to an improvement of 2.71 from \transl{English}{Finnish} parent. This two improvements are statistically significant and differ from the conclusion of \perscite{zoph2016transferLowResource}, who concluded that the more related the languages are, the better transfer learning works.
We see it as an indication that the size of the parent training set is more important than relatedness of the languages as the Czech has 40.1M sentences and the Finnish only 2.8M sentences.

The results with Russian parent for Estonian child (both directions) show that transliteration is not necessary for our approach as used in previous works \parcite{nguyen2017transfer}. Therefore there is no vocabulary sharing between Russian Cyrillic and Estonian Latin (except numbers and punctuation, see \cref{sec:warmstart_vocab_analysis} for further details), the improvement could be attributed to better coverage of English; an effect similar to domain adaptation.

We show that our method also works with the shared English on the source side. Although it is an intuitive result, we have to point out that earlier works focused only on the scenario with shared-target language \parcite{zoph2016transferLowResource,nguyen2017transfer}. Similarly to cold-start, it supports the idea that improvements are not due to the better decoder's language model.

\subsection{Combining Parent and Child Trainset}
\label{sec:combine_tag_child}

In the warm-start transfer learning scenario, we could train the parent model on a mixture of parent and child training data and then fine-tuning solely on child training data. We experiment with this setting to find out if the child performance is influenced by it. 

In multilingual learning, \perscite{johnson2017zeroshot} proposed to use a unique tag (separate word) ``<2lang>'' to identify desired target language. By adding this tag to each source sentence, they could train \ac{NMT} model on a mix of training corpora. Another option is to mix the training corpora without any unique tag.

\begin{table}[t]
\caption{Comparison of various approaches for incorporating the child data into the parent trainset. All scores are in BLEU, and neither model is significantly better than any other.}
\begin{center}
\begin{tabularx}{\textwidth}{@{}lCCC@{}}
\toprule
Child model & No-Mixing & Mix with tag & Mix without tag \\
\midrule
\transl{English}{Estonian} & \bf 20.1 & \bf 20.1 & 19.9 \\
\transl{Estonian}{English} & 23.4 & \bf 23.7 & 23.6 \\
\bottomrule
\end{tabularx}
\end{center}
\label{tab:comparison_of_tag_parent_mix_child}
\end{table}

In the following experiment, we compare Balanced Vocabulary approach with three parent models trained on various trainsets: a standard no-mixing parent-only, a mix of parent and child with added tag, a mix of parent and child \emph{without} a tag.

We use the \pair{Czech}{English} language pair as a parent and the \pair{Estonian}{English} as a child. We use \TFC{} setting.

In \cref{tab:comparison_of_tag_parent_mix_child}, we see that mixing corpora is slightly better for shared-target scenario. However, neither of approaches is significantly better than the others. Hence we see no difference between approaches. Although, this experiment could be influenced by the size of training corpora of the parent and the child. 

The previous experiment showed that the performance is not largely influenced whether we mix the child data into the parent training data or not. However, in that setting, the parent can perform translation of the child language pair, because it was trained together with the child trainset. Thus we evaluate four parent models, trained on the mix of corpora from the previous experiment, on the child testset.

\begin{table}[t]
\caption{Performance in BLEU of a parent model evaluated on the child \pair{Estonian}{English} testset. In brackets are results evaluated on individual child models from \cref{tab:comparison_of_tag_parent_mix_child}.}
\begin{center}
\begin{tabularx}{\textwidth}{@{}lCC@{}}
\toprule
Parent model & Mix with tag & Mix without tag \\
\midrule
\transl{English}{(Czech+Estonian)} & \bf 17.7 (20.1) & \p{}2.4 (19.9) \\
\transl{(Czech+Estonian)}{English} & \bf 22.0 (23.7) & 21.8 (23.6) \\
\bottomrule
\end{tabularx}
\end{center}
\label{tab:comparison_of_tag_parent_mix_parent}
\end{table}

The performance of individual parent models is reported in \cref{tab:comparison_of_tag_parent_mix_parent}. All of them perform significantly worse than after transfer learning of the child model. However, for the scenario, where the target language is shared between Czech and Estonian pair, or when the tag marks the target desired language, it is interesting that the models' performances are only slightly lower than for transfer learning than we would expect.

\subsection{Balanced Vocabulary Analysis}
\label{sec:warmstart_vocab_analysis}

Our Balance Vocabulary method relies on the vocabulary estimated jointly from the child and parent model. In the Transformer, the vocabulary is typically shared by both the encoder and the decoder. We analyzed the vocabulary in our cold-start scenario in \cref{sec:vocabulary_overlap}, where we found out that around 40\% of parent vocabulary is unused after the transfer. 

Balanced Vocabulary is prepared in the following way. We take an equal part of parent and child corpus and generate a wordpiece vocabulary that is used both by the parent as well as the child model. With a large overlap, we could expect a lot of ``information reuse'' between the parent and the child.

% Since the subword vocabulary depends on the training corpus; a little clarification is needed. We provide an analysis of the balanced vocabulary technique and what vocabulary subwords appear in various corpora.

We take the vocabulary of subword units as created for \pair{Russian}{English} parent and \pair{Estonian}{English} child experiments. This vocabulary contains 28.2k subwords in total. We then process the training corpus for each of the languages with this shared vocabulary, ignore all subwords that appear less than 10 times in each of the languages (these subwords will have little to no impact on the result of training) and break down the total of 28.2k subwords into overlapping classes based on the languages where the particular subword was observed, see \cref{tab:enruet-breakdown}.

\def\yes{\checkmark}
\def\no{-}
\begin{table}[t]
\caption{
Breakdown of subword vocabulary of experiments involving \pair{Russian}{English} parent and \pair{Estonian}{English} child.}
\begin{center}
\begin{tabularx}{\textwidth}{@{}CCCr@{}}
\toprule
Estonian & English & Russian & \% Subwords \\
\midrule
\yes & \no & \no & 29.93 \\
\no & \yes & \no & 20.69 \\
\no & \no & \yes & 29.03 \\
\yes & \yes & \no & 10.06 \\
\no & \yes & \yes & 1.39 \\
\yes & \no & \yes & 0.00 \\
\yes & \yes & \yes & 8.89 \\
\midrule
\multicolumn{3}{l}{Reused parent} & 41.03 \\
\bottomrule
\end{tabularx}
\end{center}
\label{tab:enruet-breakdown}
\end{table}

We see that the vocabulary is reasonably balanced, with each language having 20--30\% of subwords unique (see the first three rows of \cref{tab:enruet-breakdown}). English and Estonian share 10\% subwords not seen in Russian while Russian shares only 1.39\% and 0\% of subwords with each of the other languages, possibly due to the Cyrillic script. Overall, 8.89\% of subwords are used in all three languages.

A particularly interesting subset is the one where parent subwords are used by the child model. In other words, subwords appearing anywhere in English and also tokens common to Estonian and Russian. For this set of languages, this amounts to 20.69+10.06+1.39+0.0+8.89 = 41.03\%. We list this number on a separate row in \cref{tab:enruet-breakdown}, as ``Reused parent''. These subwords get their embeddings trained better thanks to the parent model.
However, the vocabulary also contains 29.04\% subwords used only by the parent and unused by the child.

\cref{tab:vocab_stats} summarizes this analysis for several language sets used in the warm-start experiments, listing what portion is shared by all the languages (column ``In All''), what portion of subwords benefits from the parent training (column ``Reused from Parent'') and what portion of vocabulary is unused by the child (column ``Unused by Child). 

We see a similar picture across the board; roughly 30\% of subwords are unused by the child model. And roughly 50\% of subwords are unused whenever both parent languages are distinct from the child. We remind that in the Direct Transfer the number was around 40\%. We already discussed this problem as these subwords only slow down the training and inference as they are useless to the child.

\observation{Balanced Vocabulary has 30\% of tokens unused by the child in the vocabulary whenever one language is shared between parent and child.}{obs:warmstart_has_thirtypercent_unused}

\begin{table}[t]
\caption{Summary of vocabulary overlaps for the various language sets. The first column specifies what is the parent language pair. The child language pair is \pair{Estonian}{English} for all rows. All figures represent percentage of the vocabulary.}
\begin{center}
\begin{tabularx}{\textwidth}{@{}l@{~~~}HC@{~~~}C@{~~~}C@{}}
\toprule
Languages & Unique in a Lang. & In All & Reused Parent & Unused by Child\\
\midrule
\pair{Finnish}{\EN{}} & 24.4-18.2-26.2 & 19.5 \% & 49.4 \% & 26.2 \%\\
\pair{Russian}{\EN{}} & 29.9-20.7-29.0 & \p{}8.9 \% &  41.0 \% & 29.0 \%\\
\pair{Czech}{\EN{}} & 29.6-17.5-21.2   & 20.3 \% & 49.2 \% & 21.2 \%\\
\pair{Arabic}{Russian} & 28.6-27.7-21.2-9.1 & \p{}4.6 \% & \p{}6.2 \% & 56.3 \%\\
\pair{Spanish}{French} & 15.7-13.0-24.8-8.8 & 18.4 \% & 34.1 \% & 28.7 \%\\
\pair{Spanish}{Russian} & 14.7-31.1-21.3-9.3 & \p{}6.0 \% & 21.4 \% & 45.8 \%\\
\pair{French}{Russian} & 12.3-32.0-22.3-8.1 & \p{}6.3 \% & 23.1 \% & 44.3 \%\\
\bottomrule
\end{tabularx}
\end{center}
\label{tab:vocab_stats}
\end{table}

We list vocabularies used in our main results in \cref{sec:broad_evaluation} as well as language pairs not containing any shared language between parent and child (English in our case) with the child as we report in \cref{sec:no_language_in_common}.

The Arabic-Russian-Estonian-English stands out with the very low number of subwords (6.2\%) available already in the parent, mainly due to the scripts of parent language not using Latin. The parent Arabic-Russian thus offered very little word knowledge to the child, and yet it leads to a performance gain (21.74 vs. 22.23 BLEU, see \cref{sec:no_language_in_common}).

Our observations indicate that the key factor is the size of the parent corpus rather than vocabulary overlaps. However, the reasons for the gains are yet to be explained in detail. 

\section{Warm-Start and Cold-Start Comparison}
\label{sec:warm_cold_comparison}

We study two approaches for transfer learning that differ in how the parent model is treated, specifically by allowing modification to the parent vocabulary before or after the training. In this section, we compare both cold-start approaches: the cold-start Direct Transform and Transformed Vocabulary with the warm-start approach of Balanced Vocabulary.

We train four parent models for warm-start approach differing in the vocabulary: two \transl{English}{Czech} and two \transl{Czech}{English} on the same parent training data. For the cold-start approaches, we used the models trained by \perscite{popel2018wmt} similarly as in cold-start experiments (see \cref{sec:cold_start_transfer}).

Results are from \persciteA{kocmi2019ecofriendly} and we use \TFB{}.

\begin{table}[t]
\caption{The scores (BLEU) for cold-start methods (Direct Transfer and Transformed Vocabulary) with the warm-start method of Balanced Vocabulary.
}
\begin{center}
\begin{tabularx}{\textwidth}{@{}lCcC@{}}
\toprule
Child language pair     & Direct Transfer & Transformed Voc. & Balanced Voc.  \\
\midrule               
\transl{\EN{}}{Estonian}   & \bf 20.75 & 20.27 &  20.62 \\ 
\transl{\EN{}}{Russian}   & 25.50 & 28.65 & \bf 29.03\significant{}\\
\transl{Estonian}{\EN{}}   & 24.36 & 24.64 & \bf 26.00\significant{}\\
\transl{Russian}{\EN{}}    & 23.41 & \bf 31.38 & 31.15 \\
\bottomrule
\end{tabularx}
\end{center}
\label{tab:cold_warm_comparison_bleu}
\end{table}

The results are tabulated in \cref{tab:cold_warm_comparison_bleu}. We see that the warm-start method reaches in most cases significantly better performance in both directions. The only exception is the \transl{Russian}{English} and \transl{English}{Estonian}, where neither warm-start Balanced Vocabulary nor the Transformed Vocabulary is significantly better than the other.

The cold-start transfer learning improves the performance of high-resource language pairs (see \cref{obs:coldstart_improves_highresource}), and the warm-start improve the performance over the cold-start transfer learning. Hence warm-start should work for high-resource language pairs. This is confirmed by the \pair{Russian}{English} language pair in \cref{tab:cold_warm_comparison_bleu} and \pair{Slovak}{English} in \cref{tab:warmstart_highresourceparent}. We study it further in the next \cref{chap:analysis}.

\observation{Both cold-start and warm-start transfer learning improve performance of low-resource and high-resource language pairs.}{obs:transfer_improves_also_highresource}

The better performance of the warm-start approach is understandable since the parent model is already trained with vocabulary prepared to accommodate a given child language pair. It does not have the high segmentation problem as the Direct Transfer has, and it does not have to retrain randomly assigned embeddings as in Transformed Vocabulary.

However, the warm-start approach has one disadvantage over the cold-start: we cannot reuse \emph{any} parent model, and we have to train the parent model for each child language pair separately. With that in mind, we investigate the number of steps needed to reach the performance as presented in \cref{tab:cold_warm_comparison_bleu}.

\cref{tab:cold_warm_comparison_steps} shows the total number of training steps. The cold-start scenarios show only steps needed for the child model convergence since we did not train the parent model by ourselves as we are using the model by \perscite{popel2018wmt}. In contrast, the steps for Balanced Vocabulary show the total number of steps the parent and the child were training altogether.

By observing the results, we see that due to the parent training, the warm-start scenario takes more steps to train. However, the total training time of the parent can vary broadly. Furthermore, in \cref{sec:parent_performance_influence}, we show that even a parent model that did not fully converge is a good parent for transfer learning.

In conclusion, the cold-start method has the advantage of not requiring to train the parent model, it does not need any modification to the training workflow (in Direct Transfer) or only a trivial modification of vocabulary (in Transformed Vocabulary) and reaches a better performance than training a baseline model. This makes it the candidate to most of the current training workflows, where researches could start using any model as a parameter initialization instead of random initialization as it is the current standard.

\begin{table}[t]
\caption{Comparison of the number of steps needed for cold-start and warm-start methods to converge.}
\begin{center}
\begin{tabularx}{\textwidth}{@{}lCcC@{}}
\toprule
Child language pair     & Direct Transfer & Transformed Voc.  & Balanced Voc. \\
\midrule               
\transl{\EN{}}{Estonian} &  \p{}75k &  \p{}75k & \p{}735k  \\
\transl{\EN{}}{Russian}  & 630k & 450k & 1510k \\
\transl{Estonian}{\EN{}} &  \p{}30k &  \p{}60k & \p{}700k  \\
\transl{Russian}{\EN{}}  & 420k & 700k & 1465k \\
\bottomrule
\end{tabularx}
\end{center}
\label{tab:cold_warm_comparison_steps}
\end{table}

On the other hand, whenever the time is not a relevant criterion, but we are focused on the performance of the model, the warm-start scenario would be the recommended approach.

\section{Related Work}
\label{sec:transfer_relatedwork}

The idea of transferring knowledge between different algorithms is as old as machine learning itself, and researchers were trying to reuse previously trained features on different tasks. \perscite{pratt1991direct} published a paper properly formulating the transfer learning problem by pretraining a neural network on a different task.

\perscite{firat2016multiway} studied multitask learning, a scenario closely related to transfer learning. They propose multi-way multilingual systems, a generalization of work by \perscite{dong2015multitask}, with the primary goal of reducing the total number of parameters needed to cater multiple source and target languages. The system uses individual encoders and decoders for each language, with only the attention mechanism shared.
The model is trained on one language pair at a time, thus to keep all the language pairs active in the model, a special training schedule is needed. Otherwise, ``catastrophic forgetting'' would hinder the ability to translate among the languages trained earlier. 
\perscite{firat2016multiway} experimented with six languages, where the training data were between English and one of the remaining five languages. They showed that the multilingual setup with shared attention outperforms the single-pair baseline on low-resource languages in all translation directions. 

Nevertheless, when using high-resource languages with more than two million training sentences, experiments yield counter-intuitive results. The multi-way system outperforms the single-pair baseline whenever English is on the target side. However, whenever English is on the source side, the baseline performs on-par or better than the multi-way. The authors did not provide any explanation for this phenomenon. We discuss it in more depth when evaluating our results in \cref{sec:more_data_is_more_important}. We believe that the improvements only for direction into the English are mainly due to the higher amount of sentences the English decoder had access to during the training in comparison to single-language-pair baseline. Therefore, the model can generate better English sentences. \perscite{firat2016multiway} suggest that transferring knowledge from the language shared between parent and child is a harder scenario than transferring with the shared language. We analyze it in \cref{sec:shared_decoder_is_easier}.

\perscite{johnson2017zeroshot} introduce another multilingual approach. They add a special word (or token) to each sentence, which indicates the desired target language. Then they mixed together all parallel sentences from all language pairs together. Thus the \ac{NMT} could recognize a target language by the token ``<2lang>'' at the beginning of each sentence. The architecture then shares all parameters between all languages, not only the attention as in \perscite{firat2016multiway}.
The model implicitly learns translation between all languages. However, it must be noted that the performance of their system is worse than the single-pair baseline. This is due to the same amount of parameters as in the baseline that is used between several language pairs, thus effectively having less amount of parameters for each used language. \perscite{johnson2017zeroshot} support the hypothesis by increasing the size of the multilingual model, which reduces the performance gap to the baseline. The most interesting result is that the model can perform zero-shot translation, i.e. translating between languages never seen in the training set in a language pair.

There are relatively few studies investigating transfer learning in the area of the \ac{NMT}. The reasons could be due to the brief history of \ac{NMT}, which beginning is dated to the work of \perscite{sutskever2014sequence}, who presented the first end-to-end \ac{NMT} model and it took until 2016 it became a major paradigm in \ac{MT} \parcite{bojar2016wmt}.

\perscite{zoph2016transferLowResource} claims to be the first to apply transfer learning to the low-resource \ac{NMT} successfully. They used the word-level \ac{RNN} translation model \parcite{sutskever2014sequence} with separate vocabularies for the source and target language. Their use shared English only in the scenario of shared-target. They exploit the cold-start scenario, where in contrast to our work, their approach required freezing the English embeddings from updating during the child's training. As for the second language, they randomly assign child words to the parent trained embeddings.
They showed that their transfer learning improves the performance of \ac{NMT} systems in translation from Hansa, Turkish, and Uzbek into English with the help of a \transl{French}{English} parent model. 
%Interestingly, they showed that the random word initialization is as good as assigning words based on their lexical similarity, we get the same conclusion in \cref{sec:various_vocabulary_transformations}. 
Unfortunately, the authors have not experimented with translation direction from English. Shared-target language scenario is an easier task as we show in \cref{sec:shared_decoder_is_easier}. Furthermore, they got the biggest improvements when using the transferred model as a re-scorer of \ac{SMT} approach.

\perscite{nguyen2017transfer} build on top of the work of \perscite{zoph2016transferLowResource}. In contrast, they focused on a scenario where both parent and the child are low-resource and linguistically related.
They improved the transfer learning by using a \ac{BPE} vocabulary instead of word-level vocabulary as in \perscite{zoph2016transferLowResource}. The vocabulary is prepared in advance from both parent and child training data, i.e. they use the warm-start scenario. Their approach is similar to our Balanced Vocabulary with several differences. They used transliteration of Arabic script in order to normalize training corpora to Latin script and used segmentation to increase the number of overlapping tokens between languages. Furthermore, they have not balanced the corpora in any way. With that setup, they show that linguistic relatedness plays an important role and that transfer learning also helps in a scenario where both parent and child are low-resource. They demonstrated that transfer learning works better on subword-level \ac{NMT} than on word-level mainly due to the higher overlap of tokens between languages. Additionally, they showed that freezing parent parameters such as English embeddings \parcite{zoph2016transferLowResource} is not necessary for the transfer learning and can even hurt its performance.
Unfortunately, they also evaluated only the scenario with shared-target language, leaving a place to investigate the effect in the more difficult setting with the shared-source language. In \cref{sec:broad_evaluation}, we showed that the relatedness of languages is not necessary for successful transfer learning.

\perscite{kim2019effective} solved the problem with mismatching vocabulary in cold-start approach by applying cross-lingual word embeddings. They trained monolingual word embeddings \parcite{mikolov2013efficient} for the parent and child source language and then computed the linear mapping between the two embedding spaces. The child model then continues the training from the parent model with modified word embeddings. The linear mapping is applied only to the encoder embeddings. The method assumes the same target language for parent and child model.

\perscite{Lakew2018dynamic} experiment with a cold-start transfer learning in the multilingual \ac{NMT} setting. They tackle the problem with vocabulary mismatch by dynamically adapting the vocabulary in a similar way as our ``Vocabulary Transformation Technique'' (see \cref{sec:vocabulary_transformation}). During the transfer, the parent vocabulary items that are not used by the child are replaced by random tokens from the child vocabulary. However, in contrast to our technique, they reset embedding weights for the randomly assigned words.
They evaluate the method on the multilingual setting of four language pairs, where instead of transferring from one parent to one child, they transfer parameters consecutively from one parent to the first child, followed by transferring from first child to the second child and lastly finishing by transfer learning of the third child. Before each change of training corpus, the vocabulary is dynamically updated for the upcoming language pair.
They proposed two approaches: either incrementally add training data to the parent training corpus, thus increasing the number of training sentences with each consecutive parent or to use only the current child's training data. The former approach can prevent worsening the performance of the child on previous parent testsets. However, the latter approach reaches a better score on the latest child language pair with the disadvantage of losing the performance on the earlier parents.

\perscite{neubig2018rapid} combined the multilingual \ac{MT} \parcite{johnson2017zeroshot} with transfer learning by \perscite{zoph2016transferLowResource}. They train a multilingual general system that can translate between all investigated language pairs followed by specializing the model to one high-resource language pair. They trained the multilingual parent on 58 language pairs. Their main goal is low-resource languages for which they come up with the technique of using a helper related language pair, which is mixed into the training set. For example, they mix Slovak and Czech sentences aligned on the English side. They confirm results from \perscite{johnson2017zeroshot} that using a single-pair baseline, in this case, mixing the related languages, performs better than the multilingual model. However, when they used the transfer learning technique on the multilingual parent model, the performance significantly improved over single-pair baseline, which shows the usefulness of multilingual parent in transfer learning.

\section{Conclusion}
\label{sec:transfer_conclusion}

We proved transfer learning to be an effective technique for \ac{NMT} under low-resource conditions both by our experiments and the related work. Existing methods require a shared-target language, language relatedness, or specific training tricks and regimes. In contrast, we describe our approach to transfer learning leveraging these constrains in both cold-start and warm-start setting. Furthermore, we showed that the relatedness of languages is not critical for transfer learning, nor we needed transliteration for our methods.
Another important ability of our approaches is to handle language pairs that have different target language between a parent and child model. 

In this chapter, we have explained the transfer learning technique on \ac{NMT} and discussed a difference between cold-start and warm-start techniques. We proposed two techniques: Direct Transfer and Vocabulary Transformation techniques in the cold-start setting, where we investigated the influence of parent vocabulary on the child language pair. In \cref{sec:warm_start_transfer}, we studied the warm-start scenario and proposed Merged and Balanced Vocabulary techniques. Furthermore, we evaluated these approaches as well as using child-specific vocabulary for the parent training, which surprisingly performed worse than other techniques. 
Lastly, in \cref{sec:warm_cold_comparison}, we compared our approaches and showed that Balance vocabulary obtains the best performance out of proposed methods. However, the Transformed Vocabulary needs much less training steps and obtains results similar to Balanced Vocabulary.

Note that all our presented methods are relatively simple to implement with current frameworks and training procedures, which could be a motivation for the community for applying them.
We believe that our Vocabulary Transformation technique can replace standard approaches of training models from random initialization as it reaches better performance in a shorter or comparable number of training steps relative to the training from random initialization.

We want to analyze what is behind the improvements, and deeper understand the technique. Therefore, in the next chapter, we focus on an extensive analysis of transfer learning with an ambition to explain how transfer learning works and if we can use this knowledge to improve the transfer learning technique.

\chapter{Analysis}
\label{chap:analysis}
% \addcontentsline{toc}{chapter}{Analysis}

In the previous chapter, we showed the advantages of the transfer learning technique and the gains it brings. However, neural networks are used without us rigorously understanding the exact gains behind individual techniques. Thus neural networks gained the reputation of being a black box. 

As \perscite{rudin2018please} said, ``A black-box model is either a function that is too complicated for any human to comprehend or a function that is proprietary; it is a model that is difficult to troubleshoot. Deep learning models, for instance, tend to be black boxes because they are highly recursive.'' In other words, neural networks are difficult to understand for their deep recursive architecture and often unpredictable behavior. For example, in image classification, \perscite{goodfellow2014explaining} showed that \ac{NN} could be deceived by a slightly modified image, which is indistinguishable from the original for humans, yet the \ac{NN} cannot classify it correctly.

In this chapter, we try to investigate what is really behind the achievements of transfer learning by analyzing the behavior, training process, and what is transferred from the parent to the child. We dive into the analysis of transfer learning. 

Although we presented several cold-start and warm-start methods in \cref{chap:transfer_learning}, we are going to analyze only the warm-start Balanced Vocabulary approach in order to make the analysis consistent. We decided on this approach as it reaches the best performance. However, we are convinced that other presented methods behave similarly.

This chapter is organized into sections that analyze various aspects of transfer learning. We start by investigating the negative effects of transfer learning in \cref{sec:negative_transfer}. In \cref{sec:shared_decoder_is_easier}, we discuss the differences between the scenarios with shared-source and shared-target language. Then in \cref{sec:more_data_is_more_important}, we try to answer if the training data size is more important than the language relatedness. We follow with a discussion if the gains stem from linguistic features or simply from a better initialization of the neural network in \cref{sec:linguistic_vs_initialization}. We conclude the results from analysis in \cref{sec:discussion_on_analysis}. Lastly, in \cref{sec:backtranslation}, we perform a case study an application of transfer learning on backtranslation.

\section{Negative Transfer}
\label{sec:negative_transfer}

In generic machine learning, transfer learning is also known for its downsides \parcite{pan2010transfersurvey,weiss2016survey}. When transferring knowledge from a less related task, it may hurt the final performance on the child task in comparison to the performance obtained without transfer learning only with the use of a child model. This harmful effect is called ``Negative transfer''.

The main reason behind the negative transfer is often the domain mismatch between the parent and child tasks or even an unrelated parent domain \parcite{pan2010domain,ge2014handling}, which prevents the model from the utilization of the parent model during transfer learning.

\perscite{wang2019characterizing} proposed a formal definition of the negative transfer and evaluated the definition on several transfer learning approaches. They evaluated the following three critical factors influencing the negative transfer:

\begin{enumerate}
  \item Divergence between the joint distributions of both tasks is hurting transfer learning.
  \item Effectiveness of transfer learning depends on the size of child data.
  \item Transfer learning should be evaluated with the same setting of the neural network to avoid adding a risk of different setups.
\end{enumerate}

As for the first factor, the ideal transfer learning should figure out and take advantage of only the similar parts of tasks, however, in the real-life scenario it often takes into account also the misleading information learned from the parent task. The second factor elaborates that the less training data is available in the child domain, the more fragments are preserved from the parent task, which decreases performance on the child task. On the other hand, when we have plentiful of child data, a better baseline can be trained, which reaches a better performance than transfer learning. Thus negative transfer is relatively more likely to occur. The last factor is to avoid misjudgment by comparing the performance of transfer learning with a baseline using different parameters. For example, fine-tuning hyper-parameters separately for the transferred model and baseline will likely lead to different results due to the hyper-parameter setting.

Transfer learning in the field of \ac{NMT} emerged recently \parcite{zoph2016transferLowResource}. Thus there is a lack of research on the negative transfer in this field. \perscite{zoph2016transferLowResource} have not discovered any problematic behavior of transfer learning. It could be due to the design of the experiment that avoids the negative transfer altogether. For example, initial works on transfer learning in \ac{NMT} examine only linguistically related language pairs \parcite{nguyen2017transfer, neubig2018rapid}. Another possible explanation is that the neural networks are robust enough that they can re-train any transferred parameters hence avoiding the negative transfer at all.

In this section, we try to shed some light on the negative transfer in \ac{NMT} by evaluating several experiments and identifying the possible downsides of transfer learning in \ac{NMT}. In \cref{sec:traces_of_parent_language_pair}, we investigate if the parent target language leaks to the outputs of the child. Then in \cref{sec:extremely_low_resource_transfer}, we study the condition of having an extremely low-resource child and if that hurts the transfer learning success. In \cref{sec:low_resource_parent}, we study the reverse effect of having a parent with less parallel sentences than the child. \perscite{wang2019characterizing} mentioned that another factor negatively influencing transfer learning is a divergence between parent and child distributions. Thus we investigate the scenario with no-shared language in \cref{sec:no_language_in_common}. Lastly, we conclude in \cref{sec:conclusion_on_negative_transfer}.

\subsection{Traces of Parent Language Pair}
\label{sec:traces_of_parent_language_pair}

\perscite{wang2019characterizing} mention as an effect of negative transfer that parent fragments appear in the child outputs whenever the child task has a low amount of data. This is mainly because the child has not entirely forgotten the parent task. In this section, we investigate traces of the parent language pair in child translation, such as text fragments.

During transfer learning, the neural network is not notified about the change in the language pair. This means that the \ac{NN} has to forget the parent task during the training on child parallel corpus. This can result in fragments of parent target language appearing in the child model output.

In order to test if there are any traces of the parent language pair in the output of the child model, LanideNN (see \cref{sec:lanidenn}) automatically identifies the language of transfer learning outputs, and we measure how often does the parent language pair appear in the child output.

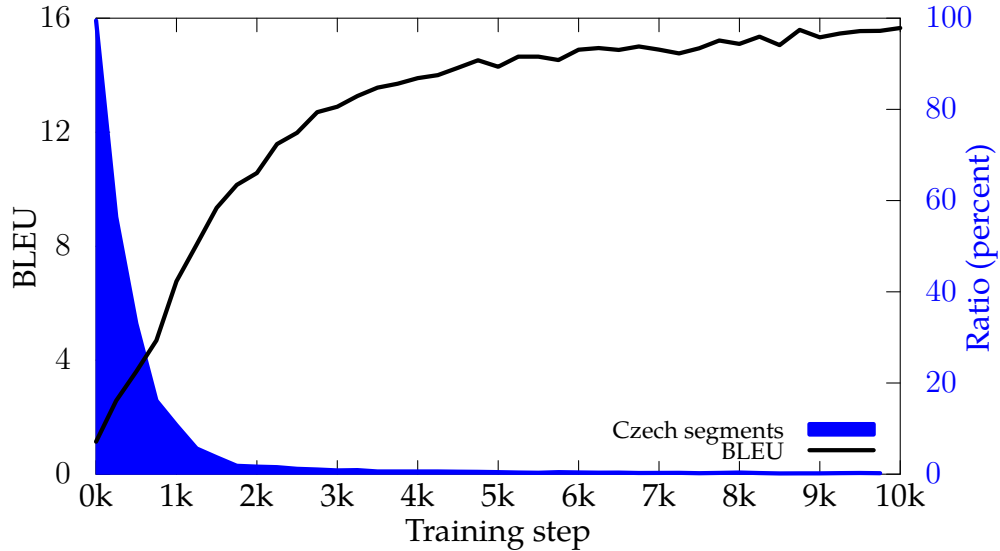
\begin{figure}
\caption{The graph represent behavior of child model during first 10k training steps. The blue area represents the ratio of Czech segments in the output of child model immediately after the transfer learning start. The black curve illustrates the BLEU score on the \transl{English}{Estonian} development set.}
\begin{center}
\begin{tikzpicture}[gnuplot]
%% generated with GNUPLOT 5.0p3 (Lua 5.1; terminal rev. 99, script rev. 100)
%% Wed 07 Aug 2019 05:47:21 PM DST
\path (0.000,0.000) rectangle (4.500,7.400);
\gpcolor{color=gp lt color border}
\gpsetlinetype{gp lt border}
\gpsetdashtype{gp dt solid}
\gpsetlinewidth{1.00}
\draw[gp path] (1.136,0.985)--(1.316,0.985);
\node[gp node right] at (0.952,0.985) {$0$};
\draw[gp path] (1.136,2.497)--(1.316,2.497);
\node[gp node right] at (0.952,2.497) {$4$};
\draw[gp path] (1.136,4.008)--(1.316,4.008);
\node[gp node right] at (0.952,4.008) {$8$};
\draw[gp path] (1.136,5.520)--(1.316,5.520);
\node[gp node right] at (0.952,5.520) {$12$};
\draw[gp path] (1.136,7.031)--(1.316,7.031);
\node[gp node right] at (0.952,7.031) {$16$};
\draw[gp path] (1.136,0.985)--(1.136,1.165);
\draw[gp path] (1.136,7.031)--(1.136,6.851);
\node[gp node center] at (1.136,0.677) {0k};
\draw[gp path] (2.194,0.985)--(2.194,1.165);
\draw[gp path] (2.194,7.031)--(2.194,6.851);
\node[gp node center] at (2.194,0.677) {1k};
\draw[gp path] (3.252,0.985)--(3.252,1.165);
\draw[gp path] (3.252,7.031)--(3.252,6.851);
\node[gp node center] at (3.252,0.677) {2k};
\draw[gp path] (4.310,0.985)--(4.310,1.165);
\draw[gp path] (4.310,7.031)--(4.310,6.851);
\node[gp node center] at (4.310,0.677) {3k};
\draw[gp path] (5.368,0.985)--(5.368,1.165);
\draw[gp path] (5.368,7.031)--(5.368,6.851);
\node[gp node center] at (5.368,0.677) {4k};
\draw[gp path] (6.427,0.985)--(6.427,1.165);
\draw[gp path] (6.427,7.031)--(6.427,6.851);
\node[gp node center] at (6.427,0.677) {5k};
\draw[gp path] (7.485,0.985)--(7.485,1.165);
\draw[gp path] (7.485,7.031)--(7.485,6.851);
\node[gp node center] at (7.485,0.677) {6k};
\draw[gp path] (8.543,0.985)--(8.543,1.165);
\draw[gp path] (8.543,7.031)--(8.543,6.851);
\node[gp node center] at (8.543,0.677) {7k};
\draw[gp path] (9.601,0.985)--(9.601,1.165);
\draw[gp path] (9.601,7.031)--(9.601,6.851);
\node[gp node center] at (9.601,0.677) {8k};
\draw[gp path] (10.659,0.985)--(10.659,1.165);
\draw[gp path] (10.659,7.031)--(10.659,6.851);
\node[gp node center] at (10.659,0.677) {9k};
\draw[gp path] (11.717,0.985)--(11.717,1.165);
\draw[gp path] (11.717,7.031)--(11.717,6.851);
\node[gp node center] at (11.717,0.677) {10k};
\draw[gp path] (11.717,0.985)--(11.537,0.985);
\gpcolor{rgb color={0.000,0.000,1.000}}
\node[gp node left] at (11.901,0.985) {$0$};
\gpcolor{color=gp lt color border}
\draw[gp path] (11.717,2.194)--(11.537,2.194);
\gpcolor{rgb color={0.000,0.000,1.000}}
\node[gp node left] at (11.901,2.194) {$20$};
\gpcolor{color=gp lt color border}
\draw[gp path] (11.717,3.403)--(11.537,3.403);
\gpcolor{rgb color={0.000,0.000,1.000}}
\node[gp node left] at (11.901,3.403) {$40$};
\gpcolor{color=gp lt color border}
\draw[gp path] (11.717,4.613)--(11.537,4.613);
\gpcolor{rgb color={0.000,0.000,1.000}}
\node[gp node left] at (11.901,4.613) {$60$};
\gpcolor{color=gp lt color border}
\draw[gp path] (11.717,5.822)--(11.537,5.822);
\gpcolor{rgb color={0.000,0.000,1.000}}
\node[gp node left] at (11.901,5.822) {$80$};
\gpcolor{color=gp lt color border}
\draw[gp path] (11.717,7.031)--(11.537,7.031);
\gpcolor{rgb color={0.000,0.000,1.000}}
\node[gp node left] at (11.901,7.031) {$100$};
\gpcolor{color=gp lt color border}
\draw[gp path] (1.136,7.031)--(1.136,0.985)--(11.717,0.985)--(11.717,7.031)--cycle;
\node[gp node center,rotate=-270] at (0.246,4.008) {BLEU};
\gpcolor{rgb color={0.000,0.000,1.000}}
\node[gp node center,rotate=-270] at (12.790,4.008) {Ratio (percent)};
\gpcolor{color=gp lt color border}
\node[gp node center] at (6.426,0.215) {Training step};
\gpfill{rgb color={0.000,0.000,1.000}} (1.136,7.000)--(1.401,4.395)--(1.665,2.983)--(1.930,1.961)%
    --(2.194,1.635)--(2.459,1.326)--(2.723,1.209)--(2.988,1.096)--(3.252,1.085)%
    --(3.517,1.078)--(3.781,1.056)--(4.046,1.047)--(4.310,1.035)--(4.575,1.039)%
    --(4.839,1.021)--(5.104,1.021)--(5.368,1.021)--(5.633,1.021)--(5.897,1.018)%
    --(6.162,1.016)--(6.427,1.012)--(6.691,1.007)--(6.956,1.004)--(7.220,1.012)%
    --(7.485,1.008)--(7.749,1.005)--(8.014,1.006)--(8.278,1.001)--(8.543,1.002)%
    --(8.807,1.003)--(9.072,0.998)--(9.336,1.002)--(9.601,1.007)--(9.865,1.000)%
    --(10.130,0.995)--(10.394,0.996)--(10.659,0.996)--(10.923,0.999)--(11.188,1.001)%
    --(11.452,0.998)--(11.452,0.985)--(1.136,0.985)--cycle;
\gpcolor{rgb color={0.000,0.000,1.000}}
\gpsetlinewidth{4.00}
\draw[gp path] (1.136,7.000)--(1.401,4.395)--(1.665,2.983)--(1.930,1.961)--(2.194,1.635)%
  --(2.459,1.326)--(2.723,1.209)--(2.988,1.096)--(3.252,1.085)--(3.517,1.078)--(3.781,1.056)%
  --(4.046,1.047)--(4.310,1.035)--(4.575,1.039)--(4.839,1.021)--(5.104,1.021)--(5.368,1.021)%
  --(5.633,1.021)--(5.897,1.018)--(6.162,1.016)--(6.427,1.012)--(6.691,1.007)--(6.956,1.004)%
  --(7.220,1.012)--(7.485,1.008)--(7.749,1.005)--(8.014,1.006)--(8.278,1.001)--(8.543,1.002)%
  --(8.807,1.003)--(9.072,0.998)--(9.336,1.002)--(9.601,1.007)--(9.865,1.000)--(10.130,0.995)%
  --(10.394,0.996)--(10.659,0.996)--(10.923,0.999)--(11.188,1.001)--(11.452,0.998);
\gpcolor{rgb color={0.000,0.000,0.000}}
\draw[gp path] (1.136,1.416)--(1.401,1.960)--(1.665,2.346)--(1.930,2.759)--(2.194,3.543)%
  --(2.459,4.028)--(2.723,4.517)--(2.988,4.821)--(3.252,4.978)--(3.517,5.362)--(3.781,5.512)%
  --(4.046,5.785)--(4.310,5.857)--(4.575,5.999)--(4.839,6.110)--(5.104,6.162)--(5.368,6.236)%
  --(5.633,6.276)--(5.897,6.373)--(6.162,6.473)--(6.427,6.387)--(6.691,6.521)--(6.956,6.521)%
  --(7.220,6.476)--(7.485,6.613)--(7.749,6.635)--(8.014,6.610)--(8.278,6.656)--(8.543,6.614)%
  --(8.807,6.564)--(9.072,6.632)--(9.336,6.735)--(9.601,6.690)--(9.865,6.786)--(10.130,6.674)%
  --(10.394,6.876)--(10.659,6.777)--(10.923,6.828)--(11.188,6.860)--(11.452,6.863)--(11.717,6.901);
\gpfill{color=gpbgfillcolor} (8.033,1.165)--(11.533,1.165)--(11.533,1.719)--(8.033,1.719)--cycle;
\gpcolor{color=gp lt color border}
\node[gp node right,font={\fontsize{9pt}{10.8pt}\selectfont}] at (10.357,1.580) {Czech segments};
\gpfill{rgb color={0.000,0.000,1.000}} (10.523,1.511)--(11.367,1.511)--(11.367,1.649)--(10.523,1.649)--cycle;
\gpcolor{rgb color={0.000,0.000,1.000}}
\draw[gp path] (10.523,1.511)--(11.367,1.511)--(11.367,1.649)--(10.523,1.649)--cycle;
\gpcolor{color=gp lt color border}
\node[gp node right,font={\fontsize{9pt}{10.8pt}\selectfont}] at (10.357,1.303) {BLEU};
\gpcolor{rgb color={0.000,0.000,0.000}}
\draw[gp path] (10.523,1.303)--(11.367,1.303);
%% coordinates of the plot area
\gpdefrectangularnode{gp plot 1}{\pgfpoint{1.136cm}{0.985cm}}{\pgfpoint{11.717cm}{7.031cm}}
\end{tikzpicture}
%% gnuplot variables
\end{center}
\label{fig:parent_traces}
\end{figure}

We used our LanideNN as it is trained to recognize language switching within one sentence instead of labeling the whole sentence by one label; we can get distribution over each character. We calculate the score by labeling each character in the testset with the language label and then calculating the ratio of labels for each language in the testset.

We evaluate the model every 250 steps, which is roughly every three minutes of training. As the parent model we use \transl{English}{Czech} and the child model is \transl{English}{Estonian}. These language pairs have different target language from different language family, which should help when automatically recognizing the language. For this evaluation, we use the \transl{English}{Estonian} development set and \TFC{}.

From \cref{fig:parent_traces}, we see that the \ac{NMT} model quickly forgets generation of Czech sentences, after just 3k steps the model generates less than 1\% of Czech data. The training time for 3k steps took only 43 minutes. We remind the reader that the standard time of training even extremely low-resource language pairs is at least 50k training steps. Therefore it takes only a fraction of time for the \ac{NN} to forget the parent target language.

\observation{During the training of a child model, the \ac{NN} almost instantly forgets the parent target language and adapts to the child target language.}{obs:child_forgets_parent_instantly}

We need to mention that the results are based on an automatic measure and that LanideNN's error rate, needs to be taken into account (in a multilingual setting, the error rate is less than 4\%, see \cref{sec:lanidenn_multilingual_experiments}). We note that this automatic language detection cannot be reliably used for fine-grained evaluation to investigate if the child occasionally generates the parent target language.

In order to evaluate how often child model produces parent (Czech) words, we used the final child model to translate 100k English sentences randomly selected from the parent training corpus. We chose the parent training corpus as these sentences could be memorized by \ac{NN} from the parent model training and it is thus most likely that the training input sentences could trigger the parent behavior in the child model. The final child model was trained for 75k steps.

\begin{table}[t]
\caption{Relic words from parent language in the child output. The English gloss is our (manual) translation of a given Czech word.}
\begin{center}
\begin{tabularx}{\textwidth}{@{}CCC@{}}
\toprule
Appearances & Czech & English Gloss\\
\midrule
49 & k{\'a}mo & buddy \\
31 & \v{C}l{\'a}nek & Article \\
20 & Pod{\'i}vej & Look \\
17 & jasn{\'y} & clear/ok \\
11 & Odpov\v{e}\v{d} & Answer\\
\p{}8 & str{\'y}c & uncle \\
\p{}7 & Pozn{\'a}mky & Notes \\
\bottomrule
\end{tabularx}
\end{center}
\label{tab:parent_words_in_child_output}
\end{table}

We evaluate the translated sentences both automatically as well as manually. The automatic evaluation is the same as in the previous analysis and identified 54 Czech sentences. We manually checked these sentences and found out that a few of them are Czech postal addresses or other named entities, the rest are Estonian sentences with Czech words inside them. The longest Czech sequence without named entities is ``u v\v{s}ech \v{c}ert'' (an incomplete idiom ``all the hell''). However, we noticed that the Czech words are often already present in the source sentence.\footnote{The \pair{Czech}{English} training corpus is noisy, as we discussed in \cref{sec:czeng}.}

In order to analyze the size of Czech words produced by the child model, we listed all sentences that use Czech characters not contained in Estonian, e.g. characters with diacritics. Then we removed sentences, where the source also contained Czech-only characters. This way, we got 624 out of 100k sentences, which often contained only one word that we manually identified as Czech. Altogether these sentences contained 645 words with Czech-only characters (the evaluated dataset contained 1.4M words). We list few of the most frequently appearing in \cref{tab:parent_words_in_child_output}.

Despite that we evaluated the child model with parent training data, i.e. the corpus that the model could have memorized, we found only a few parent relics. Therefore, we conclude that transfer learning is not negatively affected by the parent model.

\observation{Relic words (i.e. words from the parent target language) are very rare in \ac{NMT} transfer learning.}{obs:parent_relics_are_infrequent}

The rapid change in behavior, when only 3000 steps are enough to forget the parent target language, is one of the results of ``catastrophic forgetting'', a nature of a network to quickly forget or re-train previously learned features. This phenomenon has been widely studied \parcite{kirkpatrick2017overcoming,kemker2018measuring} as the researchers develop methods to overcome this issue. Furthermore, we have tackled problems connected with catastrophic forgetting in \persciteA{kocmi2017curriculum} when experimenting with curriculum learning \parcite{bengio2009curriculum}.

Despite catastrophic forgetting being a problem in machine learning in general, in the scenario of transfer learning, we believe it helps to avoid negative transfer from the parent model by forgetting it. However, we need to keep in mind that in future, when algorithms become more robust in terms of catastrophic forgetting, the negative transfer could emerge as a problem for transfer learning because we saw some words attributed to the parent language pair.

\subsection{Extremely Low-Resources Language Pairs}
\label{sec:extremely_low_resource_transfer}

We showed that \ac{NMT} systems quickly forget the parent language in the low-resource scenario. Now, we evaluate an \emph{extremely} low-resource language pair to find out, if our approach helps in the extremely low-resource scenario (see \cref{sec:lowresource_definition}) or if insufficient data lead to negative transfer as \perscite{wang2019characterizing} described. The results in this subsection are from our paper \persciteA{kocmi2018trivial} (\TFA{}).

%Although this thesis focuses on low-resource language pairs with the number of parallel sentences in order of a hundred thousand. We evaluate our approaches also on extremely low-resource language pairs with tens of thousands of sentences to show the applicability of our approach. The results are for warm-start transfer learning from our paper \persciteA{kocmi2018trivial}.

We simulate extremely low-resource settings by downscaling the data for the child model but maintaining the same parent model. It is a common knowledge that gains from transfer learning are more pronounced for child models with smaller training data. We use the \transl{English}{Finnish} as a parent model for \transl{English}{Estonian}. We mention that shared-source is the harder transfer scenario as the model cannot benefit from the parent English language model because the target language changes from Finnish to Estonian (more analysis in \cref{sec:shared_decoder_is_easier}).

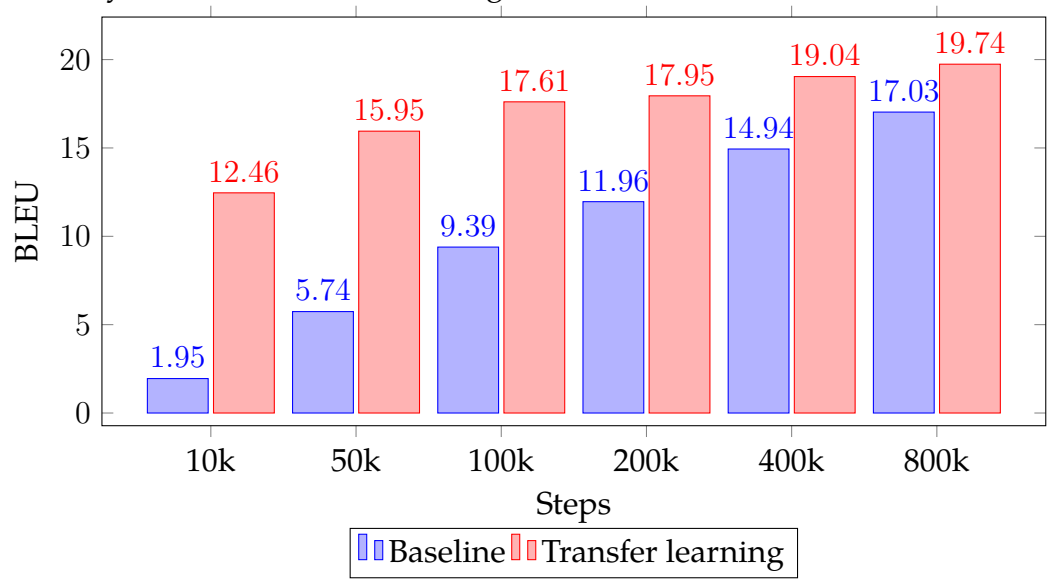
\begin{figure}
\caption{Maximal score reached by \transl{English}{Estonian} child for decreasing sizes of child training data, trained off an \transl{English}{Finnish}  parent. The baselines use only the reduced \pair{Estonian}{English} data.
}
\begin{tikzpicture}
\begin{axis}[
    ybar,
    bar width=.8cm,
    width=14cm,
    height=7cm,
    enlargelimits=0.15,
    legend style={at={(0.5,-0.25)},
      anchor=north,legend columns=-1},
    ylabel={BLEU},
    xlabel={Steps},
    symbolic x coords={10k,50k,100k,200k,400k,800k},
    xtick=data,
    nodes near coords,
    nodes near coords align={vertical},
    ]
\addplot coordinates {(10k,1.95) (50k,5.74) (100k,9.39) (200k,11.96) (400k,14.94) (800k,17.03)};
\addplot coordinates {(10k,12.46) (50k,15.95) (100k,17.61) (200k,17.95) (400k,19.04) (800k,19.74)};
\legend{Baseline, Transfer learning}
\end{axis}
\end{tikzpicture}
\label{tab:simulated_lowresource}
\end{figure}

The results of downscaling the child training corpus are shown in \cref{tab:simulated_lowresource}. We see that our approach applies even to extremely low-resource language pairs with as few as 10k sentence pairs. We see this behavior on the 10k training corpus, where the baseline reaches 1.95 BLEU. This behavior is in accordance with observations done by \perscite{koehn2017six}. For such a small amount of training data, the \ac{NMT} baseline cannot be properly trained. With transfer learning, \ac{NMT} suddenly becomes able to train the model and reaches 12.46 BLEU.

\observation{Transfer learning helps \ac{NMT} to train models for extremely low-resource language pairs that are not possible to properly train on their own.}{obs:transfer_works_in_extremely_lowresource}

\perscite{sennrich2019revisiting} recently revisited the problem of extremely low-resource language pairs and showed that it could be tackled with various tricks. Furthermore, transfer learning could be used as another way of improving the extremely low-resource language pairs hand in hand with other techniques mentioned by \perscite{sennrich2019revisiting}.

As \perscite{wang2019characterizing} summarized, transfer learning can lower the performance of the child task whenever the amount of child training data is too low. We showed that this is not the case in \ac{NMT} because transfer learning can help training the model even when the baseline cannot be trained in the first place.

\subsection{Low-Resource Parent Language Pair}
\label{sec:low_resource_parent}

We showed that transfer learning is not restricted to the low-resource scenarios and improves the performance even when the child is a high-resource language pair (see \cref{obs:coldstart_improves_highresource} and \cref{obs:transfer_improves_also_highresource}). However, it is in the scenario, where both the parent and child model are high-resource.

\perscite{wang2019characterizing} summarized that transfer learning is negatively influenced by the parent model whenever the parent has a low number of training examples. In this section, we examine this condition in the area of \ac{NMT}, and we investigate the scenario where the parent has a lower amount of parallel sentences than the child model.

\perscite{zoph2016transferLowResource} conclude that the relatedness of languages is the main factor influencing the success of transfer learning, which we already showed it is not a necessary condition in \cref{chap:transfer_learning}. However, we use linguistically related languages in this experiment, because the secondary goal is to test what plays a bigger role in transfer learning -- the relatedness of languages or the size of parent data.

\begin{table}[t]
\caption{The column ``Transfer'' is our warm-start method, baselines correspond to training on child corpus only. We show the sizes of corpora in millions sentences. The \significantmark{} represents significantly better results.}
\begin{center}
\begin{tabularx}{\textwidth}{@{}llll@{\hskip 1cm}CC@{}}
\toprule
Parent & Size & Child & Size & Transfer & Baseline\\
\midrule
\transl{\EN{}}{Estonian} & 0.8M & \transl{\EN{}}{Finnish} & \p{}2.8M & \textbf{20.07}\significant{} & 19.50 \\
\transl{Estonian}{\EN{}} & 0.8M &  \transl{Finnish}{\EN{}} & \p{}2.8M & 23.95 & \textbf{24.40}\\
\transl{\EN{}}{Slovak} & 4.3M & \transl{\EN{}}{Czech} & 40.1M & 22.99 & \textbf{23.48}\significant{} \\
\transl{Slovak}{\EN{}} & 4.3M & \transl{Czech}{\EN{}} & 40.1M & 28.20 & \textbf{29.61}\significant{} \\
\bottomrule
\end{tabularx}
\end{center}
\label{tab:low_resource_parent}
\end{table}

Results from \persciteA{kocmi2018trivial} (\TFA{}) are presented in \cref{tab:low_resource_parent}. We see that low-resource parents do not generally improve the performance of sufficiently resourced children. The only exception is the child \transl{English}{Finnish}, where the child has only 3.5 times more parallel sentences than the \transl{English}{Estonian} parent. 

\observation{Transfer learning harms the child performance whenever the parent has substantially less training data than the child.}{obs:lowresource_parent_hurt_performance}

Whenever the child has notably more training data, e.g. ten times more for \pair{Czech}{English} it even (significantly) decreases the child's performance compared to the baseline. Therefore we conclude that transfer learning is negatively influenced in scenarios where the parent has substantially less training data. We suppose it could be due to the initial warm-up steps when the network changes rapidly, thus low-resource language can skew it. However, more analysis is needed to study this behavior properly.

Furthermore, we evaluated linguistically related languages where the relatedness could help improve the model. For example the Czech and Slovak are related to such extent that evaluating \transl{English}{Czech} system on \transl{English}{Slovak} testset output leads to 6.51 BLEU (see \cref{sec:broad_evaluation}). However, the relatedness did seem not to play any role in our experiments, and transfer learning led to worse performance than training on child parallel corpus only.

\observation{For a high-resource child the linguistic relatedness of parent and child language pairs is less important than the size of the parent training corpus.}{obs:related_lowresource_parent_doesnot_work}

\subsection{No-Shared Language Scenario}
\label{sec:no_language_in_common}

One of the main factors of negative transfer is the divergence in distributions between parent and child training data \parcite{wang2019characterizing}. In \ac{NMT}, one would assume that the languages in question are the key element affecting task similarity.

In \cref{sec:broad_evaluation}, we showed that the relatedness of the languages is not the most critical for transfer learning. Moreover, the related \transl{English}{Finnish} parent performed worse even when compared to a parent that uses a different writing script, in our case, \transl{English}{Russian} with Cyrillic. Thus we have not detected any negative transfer when evaluated on less linguistically related languages.

However, our experiments always contained a language shared between the parent and child, e.g. English, which could work as a connecting bridge during transfer learning, thus preventing the negative effects. In order to test the negative transfer in \ac{NMT}, we experiment with a no-shared language scenario.

We examine the performance of \transl{Estonian}{English} child trained on top of parents using unrelated languages, specifically \transl{Arabic}{Russian}, \transl{Spanish}{French}, \transl{Spanish}{Russian}, and \transl{French}{Russian}. The parents are trained with the UN corpus \parcite{ziemski2016united}, which has 10M multi-parallel sentences across six languages. 

\begin{table}[t]
\caption{No-shared language scenario of transfer learning. The child model is \transl{Estonian}{English}. Each row represents various metrics for measuring \ac{MT} performance, where higher number is better for all metrics. The significance \significantmark{} is computed pairwise relative to the baseline ``No transfer learning''.}
\begin{center}
\begin{tabularx}{\textwidth}{@{}lCCCHCCc@{}}
\toprule
Parent &  BLEU  & nPER & nTER & nCDER & nWER & chrF3 & nCharacTER\\
\midrule
No transfer learning     & 21.74 & 54.33 & 35.66 & 41.28 & 32.70 & 49.87 & 37.70\\
\transl{Arabic}{Russian} & 22.23 & 55.05 & 36.66 & 42.23 & 33.59 & 50.86 & \textbf{40.11}\\
\transl{Spanish}{French} & 22.24\significant{} & \textbf{55.32} & 36.58 & 42.05 & 33.69 & 50.88 & 39.59 \\
\transl{Spanish}{Russian} & \textbf{22.52}\significant{} & 55.26 & \textbf{36.85} & \textbf{42.53} & \textbf{33.79} & \textbf{51.28} & 39.92 \\
\transl{French}{Russian} & 22.40\significant{} & 54.99 & 36.50 & 42.06 & 33.39 & 50.93 & 39.60 \\
\bottomrule
\end{tabularx}
\end{center}
\label{tab:no_shared}
\end{table}

The results from \persciteA{kocmi2018trivial} (\TFA{}) are shown in \cref{tab:no_shared}. We see mostly significant gains from transfer learning in all cases. The only non-significant gain is from \transl{Arabic}{Russian}, which does not share the script with the child's Latin at all, only sharing of punctuation and numbers is possible across all the tested scripts.

\observation{Transfer learning improves the performance even for the no-shared language scenario.}{obs:transfer_works_when_no_language_shared}

There is no loss of performance in comparison to the baseline. This can be seen either as the evidence that transfer learning is not negatively affected by the difference in data distributions between parent and child, or that the mere distributional properties of all (tested) languages are sufficiently similar to be useful for transfer learning in \ac{NMT}.

Surprisingly, the \transl{Spanish}{Russian} (with a target languages that uses the Cyrillic script) reached a better performance than the \transl{Spanish}{French}, a target language that is linguistically closest to the \transl{Estonian}{English} from all four investigated language pairs. However, neither of these two systems is significantly better than the other.
Furthermore, the gains are quite similar (+0.49 up to +0.78 BLEU), which supports an assumption that the major factor influencing transfer learning is the size of the parent (here, all parents have 10M sentence pairs). We are going to discuss this aspect in \cref{tab:datasize_vs_relatedness}. This result can also be explained with a similar domain of parent training set. In comparison, the \transl{Czech}{English} parent, which has 40.1M sentences from a broader range of domains and has a shared language (English), improved the performance of \transl{Estonian}{English} by 3.38 BLEU.

\observation{The exact parent language pair does not seem to affect the performance given a particular domain and parent data size.}{obs:any_parent_performs_similarly_when_same_amount_of_data}

In the no-shared language scenario, the gains cannot be attributed to the language model or model parts such as shared English word embeddings. The subword vocabulary overlap is mostly due to short subwords or numbers and punctuation.

In \cref{sec:automatic_metrics}, we discuss issues with the BLEU metrics, e.g. ignoring the importance of various $n$-grams or high-influence of tokenization. For this reason, we computed the scores for several other automatic methods. We see that in all metrics, transferred models perform better than the baseline.

The experiments presented in this section indicate that the parent simply works as a better model weight initialization in comparison to the random initialization. We investigate it more in \cref{sec:linguistic_vs_initialization}.

% In multi-task learning, \perscite{firat2016multiway} hinted possible gains even when the source and target languages are distinct from the low-resource languages. We examine the effect of using a parent language pair that does not share either of languages with the child language pair.

\subsection{Conclusion on Negative Transfer}
\label{sec:conclusion_on_negative_transfer}

In this section, we investigated the effect of negative learning. We evaluated various scenarios and identified two weaknesses of transfer learning. We showed that the final child model could exhibit relics of the parent target language in \cref{sec:traces_of_parent_language_pair}. Moreover, in \cref{sec:low_resource_parent}, we showed that when the parent has substantially less training data than the child, transfer learning may hurt the final performance. In this scenario, the training model without transfer learning performs better.

Lastly, we reevaluate three critical factors proposed by \perscite{wang2019characterizing} that influence the negative transfer:

\begin{enumerate}
  \item Bigger divergence between the distributions of both tasks is hurting transfer learning.
  \item Effectiveness of transfer learning depends on the size of child data.
  \item Negative transfer should be evaluated with the same setting of the neural network, to avoid adding risk from a different network.
\end{enumerate}

We showed that divergence in distribution does not play a negative role in \ac{NMT} transfer learning because even the no-shared language scenario improves child performance (see \cref{sec:no_language_in_common}).

The second factor plays a role only in the case when the parent has less training data in comparison to the child (see \cref{sec:low_resource_parent}).

We used an identical parameter setting for both baseline and transfer learning. Thus the third factor does not play a role in our evaluation.

In conclusion, we provided an analysis of the negative effects of transfer learning and discovered two possible issues of transfer learning. We are not aware of any previous study discussing negative transfer in \ac{NMT} transfer learning.

\section{Does Position of Shared Language Influence Transfer Learning?}
\label{sec:shared_decoder_is_easier}
% \XX{Ondra: zvazit jestli dat pred negativ transfer ... Kocmi: asi spis ne, prijde mi ze negative transfer je samostatnejsi nez zbyvajici kapitoly}

We noticed that transfer learning has different behavior for the shared-source and shared-target language scenarios. For example, \transl{English}{Russian} parent improved the Estonian child more than \transl{English}{Finnish} (20.41 vs. 19.74 BLEU), however in the opposite direction \transl{Russian}{English} worked as a worse parent than \transl{Finnish}{English} (23.54 vs. 24.18 BLEU), as we showed in \cref{sec:broad_evaluation}.

In this section, we investigate the influence of the position of the shared language on transfer learning. Moreover, we discuss which of those tasks is harder for the \ac{NMT} transfer learning.

\subsection{Shared Language Position Effect on Convergence Speed}
\label{sec:shared_decoder_converges_faster}

We start with an investigation of the training time needed for each direction to converge.
We recall the results from \cref{sec:warm_cold_comparison} (\TFB{}) in \cref{tab:shared_decoder_converges_faster}. We also added the \pair{Gujarati}{English} pair from our paper \persciteA{kocmi2019wmt} (\TFC{}).
We subtracted the number of parent steps needed for the convergence and showed the results of the child model training in \cref{tab:shared_decoder_converges_faster}.\footnote{Training steps in \cref{sec:warm_cold_comparison} are the sum of parent plus child training steps, but we show only the child number of steps in \cref{tab:shared_decoder_converges_faster}.}

\begin{table}[t]
\caption{Number of steps needed for a model to converge. The size shows the number of sentences in the corpora of each language. For both child the second language is English. The results are from \cref{sec:warm_cold_comparison} with subtracted time of parent model training. }
\begin{center}
\begin{tabularx}{\textwidth}{@{}lCCC@{}}
\toprule
Child language  & Size & Shared-source & Shared-target \\
\midrule               
Estonian & \p{}0.8M & \p{}75k & \p{}25k  \\
Gujarati & \p{}0.2M & 195k & 180k \\
Russian & 12.6M & 935k & 790k \\
\bottomrule
\end{tabularx}
\end{center}
\label{tab:shared_decoder_converges_faster}
\end{table}

In \cref{tab:shared_decoder_converges_faster} we can see that the shared-target language scenario converges faster for both the low-resource Estonian and the high-resource Russian. In the case of \pair{Gujarati}{English}, the convergence is only slightly faster for the shared-target. 

\observation{Transfer learning with a shared-target language converges in fewer steps than with a shared-source language.}{obs:shared_target_converges_faster}

We need to mention that the total number of training steps does not always reflect the convergence speed because the performance usually fluctuates and thus the model can converge after a different number of steps, which depends on randomness in training. Moreover, the training time does not explicitly mean if the task is easier; there are many other factors like shared language, the noisiness of training data, and other factors.

Therefore we investigate the learning curves and look for a distinctive behavior between these tasks.

\subsection{Shared Language Position Affects Slope of Learning Curve}
\label{sec:shared_decoder_has_higher_slope}

In \cref{fig:comparison_transfer_non}, we described three impacts of transfer learning on the learning curve -- namely, higher start, higher performance, and higher slope. The effect of higher slope suggests that the model trains faster, therefore we compare learning curves of shared-source and shared-target scenarios in order to find out which one learns faster.

We report the learning curve's Y-axis in BLEU, but any other metric could be used. The BLEU score has a disadvantage that it cannot be compared across various languages or even testsets. Therefore, in order to study the slope of the learning curve, we need to scale it. We scale each learning curves by multiplicating the performance (BLEU) by a fixed constant. The constant is selected in order to align the best-reached performance of both translation directions.

\begin{figure}
\caption{Learning curves for various language pairs in both directions. The Y-axis has been scaled for each learning curve by a constant in order to match their final performance. The bracket specifies the child's second language that is paired with English. The convergence is seen only on Gujarati pair as other languages converged later than within first 30k steps.}
\begin{center}
\begin{tikzpicture}[gnuplot]
%% generated with GNUPLOT 5.0p3 (Lua 5.1; terminal rev. 99, script rev. 100)
%% Fri 27 Sep 2019 03:06:16 PM DST
\path (0.000,0.000) rectangle (4.500,7.400);
\gpcolor{color=gp lt color border}
\gpsetlinetype{gp lt border}
\gpsetdashtype{gp dt solid}
\gpsetlinewidth{1.00}
\draw[gp path] (0.952,0.985)--(1.132,0.985);
\draw[gp path] (12.945,0.985)--(12.765,0.985);
\node[gp node right] at (0.768,0.985) { };
\draw[gp path] (0.952,1.741)--(1.132,1.741);
\draw[gp path] (12.945,1.741)--(12.765,1.741);
\node[gp node right] at (0.768,1.741) { };
\draw[gp path] (0.952,2.497)--(1.132,2.497);
\draw[gp path] (12.945,2.497)--(12.765,2.497);
\node[gp node right] at (0.768,2.497) { };
\draw[gp path] (0.952,3.252)--(1.132,3.252);
\draw[gp path] (12.945,3.252)--(12.765,3.252);
\node[gp node right] at (0.768,3.252) { };
\draw[gp path] (0.952,4.008)--(1.132,4.008);
\draw[gp path] (12.945,4.008)--(12.765,4.008);
\node[gp node right] at (0.768,4.008) { };
\draw[gp path] (0.952,4.764)--(1.132,4.764);
\draw[gp path] (12.945,4.764)--(12.765,4.764);
\node[gp node right] at (0.768,4.764) { };
\draw[gp path] (0.952,5.520)--(1.132,5.520);
\draw[gp path] (12.945,5.520)--(12.765,5.520);
\node[gp node right] at (0.768,5.520) { };
\draw[gp path] (0.952,6.275)--(1.132,6.275);
\draw[gp path] (12.945,6.275)--(12.765,6.275);
\node[gp node right] at (0.768,6.275) { };
\draw[gp path] (0.952,7.031)--(1.132,7.031);
\draw[gp path] (12.945,7.031)--(12.765,7.031);
\node[gp node right] at (0.768,7.031) { };
\draw[gp path] (0.952,0.985)--(0.952,1.165);
\draw[gp path] (0.952,7.031)--(0.952,6.851);
\node[gp node center] at (0.952,0.677) {$0$};
\draw[gp path] (2.951,0.985)--(2.951,1.165);
\draw[gp path] (2.951,7.031)--(2.951,6.851);
\node[gp node center] at (2.951,0.677) {$5$};
\draw[gp path] (4.950,0.985)--(4.950,1.165);
\draw[gp path] (4.950,7.031)--(4.950,6.851);
\node[gp node center] at (4.950,0.677) {$10$};
\draw[gp path] (6.949,0.985)--(6.949,1.165);
\draw[gp path] (6.949,7.031)--(6.949,6.851);
\node[gp node center] at (6.949,0.677) {$15$};
\draw[gp path] (8.947,0.985)--(8.947,1.165);
\draw[gp path] (8.947,7.031)--(8.947,6.851);
\node[gp node center] at (8.947,0.677) {$20$};
\draw[gp path] (10.946,0.985)--(10.946,1.165);
\draw[gp path] (10.946,7.031)--(10.946,6.851);
\node[gp node center] at (10.946,0.677) {$25$};
\draw[gp path] (12.945,0.985)--(12.945,1.165);
\draw[gp path] (12.945,7.031)--(12.945,6.851);
\node[gp node center] at (12.945,0.677) {$30$};
\draw[gp path] (0.952,7.031)--(0.952,0.985)--(12.945,0.985)--(12.945,7.031)--cycle;
\node[gp node center,rotate=-270] at (0.246,4.008) {Scaled BLEU};
\node[gp node center] at (6.948,0.215) {Child's Training Steps (in thousands)};
\node[gp node right,font={\fontsize{8pt}{9.6pt}\selectfont}] at (11.699,2.518) {Shared target (Russian)};
\gpcolor{rgb color={0.000,1.000,0.000}}
\gpsetlinewidth{4.00}
\draw[gp path] (11.846,2.518)--(12.614,2.518);
\draw[gp path] (0.952,1.127)--(2.951,5.694)--(4.950,6.061)--(6.949,6.242)--(8.947,6.305)%
  --(10.946,6.455)--(12.945,6.479);
\gpcolor{color=gp lt color border}
\node[gp node right,font={\fontsize{8pt}{9.6pt}\selectfont}] at (11.699,2.272) {Shared source (Russian)};
\gpcolor{rgb color={0.000,1.000,0.000}}
\gpsetlinetype{gp lt axes}
\gpsetdashtype{gp dt axes}
\draw[gp path] (11.846,2.272)--(12.614,2.272);
\draw[gp path] (0.952,1.231)--(2.951,4.963)--(4.950,5.775)--(6.949,6.075)--(8.947,6.316)%
  --(10.946,6.529)--(12.945,6.542);
\gpcolor{color=gp lt color border}
\node[gp node right,font={\fontsize{8pt}{9.6pt}\selectfont}] at (11.699,2.026) {Shared target (Estonian)};
\gpcolor{rgb color={0.933,0.510,0.933}}
\gpsetlinetype{gp lt border}
\draw[gp path] (11.846,2.026)--(12.614,2.026);
\draw[gp path] (0.952,1.176)--(1.352,3.496)--(1.752,3.983)--(2.151,4.318)--(2.551,4.500)%
  --(2.951,4.616)--(3.351,4.800)--(3.750,4.831)--(4.150,4.880)--(4.550,5.009)--(4.950,5.023)%
  --(5.349,5.050)--(5.749,5.101)--(6.149,5.152)--(6.549,5.130)--(6.949,5.145)--(7.348,5.181)%
  --(7.748,5.171)--(8.148,5.239)--(8.548,5.271)--(8.947,5.242)--(9.347,5.234)--(9.747,5.208)%
  --(10.147,5.272)--(10.546,5.270)--(10.946,5.336)--(11.346,5.364)--(11.746,5.280)--(12.145,5.315)%
  --(12.545,5.274)--(12.945,5.295);
\gpcolor{color=gp lt color border}
\node[gp node right,font={\fontsize{8pt}{9.6pt}\selectfont}] at (11.699,1.780) {Shared source (Estonian)};
\gpcolor{rgb color={0.933,0.510,0.933}}
\gpsetlinetype{gp lt axes}
\gpsetdashtype{gp dt axes}
\draw[gp path] (11.846,1.780)--(12.614,1.780);
\draw[gp path] (0.952,1.267)--(1.352,2.575)--(1.752,3.464)--(2.151,3.955)--(2.551,4.233)%
  --(2.951,4.395)--(3.351,4.601)--(3.750,4.763)--(4.150,4.834)--(4.550,4.872)--(4.950,4.910)%
  --(5.349,4.930)--(5.749,5.023)--(6.149,4.960)--(6.549,5.109)--(6.949,5.113)--(7.348,5.082)%
  --(7.748,5.208)--(8.148,5.162)--(8.548,5.141)--(8.947,5.222)--(9.347,5.250)--(9.747,5.222)%
  --(10.147,5.230)--(10.546,5.295)--(10.946,5.258)--(11.346,5.298)--(11.746,5.279)--(12.145,5.296)%
  --(12.545,5.351)--(12.945,5.306);
\gpcolor{color=gp lt color border}
\node[gp node right,font={\fontsize{8pt}{9.6pt}\selectfont}] at (11.699,1.534) {Shared target (Gujarati)};
\gpcolor{rgb color={1.000,0.000,0.000}}
\gpsetlinetype{gp lt border}
\draw[gp path] (11.846,1.534)--(12.614,1.534);
\draw[gp path] (0.952,1.023)--(2.951,2.748)--(4.950,3.751)--(6.949,3.993)--(8.947,3.909)%
  --(10.946,3.977)--(12.945,4.078);
\gpcolor{color=gp lt color border}
\node[gp node right,font={\fontsize{8pt}{9.6pt}\selectfont}] at (11.699,1.288) {Shared source (Gujarati)};
\gpcolor{rgb color={1.000,0.000,0.000}}
\gpsetlinetype{gp lt axes}
\gpsetdashtype{gp dt axes}
\draw[gp path] (11.846,1.288)--(12.614,1.288);
\draw[gp path] (0.952,1.265)--(2.951,2.154)--(4.950,3.055)--(6.949,3.570)--(8.947,3.772)%
  --(10.946,4.024)--(12.945,4.044);
\gpcolor{color=gp lt color border}
\gpsetlinetype{gp lt border}
\gpsetdashtype{gp dt solid}
\gpsetlinewidth{1.00}
\draw[gp path] (0.952,7.031)--(0.952,0.985)--(12.945,0.985)--(12.945,7.031)--cycle;
%% coordinates of the plot area
\gpdefrectangularnode{gp plot 1}{\pgfpoint{0.952cm}{0.985cm}}{\pgfpoint{12.945cm}{7.031cm}}
\end{tikzpicture}
%% gnuplot variables
\end{center}
\label{fig:higher_slope_of_shared_decoder}
\end{figure}
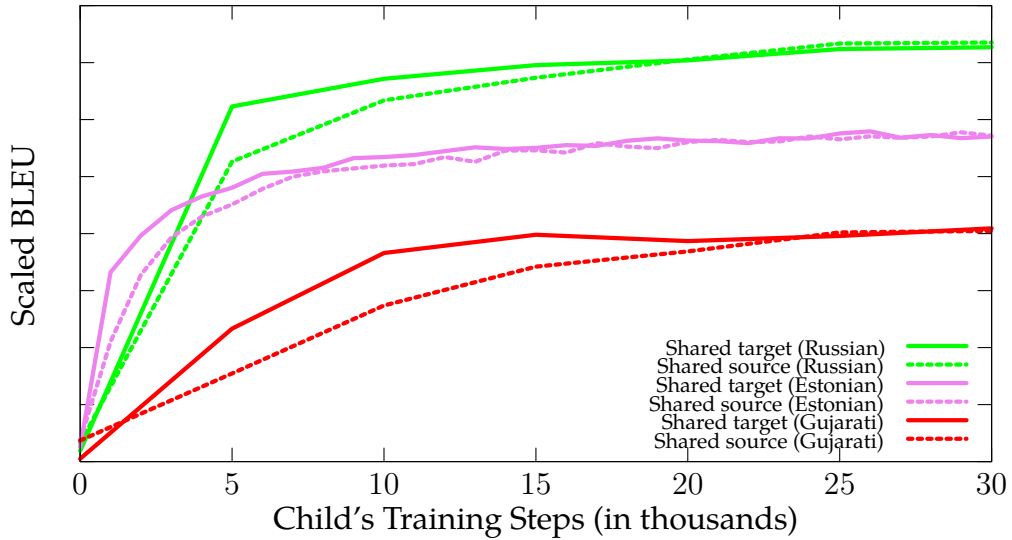

\cref{fig:higher_slope_of_shared_decoder} presents the learning curves of three language pairs evaluated in the previous section on their respective development sets. We investigate only the first 30k training steps, where the difference in slopes is most visible.

When comparing the learning curves of shared-source and shared-target scenarios, we see that for all three language pairs, the shared-target has a higher slope than the shared-source. 

\observation{Transfer learning with a shared-target language has a higher slope of the learning curve.}{obs:shared_target_have_higher_slope}

This observation suggests that shared-target, i.e. having shared language (e.g. English) on the target side, is easier for transferring knowledge from parent to child. In this scenario, \ac{NN} reaches higher performance in a shorter time compared to the shared-source scenario.
This behavior is not surprising. From the neural network's point of view, it is learning to predict the shared language through the whole training process. Therefore it can utilize the language model from a parent with only learning different encoder's part of the model. We further analyze the behavior by freezing various parts of a model in \cref{sec:linguistic_vs_initialization}.

\subsection{Parent Performance Drop}
\label{sec:parent_performance_dropping}

In transfer learning, we do not pay attention to the final parent's performance as is customary in multi-task learning. In \cref{sec:traces_of_parent_language_pair}, we showed that for translation direction with shared-source, the child model quickly forgets the parent target language.

In this section, we investigate how the parent translation deteriorates during the child's training phase in both directions, and if the shared-source and shared-target scenarios behave differently. 

We evaluate two scenarios, the shared-source and shared-target under our \transl{English}{Gujarati} and \transl{Gujarati}{English} child models transferred from \pair{Czech}{English} parent \parciteA{kocmi2019wmt} (\TFC{}).

We evaluate both models each by the corresponding child's and parent's development set. The scores between language pairs are not comparable as they are for different languages.

\begin{figure}
\caption{Performance of child model on a parent development set. Both child are \pair{Gujarati}{English}. Learning curves are evaluated on a development set of analogous language pairs.}
\begin{center}
\begin{tikzpicture}[gnuplot]
%% generated with GNUPLOT 5.0p3 (Lua 5.1; terminal rev. 99, script rev. 100)
%% Sun 11 Aug 2019 02:13:18 PM DST
\path (0.000,0.000) rectangle (4.500,7.400);
\gpcolor{color=gp lt color border}
\gpsetlinetype{gp lt border}
\gpsetdashtype{gp dt solid}
\gpsetlinewidth{1.00}
\draw[gp path] (1.136,0.985)--(1.316,0.985);
\draw[gp path] (12.945,0.985)--(12.765,0.985);
\node[gp node right] at (0.952,0.985) {$0$};
\draw[gp path] (1.136,2.497)--(1.316,2.497);
\draw[gp path] (12.945,2.497)--(12.765,2.497);
\node[gp node right] at (0.952,2.497) {$5$};
\draw[gp path] (1.136,4.008)--(1.316,4.008);
\draw[gp path] (12.945,4.008)--(12.765,4.008);
\node[gp node right] at (0.952,4.008) {$10$};
\draw[gp path] (1.136,5.520)--(1.316,5.520);
\draw[gp path] (12.945,5.520)--(12.765,5.520);
\node[gp node right] at (0.952,5.520) {$15$};
\draw[gp path] (1.136,7.031)--(1.316,7.031);
\draw[gp path] (12.945,7.031)--(12.765,7.031);
\node[gp node right] at (0.952,7.031) {$20$};
\draw[gp path] (1.136,0.985)--(1.136,1.165);
\draw[gp path] (1.136,7.031)--(1.136,6.851);
\node[gp node center] at (1.136,0.677) {$2000$};
\draw[gp path] (4.088,0.985)--(4.088,1.165);
\draw[gp path] (4.088,7.031)--(4.088,6.851);
\node[gp node center] at (4.088,0.677) {$2100$};
\draw[gp path] (7.041,0.985)--(7.041,1.165);
\draw[gp path] (7.041,7.031)--(7.041,6.851);
\node[gp node center] at (7.041,0.677) {$2200$};
\draw[gp path] (9.993,0.985)--(9.993,1.165);
\draw[gp path] (9.993,7.031)--(9.993,6.851);
\node[gp node center] at (9.993,0.677) {$2300$};
\draw[gp path] (12.945,0.985)--(12.945,1.165);
\draw[gp path] (12.945,7.031)--(12.945,6.851);
\node[gp node center] at (12.945,0.677) {$2400$};
\draw[gp path] (1.136,7.031)--(1.136,0.985)--(12.945,0.985)--(12.945,7.031)--cycle;
\node[gp node center,rotate=-270] at (0.246,4.008) {BLEU};
\node[gp node center] at (7.040,0.215) {Steps (in thousands)};
\node[gp node right,font={\fontsize{8pt}{9.6pt}\selectfont}] at (11.699,2.026) {Shared target (Parent devset)};
\gpcolor{rgb color={1.000,0.000,0.000}}
\gpsetlinetype{gp lt axes}
\gpsetdashtype{gp dt axes}
\gpsetlinewidth{4.00}
\draw[gp path] (11.846,2.026)--(12.614,2.026);
\draw[gp path] (3.556,7.031)--(3.645,7.017)--(3.793,6.928)--(4.236,6.850)--(4.383,6.848)%
  --(4.679,6.765)--(4.974,6.729)--(5.269,6.593)--(5.417,6.654)--(5.564,6.538)--(5.712,6.458)%
  --(5.860,6.311)--(6.007,6.451)--(6.155,6.412)--(6.302,6.397)--(6.450,6.307)--(6.598,6.226)%
  --(6.745,6.293)--(6.893,6.222)--(7.041,6.241)--(7.188,6.087)--(7.336,6.095)--(7.483,6.099)%
  --(7.631,6.153)--(7.779,6.088)--(7.926,6.072)--(8.221,6.103)--(8.369,5.907)--(8.517,5.955)%
  --(8.664,6.013)--(8.812,5.899)--(8.959,5.837)--(9.107,5.838)--(9.255,5.655)--(9.402,5.804)%
  --(9.550,5.701)--(9.698,5.812)--(9.845,5.656)--(9.993,5.610)--(10.140,5.640)--(10.288,5.683)%
  --(10.436,5.638)--(10.583,5.604)--(10.731,5.575)--(10.878,5.584)--(11.026,5.518)--(11.174,5.320)%
  --(11.469,5.382)--(11.764,5.342)--(11.912,5.300)--(12.059,5.274)--(12.502,5.357)--(12.650,5.200)%
  --(12.797,5.180)--(12.945,5.173);
\gpcolor{color=gp lt color border}
\node[gp node right,font={\fontsize{8pt}{9.6pt}\selectfont}] at (11.699,1.780) {Shared target (Child devset)};
\gpcolor{rgb color={0.545,0.000,0.000}}
\draw[gp path] (11.846,1.780)--(12.614,1.780);
\draw[gp path] (1.136,1.014)--(1.284,2.341)--(1.431,3.113)--(1.579,3.299)--(1.726,3.234)%
  --(1.874,3.287)--(2.022,3.364)--(2.169,3.491)--(2.317,3.387)--(2.465,3.360)--(2.612,3.423)%
  --(2.760,3.555)--(2.907,3.531)--(3.055,3.738)--(3.203,3.577)--(3.350,3.719)--(3.498,3.713)%
  --(3.645,3.719)--(3.793,3.759)--(3.941,3.742)--(4.088,3.818)--(4.236,3.689)--(4.383,3.745)%
  --(4.531,3.651)--(4.679,3.621)--(4.826,3.733)--(4.974,3.798)--(5.122,3.754)--(5.269,3.795)%
  --(5.417,3.789)--(5.564,3.745)--(5.712,3.694)--(5.860,3.690)--(6.007,3.704)--(6.155,3.771)%
  --(6.302,3.793)--(6.450,3.848)--(6.598,3.766)--(6.745,3.812)--(6.893,3.782)--(7.041,3.816)%
  --(7.188,3.700)--(7.336,3.778)--(7.483,3.808)--(7.631,3.823)--(7.779,3.821)--(7.926,3.813)%
  --(8.074,3.765)--(8.221,3.777)--(8.369,3.714)--(8.517,3.803)--(8.664,3.828)--(8.812,3.835)%
  --(8.959,3.837)--(9.107,3.793)--(9.255,3.802)--(9.402,3.827)--(9.550,3.805)--(9.698,3.829)%
  --(9.845,3.780)--(9.993,3.845)--(10.140,3.743)--(10.288,3.791)--(10.436,3.776)--(10.583,3.753)%
  --(10.731,3.842)--(10.878,3.843)--(11.026,3.845)--(11.174,3.847)--(11.321,3.752)--(11.469,3.744)%
  --(11.616,3.838)--(11.764,3.795)--(11.912,3.830)--(12.059,3.841)--(12.207,3.720)--(12.355,3.800)%
  --(12.502,3.819)--(12.650,3.829)--(12.797,3.746)--(12.945,3.834);
\gpcolor{color=gp lt color border}
\node[gp node right,font={\fontsize{8pt}{9.6pt}\selectfont}] at (11.699,1.534) {Shared source (Parent devset)};
\gpcolor{rgb color={0.000,1.000,0.000}}
\gpsetlinetype{gp lt border}
\gpsetdashtype{gp dt 3}
\draw[gp path] (11.846,1.534)--(12.614,1.534);
\draw[gp path] (1.229,7.031)--(1.579,1.175)--(1.726,1.148)--(2.022,1.136)--(2.169,1.143)%
  --(2.317,1.131)--(2.760,1.127)--(2.907,1.128)--(3.055,1.130)--(3.203,1.132)--(3.350,1.130)%
  --(3.498,1.130)--(3.645,1.129)--(3.793,1.136)--(4.088,1.149)--(4.236,1.139)--(4.383,1.134)%
  --(4.531,1.146)--(4.679,1.128)--(4.974,1.087)--(5.122,1.143)--(5.269,1.144)--(5.417,1.139)%
  --(5.564,1.150)--(5.712,1.139)--(5.860,1.148)--(6.007,1.142)--(6.155,1.137)--(6.302,1.137)%
  --(6.450,1.132)--(6.598,1.125)--(6.745,1.147)--(6.893,1.139)--(7.041,1.141)--(7.188,1.142)%
  --(7.336,1.104)--(7.483,1.137)--(7.631,1.127)--(7.779,1.132)--(7.926,1.160)--(8.074,1.151)%
  --(8.221,1.152);
\gpcolor{color=gp lt color border}
\node[gp node right,font={\fontsize{8pt}{9.6pt}\selectfont}] at (11.699,1.288) {Shared source (Child devset)};
\gpcolor{rgb color={0.000,0.392,0.000}}
\draw[gp path] (11.846,1.288)--(12.614,1.288);
\draw[gp path] (1.136,1.209)--(1.284,1.920)--(1.431,2.641)--(1.579,3.053)--(1.726,3.215)%
  --(1.874,3.416)--(2.022,3.433)--(2.169,3.375)--(2.317,3.465)--(2.465,3.335)--(2.612,3.453)%
  --(2.760,3.352)--(2.907,3.401)--(3.055,3.355)--(3.350,3.400)--(3.645,3.448)--(3.793,3.373)%
  --(3.941,3.444)--(4.088,3.413)--(4.236,3.427)--(4.383,3.390)--(4.531,3.438)--(4.679,3.416)%
  --(4.826,3.463)--(4.974,3.411)--(5.122,3.435)--(5.269,3.463)--(5.417,3.418)--(5.712,3.470)%
  --(6.155,3.394)--(6.302,3.415)--(6.450,3.428)--(6.598,3.499)--(6.893,3.519)--(7.188,3.483)%
  --(7.336,3.447)--(7.483,3.364)--(7.779,3.379)--(7.926,3.424)--(8.074,3.470)--(8.221,3.506);
\gpcolor{color=gp lt color border}
\gpsetdashtype{gp dt solid}
\gpsetlinewidth{1.00}
\draw[gp path] (1.136,7.031)--(1.136,0.985)--(12.945,0.985)--(12.945,7.031)--cycle;
%% coordinates of the plot area
\gpdefrectangularnode{gp plot 1}{\pgfpoint{1.136cm}{0.985cm}}{\pgfpoint{12.945cm}{7.031cm}}
\end{tikzpicture}
%% gnuplot variables
\end{center}
\label{fig:parent_performance_dropping}
\end{figure}
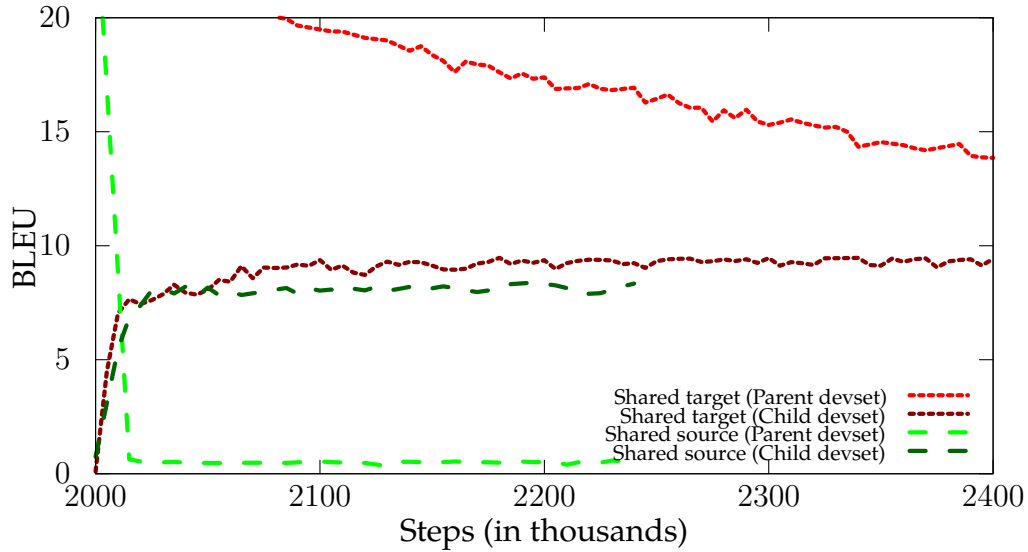

\cref{fig:parent_performance_dropping} presents the results of our experiment. The figure starts at a global step of 2M, the end of the parent models. At this point, all parent models have the final performance higher than 20 BLEU (not visible in the figure). The learning curves with the same dashing correspond to the same model evaluated either on the child or parent development set.

We can see that from the start of the child model's training, the performance of the parent largely deteriorates on the parent's development set and that the behavior for shared-source and the shared-target is distinct. In the scenario with shared-source language, the model does not know to which language it should translate. Therefore it learns to always translate to the child target language.

Interestingly, in the shared-source language scenario, the performance drops nearly immediately. In contrast, whenever we investigate the shared-target scenario, the performance is deteriorating slower, and even after finishing the child's training, it is still able to translate \transl{Czech}{English} parent language pair with ~15 BLEU accuracy.

\observation{Parent performance deteriorates during the child training at different speed depending on shared language position. The shared-source language scenario declines almost instantly. The shared-target language scenario deteriorates slowly.}{obs:parent_deteriorates_at_different_speed}

\cref{fig:parent_performance_dropping} shows that the neural network forgets the parent source language slower. Thus the phenomenon of critical forgetting is mostly concerned with the decoder part. If it forgot both of them in the same way, the drop in parent performance would be similar in both directions, i.e. \transl{English}{XX} and \transl{XX}{English}. 

\observation{A converged child model in the shared-target scenario can still translate the parent language pair to some extent. It is not possible in the shared-source scenario.}{obs:child_with_shared_target_can_perform_parent_task}

This behavior could be a result of an error backpropagation, where the gradient vanishes as it travels back through the network, thus updates the encoder layers less than the decoder layers. This is especially true in transfer learning with shared-target because the network already knows how to generate the target language, e.g. English. Thus it does not produce sufficient errors that would modify the encoder. Thus the encoder does not forget the parent task as quickly. This suggests that the most important part of the model is the decoder and therefore, transfer learning with the shared-target language is easier to learn as it already knows how target language should look like.

As future work, we could increase the error backpropagation to the encoder using ultra-residual connections similar to \perscite{emelin2019widening} that would connect various layers of the encoder directly with the decoder's layers. Moreover, we could add a special tag at the beginning of the source language specifying the desired target language \parcite{johnson2017zeroshot}.

\subsection{Conclusion}

In this section, we investigated if there is a difference between the shared-source and shared-target scenario. In \cref{sec:shared_decoder_converges_faster}, we showed that shared-target converges faster than shared-source, which is associated with a higher slope as demonstrated in \cref{sec:shared_decoder_has_higher_slope}. Furthermore, we revealed that \ac{NMT} forgets the parent model slower in the shared-target scenario in \cref{sec:parent_performance_dropping}.

We cannot compare our findings with other works in the \ac{NMT} transfer learning because we are the first to show transfer learning with a shared-target language. However, we can investigate a closely related task of multi-task \ac{NMT}. 
\perscite{firat2016multiway} trained a multilingual system with six languages and showed that their model outperforms the single-pair baseline only in case of the shared-target language. In the case of the shared-source language scenario, the baseline performs on-par or better than their multilingual system. \perscite{johnson2017zeroshot} showed that the many-to-one scenario (shared-target) leads to improvements in most cases, unlike the one-to-many scenario (shared-source).

In summary, based on previous observations, we conclude that the shared-\errata{target}{source} language scenario is a harder task for transfer learning compared to the shared-\errata{source}{target} language scenario.

Lastly, there are other scenarios that do not fall into our investigated categories. We can have the shared language on a different translation side, e.g. parent of \transl{English}{XX} and child \transl{XX}{English}, which we are going to investigate in \cref{sec:direction_swap}. And we already examined a no-shared language scenario in \cref{sec:no_language_in_common}.

% \ac{NMT} models are trained by backpropagation of an error from the predicted output for a given input sentence. Whenever the predicted output is closer to the correct translation, then a smaller error is backpropagated through the network making smaller changes. 
% In the case of the shared source, model obtains input in the same language as it knows, therefore generates a translation for the parent target language, that is different from the child. 
% In the case of the shared decoder, model obtains input in a different language than used to know. However, it is trained to predict English sentences, thus is initially guess some English sentences.  

\section{Language Relatedness versus Data Size}
\label{sec:more_data_is_more_important}

Whenever humans learn a new language, it is much easier for them if they know a related language. Therefore we suppose that similar works for \ac{NN} with transfer learning. \perscite{zoph2016transferLowResource} in their transfer learning approach concluded that "the choice of parent model can have a strong impact on transfer models, and choosing better [related] parents for our low-resource languages could improve the final results".
Furthermore, the use of related language pair as a way to improve the performance of a model has been widely studied, and researchers showed that related language pairs can be used as a source of improvements \parcite{nakov2009improved, nguyen2017transfer}.

However, in \cref{sec:broad_evaluation}, we saw that \transl{English}{Russian} is a better parent to \transl{English}{Estonian} than \transl{English}{Finnish} despite being linguistically related, moreover having the same script.\footnote{We do not transliterate the Cyrillic as done by other works \parcite{nguyen2017transfer}.} Thus the main difference could be in the number of parent parallel sentences, where \pair{Russian}{English} has 12.6M sentences, and \pair{Finnish}{English} has only 2.8M sentences.

Besides, we showed in \cref{sec:no_language_in_common} that entirely unrelated languages still yields improvements in the child model. On the other side, having less resourceful parent can harm the performance of the child as we showed in \cref{sec:low_resource_parent}.

These are indications that the relatedness of languages is not the main factor in transfer learning, and the size of the parent model has a bigger influence on child performance. In this section, we investigate the phenomenon of language relatedness in contrast to the training size of the parent model.
%shed some light on a question if the language relatedness plays an important role or if other factors such as the size of parent corpus are more critical.

\subsection{Artificially Related Language Pair}
\label{sec:artificial_related_pair}

Many factors influence language relatedness: linguistic family, writing script, grammar phenomena, etc. Therefore it is hard to measure the relatedness of various languages, especially comparing training data sizes with various degree of relatedness. 
In this section, we evaluate the effect of various degree of relatedness in contrast to various training data sizes. We present artificially related language where we can influence the degree of the language relatedness and measure the performance of the child. The artificial language is prepared by harmful modifications of the original training set.

There are several options on how to prepare artificial related language pair from the original training data. We can either shuffle words in the sentence, which creates a language pair with the same vocabulary, but different word order. The second option is to shuffle words within the language, e.g. ``cat'' would always be replaced by ``juggling''. The third option is to shuffle individual characters within each word type based on an exact replacement rule. The variants are visualized in \cref{fig:various_ways_of_noise}.

\begin{figure}
\caption{Various ways of damaging original sentences. Each column corresponds to the original word. All occurencess of a word type are replaced with the same string anywhere in the corpus, except for option 1 with shuffled words.}
\begin{center}
\begin{tabularx}{\textwidth}{@{}lllllllll@{}}
\toprule
  Original: & my     & cat         & likes  & playing & with     & my & other   & cats \\
  Option 1: & likes  & playing       & other   & cats    & my       & cat & with & my \\
  Option 2: & has    & juggling    & rather & study   & research & has & those   & set \\
  Option 3: & tf     & jha         & sprlz  & wshfpun & dpao     & tf & vaoly   & jhaz \\
\bottomrule
\end{tabularx}
\end{center}
\label{fig:various_ways_of_noise}
\end{figure}

The first option is prone to the word order, if we would like to have artificial language with consistent word order we need to rely on some linguistic analysis, which would add another layer of uncertainty to this experiment, however, we investigate this option in \cref{sec:broken_word_order} as a way to study word order). The second option is problematic as we want the related language to have a property that similar words behave similarly, for example, a word ``cats'' should be replaced with something similar to ``juggling'', however this would need an in-depth analysis of clusters of words and generation of rules which words are mapped to which. It is very hard, especially for an inflected language such as Czech. The third option generates a language with the same word order and language where visually similar words appear in a similar context. However, the language is unintelligible from the original. Moreover, due to the subword segmentation, the actual lengths of sentences as seen by \ac{NMT} vary and thus the \ac{NN} cannot learn easy mapping across the sentences.

We decided to create the artificial language pair by the last option of substituting characters. We mix only the alphabet characters to match real-life conditions where most languages use the same punctuation and numbers. Furthermore, we preserved the capitalization. We modify both source and target language. Therefore there is no shared unmodified language between parent and child.

In order to scale the relatedness of languages, we randomly select X\% of words from source and target language and substitute characters only in the remaining words; thus we obtain corpus with X\% words unchanged. This way we get a pseudo related language pair with a varying degree of relatedness where 0\% is almost unrelated, and 100\% is identical language. An example is in \cref{fig:pseudo_related_language_pair}.

\begin{figure}
\caption{An example sentence in various ratio of substitution.}
\begin{center}
\begin{tabularx}{\textwidth}{@{}ll@{}}
\toprule
Original: & \bf Pardon? Have you seen this cat? \\
70\% related: & \textbf{Pardon?} Crnk \textbf{you seen this} tre? \\
50\% related: & Irfjwh? \textbf{Have you} ykkh ecsy \textbf{cat}? \\
30\% related: & Irfjwh? Crnk bwm \textbf{seen} ecsy \textbf{cat}? \\
0\% related: & Irfjwh? Crnk bwm ykkh ecsy tre? \\
\bottomrule
\end{tabularx}
\end{center}
\label{fig:pseudo_related_language_pair}
\end{figure}

In this experiment (\TFC{}), we use \transl{English*}{Czech*} as a parent model with various degree of relatedness and various amount of training data. The corpus is created from CzEng 1.7 (see \cref{sec:czeng}).
We experiment with 80\%, 50\%, and 0\% related corpus and each in 2M, 5M, 10M, and 20M parallel sentences. We use a warm-start technique where all models use the same vocabulary that is created from 50\% related corpus. 

As the child language pair, we use 100k unmodified random sentences from the same corpus that have not appeared in our parent corpus. The performance of a child when trained solely on its training data is 7.17 BLEU.

The training process is as follows: train the parent model for 1M training steps, take the last model, and continue with the training of the child model for additional 200k steps. The best child model is selected based on the development data and evaluated against \transl{English}{Czech} testset.

The results in \cref{tab:datasize_vs_relatedness} present an interesting pattern. We can see that having more data can be more useful than a related language with fewer data. For example, 50\% parent with 2M parallel sentences reached 16.76 BLEU, in contrast, having ten times more data but 0\% related parent yields 18.25 BLEU. On the opposite, whenever the difference is only double, then the relatedness helps more with fewer data (19.13 BLEU vs. 18.25 BLEU). With 5M training data, the 50\% parent already performs better than 20M with 0\% (19.63 BLEU vs. 18.25).

We obtained similar results in real-life experiments as noted in the previous chapter, where more resourceful language (Czech or Russian) performed better than related language with less data (Finnish) with the Estonian child.

\observation{Language relatedness plays a role in transfer learning. However, the amount of parent training data can improve performance even more than language relatedness.}{obs:relatedness_is_less_important_than_datasize}

\begin{table}[t]
\caption{The results of various parent models. Each column specifies the size of parent training data, which is randomly downsampled from the original. Each row specifies parent model relatedness. The scores are in BLEU and specify performance of child model trained from the parent. 
For models with the star the performance dropped quickly during child training.}
\begin{center}
\begin{tabularx}{\textwidth}{@{}lCCCC@{}}
\toprule
       & 2M & 5M & 10M & 20M \\
\midrule
80\% related & 18.11 & *20.46 & *21.81 & *22.10 \\
50\% related & 16.76 & \ps{*}19.63 & \ps{*}19.13 & \ps{*}19.61 \\
0\% related & 15.09 & \ps{*}16.83 & \ps{*}17.91 & \ps{*}18.25 \\
\bottomrule
\end{tabularx}
\end{center}
\label{tab:datasize_vs_relatedness}
\end{table}

This finding can help when deciding which parent language pair to choose for a particular child. For example, whenever related language pair does not have enough training data, we could choose any training corpus with a high number of parallel sentences (for example \pair{Czech}{English}). However, our experiment is performed on artificially related languages, which could be considered as a noisy parent model. Moreover, we evaluated only 12 settings of relatedness and training size.

We noticed that with 80\% relatedness the child model (labeled by a star) performs best without the training on its training data as the performance quickly deteriorated during the child training.

The language relatedness is not the only criterion or the most important one. Even unrelated parent can improve the performance of the child. In the next section, we examine the effect using parent language pair with an artificially huge number of parallel sentences.

\subsection{Parent Trained on Large Mix of Languages}
\label{sec:any_data_are_good}

We showed that even completely unrelated language can yield improvements in the child model and that the number of parallel sentences plays an important role. Therefore we investigate scenario, where we create the biggest possible training corpus by mixing all corpora we have at hand.

We created artificial training corpus that translates between English and a mix of languages. Our target child is \pair{Estonian}{English}. Therefore we avoided the training corpus of this language pair. We collected training corpora of most languages from \ac{WMT} 2019, most of them we already used throughout this thesis. Then we mix them together by aligning the English on one translation side and putting the other languages together on the other side. This way, we create \pair{Mix}{English} corpus. 

We have not added any target language label as is used in the multi-task learning \parcite{lu2018interlingua} so our resulting model is not suitable for any practical translation; it picks a target language at random for every input segment. 

We collected following language pairs with various training sizes: German-\EN{} with 42.2M sentences, Czech-\EN{} with 40.1M, Russian-\EN{} with 14.7M, Italian-\EN{} with 8.3M, Slovak-\EN{} with 4.3M, Finnish-\EN{} with 2.8M, Dutch-\EN{} with 2.6M and Basque-\EN{} with 0.6M parallel sentences. The final shuffled corpus has 117M parallel sentences.

We trained two models (\TFC{}): \transl{English}{Mix} and \transl{Mix}{English} for 11.6 million training steps each. For training, we have used multi-GPU training with 8 NVidia GeForce 1080 graphical cards. 

%Because the training process takes too long and we wanted to evaluate mode than one language pair, 
We decided to use our cold-start technique in order to leave the vocabulary selection for the child instead of warm-start transfer learning, which specifies a vocabulary in advance of parent training.

Before we evaluate transfer learning, we investigate the trained parent models. We took the \pair{Estonian}{English} testset and translated it with both \transl{English}{Mix} and \transl{Mix}{English} models. We have noticed that the \transl{English}{Mix} generates outputs in various languages but never mixes them within a sentence. For example, whenever the translation starts with Czech, the entire output is in proper Czech. Thus we automatically evaluated the translated sentences and counted how many languages appear in the output of translated \pair{Estonian}{English} testset:

\begin{itemize}
\item \transl{Mix}{English} -- 80.4\% Estonian, 18.2\% English and 1.4\% others.
\item \transl{English}{Mix} -- 43.1\% German, 26.5\% Czech, 22.0\% Russian, 3.5\% Italian, 1.7\% Dutch and 3.2\% others.
\end{itemize}

\transl{Mix}{English} often only copies the source into the target. Therefore it generates mostly Estonian sentences (sentences copied from the source). \transl{English}{Mix} on the other hand somewhat randomly select translation language, we can see it reflects proportions from training corpus, which contains the most sentences from German and Czech, followed by the Russian and Italian.

\observation{\ac{NMT} reflects the domain (language) distribution of sentences from the training corpus despite any additional knowledge.}{obs:nmt_reflect_distribution}

\begin{table}[t]
\caption{The result of \pair{Estonian}{English} child trained off of various parent models. The \significantmark{} represents significantly better results.}
\begin{center}
\begin{tabularx}{\textwidth}{@{}lCC@{}}
\toprule
 & Czech (Warm-start) & Mix (Cold-start) \\
\midrule
\transl{\EN{}}{Estonian} & 20.07 & \bf 20.72 \significant{} \\
\transl{Estonian}{\EN{}} & 23.35 & \bf 24.76 \significant{} \\
\bottomrule
\end{tabularx}
\end{center}
\label{tab:huge_parent_results}
\end{table}

Results of \pair{Estonian}{English} child trained off of \pair{Mix}{English} parent language pair are in \cref{tab:huge_parent_results}. Before discussing the result, we have to mention that we are comparing the cold-start direct transfer (column ``Mix'') with warm-start balanced vocabulary (column ``Czech''). The reason is that we trained the Mix model by \TFC{} and we realized too late that our cold-start experiments are trained by \TFB{} setup (see the differences in \cref{sec:used_architecture}). Thus instead of training the model again, we are comparing the Mix model with warm-start setup also trained on \TFC{}. 

Therefore we remind the results from \cref{sec:warm_cold_comparison}, where we found out that the warm-start technique always has better performance than the cold-start direct transfer.

In \cref{tab:huge_parent_results} we see that Mix model performs significantly better than the warm-start technique on Czech. Moreover, based on our findings in \cref{sec:warm_cold_comparison} whenever the warm-start technique would train the Mix model, it should obtain even better performance.

This result is unusual on its own because the parent Mix model is not a regular \ac{MT} system as it was trained on a mix of languages and cannot be used for translating in practice because it generates sentences from various training languages at random.

\observation{More data in the parent model yields better performance of a child even when the parent model is trained on a mix of languages.}{obs:more_data_are_just_better}

We showed that the parent model does not have to be a model trained on the same distribution of data, for example, the same language pair or similar modification of parent training data as in \cref{sec:artificial_related_pair}.

\subsection{Conclusion on Language Relatedness}

We illustrated that less related language pairs with a higher number of parallel sentences can improve the performance of the child more than a related language with less parallel sentences. For example in \cref{sec:broad_evaluation} in \cref{tab:warmstart_highresourceparent} we showed that \transl{English}{Czech} parent works better for \transl{English}{Estonian} than linguistically related \transl{English}{Finnish} (20.41 BLEU vs. 19.74 BLEU). Furthermore, even languages with a different script and more training data such as \transl{English}{Russian} can improve the performance of the child \transl{English}{Estonian} more than the related \transl{English}{Finnish} (20.09 BLEU vs. 19.74 BLEU). On the other hand, whenever the shared language (in our case English) is on the target side, the more related language \transl{Finnish}{English} is a better parent to \transl{Estonian}{English} than \transl{Russian}{English} (24.18 BLEU vs. 23.54 BLEU). These examples show that the relatedness of the parent is important. However, the size of parent corpus also plays a critical role in transfer learning.

In \cref{sec:no_language_in_common} we experimented with languages that have both parent languages different from the child. Moreover, they have the same number of parallel sentences (10M sentences). We observed that the improvements of the child model have been similar across all four parents, which suggest that the same amount of training data leads to the same improvements. The improvements were even for languages that do not share writing script.

In \cref{sec:artificial_related_pair} we showed on an artificial parent that language relatedness plays a role whenever the differences in training data are small (e.g. twice as much data), however having parent trained on much more parallel sentences leads to better child performance regardless the relatedness.

\cref{sec:any_data_are_good} we confirmed the observations by training a parent model on a mix of all training corpora we have available and obtaining improvements in child performance. 

Based on our observations across the thesis, we conclude that although the relatedness of languages plays a role the size of training corpus seems to overcome the relatedness whenever the parent can be trained on much more parallel sentences, even up to the point where the parent model is not useful to the translation by itself.

\section{Linguistic Features versus Better Initialization}
\label{sec:linguistic_vs_initialization}

Neural networks have the reputation of being a black box. In this section, we try to understand if the gains can be attributed to some linguistic features or if the main contribution of transfer learning is simply a better initialization of weights than random initialization. There are several possible explanations of what are the reasons for the improvements:

\begin{itemize}
  \item Knowledge of the shared language, e.g. English.
  \item General linguistic knowledge transferred from the parent model (e.g. word order patterns or typical lengths of sentences).
  \item Better hyper-parameters that are changed after the start (especially learning rate).
  \item Better initialization of weights (weights transformed from the parent could have more suitable distributions than the random initialization).
\end{itemize}

The main contribution could be due to the shared language between parent and child, e.g. English. Although it is possibly one of the main aspects, it is not the only one. In \cref{sec:no_language_in_common}, we showed that transfer learning also helps languages that have both source and target language different from the parent model, i.e. \transl{Spanish}{Russian} helping \transl{Estonian}{English}.

\perscite{zoph2016transferLowResource} showed that transfer learning does not utilize only the shared English, but also other parameters from the second language. Thus other options could lead to the improvements of the child model. 

Alternatively, the improvements could yield from linguistic features. For example, the child could transfer some knowledge about the word order or the ratio in source and target lengths.
On the other hand, the main benefits could be attributed only to \ac{NN} layout. If we compare the training model from a random initialization or a parent model, there are two main differences. The first difference is in the initial weights, where transfer learning has weights initialized already in some informed part of the parameter space compared to the random initialization. The second difference is in the learning rate because we are using non-constant learning rate depending on the global number of steps it changes through the training process, which can simply mean that different learning rate could lead to the same improvements. 

In this section, we start by discussing the effect of shared language on the \ac{NMT} by analyzing various layers of the network in \cref{sec:freezing} as well as investigating the behavior of the model when the shared language changes position from parent to child in \cref{sec:direction_swap}. Then we explore some linguistic features of parent training data and analyze the generated output in \cref{sec:broken_word_order}, \cref{sec:output_analysis} and \cref{sec:various_length_of_parent_training_corpus}. We conclude the section by analysis of learning rate influence in \cref{sec:parent_performance_influence} and comparing transfer learning to random initialization in \cref{sec:same_lang_in_reverse}.

\subsection{Freezing Parameters}
\label{sec:freezing}

\perscite{thompson2018freezing} investigated, which parts of a model are responsible for the gains during domain adaptation. They used the technique of freezing model sub-networks to gain an insight into \ac{NMT} system behavior during the continued training.

They segmented the RNN model \parcite{bahdanau2015neural} into five sub-networks: source embeddings, target embeddings, encoder, decoder with attention mechanism and the softmax layer responsible for the generation of the output. Then they follow standard scenario of domain adaptation (see \cref{sec:domain_adaptation}).

In this section, we use their technique to evaluate which parts of the neural network are crucial for transfer learning. In order to analyze transferred parameters that are the most helpful for the child model and which need to be updated the most, we follow the strategy by \perscite{thompson2018freezing}. We carry out the analysis on \pair{Estonian}{English} pair with \pair{Czech}{English} parent.

Based on the internal layout of Transformer model parameters in the \ac{T2T}, we divided the model into four parts. (i) Word embeddings map each subword unit to a dense vector representation. The same embeddings are shared between the encoder and decoder. (ii) The encoder part includes all the six feed-forward layers converting input sequence to the deeper representation. (iii) The decoder part is again six feed-forward layers preparing the choice of the next output subword unit. (iv) The multi-head attention is used throughout encoding as well as decoding, as self-attention layers interleaved with the feed-forward layers (see \cref{sec:transformer}). We do not separate the self-attention layers used in the encoder or decoder, therefore when freezing encoder (resp. decoder) we also freeze some of the attention layer matrices.

We run two sets of experiments (\TFC{}): either freeze only one out of the four parts and leave updating the rest of the model or freeze everything except for the examined part.

\begin{table}[t]
\caption{Child BLEU scores when trained with some parameters frozen. Each row represents a parameter set that was fixed at the pre-trained values of the \transl{Czech}{English} parent. Best result for frozen parts in each column in bold.}
\begin{center}
\begin{tabularx}{\textwidth}{@{}lCC@{}}
\toprule
Frozen part  & \transl{\EN{}}{Estonian} & \transl{Estonian}{\EN{}} \\
\midrule   
All          & \p{}1.99 & \p{}1.39\\       
Embeddings   & 19.79 & 22.89\\
Encoder      & 19.65 & 20.61\\  
Decoder      & 18.76 & \bf 23.95 \significant{}\\
Attention    & 19.73 & 23.00\\
None         & \bf 20.07 & 23.35 \\
\bottomrule
\end{tabularx}
\end{center}
% in bold neni preklep, je to neslovesna veta
\label{tab:freeze_one}
\end{table}

\begin{table}[t]
\caption{
Child BLEU scores when trained with most parameters frozen. Each row represents a parameter set that was free to train; all other parameters were fixed at their pre-trained values. Best result for non-frozen parts in each column in bold. The results marked with * diverged as the model could not train anything.}
\begin{center}
\begin{tabularx}{\textwidth}{@{}lCC@{}}
\toprule
Non-frozen part & \transl{\EN{}}{Estonian} & \transl{Estonian}{\EN{}} \\
\midrule   
All           & \bf 20.07 \significant{} & \bf 23.35 \significant{}\\ 
Embeddings    & * & *\\ 
Encoder       & * & 13.21\\   
Decoder       & \p{}7.87 & \p{}5.76\\ 
Attention     & \p{}6.19 & 10.69\\
None          & \p{}1.99 & \p{}1.39\\    
\bottomrule
\end{tabularx}
\end{center}
\label{tab:freeze_but_one}
\end{table}

The results are in \cref{tab:freeze_one}. Based on the results the most important part of \ac{NN} that has to be changed is the decoder for \transl{\EN{}}{Estonian} (resp. encoder for \transl{Estonian}{\EN{}}) that handles the language that changes from the parent to child. With this part fixed, the performance drops the most. 

The same observation is confirmed in \cref{tab:freeze_but_one}: all the model parts (including the multi-head attention) can be reused precisely from the parent model as long as the decoder for \transl{\EN{}}{Estonian} (resp. encoder for \transl{Estonian}{\EN{}}) can learn the new language.

We got a significantly \significantmark{} better score when the decoder was frozen compared to when all the network were free to train in \transl{Estonian}{English} (23.95 vs. 23.25 BLEU). This shows that at least for examined language pair, the Transformer model lends itself very well to decoder reuse. However, we do not see the same in the opposite direction, which confirms that the position of the shared language makes the task different as we discussed in \cref{sec:shared_decoder_is_easier}.

\observation{Freezing decoder when the target language is shared during child training can significantly improve the final performance.}{obs:freezing_can_improve_performance}

\perscite{zoph2016neural} needed freezing embeddings for their transfer learning to successfully work. On the other hand, freezing embeddings is harmful to our transfer learning.

Other results in \cref{tab:freeze_one} reveal that the architecture can compensate for some of the training deficiencies. Freezing the encoder (resp. decoder for \transl{Estonian}{\EN{}}) or attention
is not that critical as a frozen decoder (resp. encoder). 

\observation{Transformer model is robust enough to compensate for some frozen parts and reach a comparable performance.}{obs:nmt_is_robust_when_frozen}

Whenever we freeze everything except a particular layer, we get a completely new picture. Results in \cref{tab:freeze_but_one} show that the network struggles to change the behavior from the parent when most of the network is frozen. Especially the parent embeddings are the least useful for the child because keeping only them leads to diverged training. The diverging results show that \ac{NN} is not capable of providing all the needed capacity for the child, unlike the self-attention.

All in all, these experiments illustrate the robustness of the Transformer model in that it is able to train and reasonably well utilize parent weights even when the training is severely crippled. Interestingly, the attention is a crucial part of the network as it can compensate for the harmful effect to some extent. We use this knowledge in \cref{sec:broken_word_order}, where we evaluate the parent model with damaged word order.

\subsection{Direction Swap in Parent and Child}
\label{sec:direction_swap}

In the previous section, we showed that the side of the network with shared language is modified the least, and when frozen, it leads to the best performance.

We experimented with scenarios of shared-source, shared-target, and no-shared language. In this section, we investigate the scenario, where the shared language is switched to the other side between parent and child in order to investigate if the technique can transfer other features across various parts of the network.

In other words, we now allow a mismatch in the translation direction of the parent and child. 
The parent \transl{XX}{English} is thus followed by an \transl{English}{YY} child or vice versa. We use \pair{Estonian}{English} language pair as the child with various parents. The results are from our paper \persciteA{kocmi2018trivial} (\TFA{}). 

This way, the child cannot use the parent target language model as languages on both sides changed. It is important to note that Transformer shares word embeddings for the source and target side. However, we showed in the previous section that the word embeddings are not crucial for the training, although some improvements could be due to better English embeddings.

\begin{table}[t]
\caption{Results of child following a parent with swapped English side.
``Baseline'' is trained on child data only. ``Aligned \EN{}'' is the more natural setup with English appearing on the ``correct'' side of the parent, the numbers in this column thus correspond to those in \cref{tab:warmstart_highresourceparent}. The \significantmark{} represents significantly better model when comparing ``Transfer'' and ``Baseline''.
}
\begin{center}
\begin{tabularx}{\textwidth}{@{}llCC|C@{}}
\toprule
Parent & Child & Transfer & Baseline & Aligned \EN{}\\
\midrule
\transl{Finnish}{\EN{}} & \transl{\EN{}}{Estonian} & \bf 18.19\significant{} & 17.03 & 19.74\\
\transl{Russian}{\EN{}} & \transl{\EN{}}{Estonian} & \bf 18.16\significant{} & 17.03 & 20.09\\
\transl{\EN{}}{Finnish} & \transl{Estonian}{\EN{}} & \bf 22.75\significant{} & 21.74 & 24.18\\
\transl{\EN{}}{Russian} & \transl{Estonian}{\EN{}} & \bf 23.12\significant{} & 21.74 & 23.54 \\
% \transl{\EN{}}{Czech} & \transl{Estonian}{\EN{}} & \bf 22.80\significant{} & 21.74 & not run \\
\bottomrule
\end{tabularx}
\end{center}
\label{tab:swapped_english}
\end{table}

The results in \cref{tab:swapped_english} document that an improvement can be reached even when none of the involved languages is reused on the same side.
This suggests that the model can transfer further knowledge across various parts or layers of the network. Although there are too many factors that could influence it (language relatedness, parent training size, etc.), thus we cannot make any definite conclusion.

More importantly, the improvements are better when the shared language is aligned (column ``Aligned EN''), which concludes that the shared language does play a significant role in transfer learning. This finding could be used whenever we want to train only one parent, for example, due to the time or resource restrictions and use it for more children even those with the shared language on the other side.

However, it cannot be the only source of improvements as we showed in \cref{sec:no_language_in_common} that the no-shared language scenario improves the performance of the child.

\observation{The improvements of transfer learning are partly but not fully attributed to the shared language between parent and child.}{obs:shared_english_do_play_a_role}

\subsection{Broken Word Order in Parent Model}
\label{sec:broken_word_order}

We showed that the shared language plays a vital role in transfer learning, now we attempt to find some particular linguistic features explaining the gains.

Intuitively, the child model could transfer linguistic knowledge from the parent such as word order, length of the target language sentences, etc. In this section, we experiment with the somehow modified parent to study the effect on the transfer of knowledge.

We use the \transl{English}{Czech} as a parent model and the child is \transl{English}{Estonian}. We used the shared-source language scenario as it is harder for transfer learning (see \cref{sec:shared_decoder_is_easier}), and the improvements cannot be attributed to the shared English target side as a better language model.

In the first experiment, we change the word order of a parent language pair in order to find out if the language word order plays an important role in transfer learning. Both of examined languages have mostly SVO (subject-verb-object) word order.

The word order is important for the attention mechanism that learns which parts of the source it should consider most.

We use sorting or shuffling of words (tokenized on whitespace) as a way to create a broken parent language. We modify only the word order of the parent model, therefore ``shuffle source'' means shuffling the source sentences (English in this experiment) and leaving the target language unmodified.

Additionally, we created an experiment, where we shuffled all sentences which breaks the sentence pairs (row ``Shuffled sentences''). This way, the parent model could learn to generate random sentences that are not related to the source sentence. Thus it can mainly learn the decoder's language model.

\begin{table}[t]
\caption{Results of transfer learning with modified parent word order. The performance is measured in BLEU.}
\begin{center}
\begin{tabularx}{\textwidth}{@{}lCC@{}}
\toprule
Parent & Child Performance & Parent Performance \\
\midrule   
Unmodified parent & 20.07  & 23.48  \\
Shuffle source & 19.18  & 12.63 \\
Shuffle target & 19.16  & \p{}2.78 \\
Shuffle both & 18.43  & \p{}2.23 \\
Sort target & 19.45  & \p{}2.29 \\
Shuffled sentences & \p{}0.03  & \p{}0.00 \\
% \transl{English}{English} & 14.48  & 1.85  \\
Baseline & 15.80  & -- \\
\bottomrule
\end{tabularx}
\end{center}
\label{tab:broken_parent}
\end{table}

Results (\TFC{}) are tabulated in \cref{tab:broken_parent}. The sorting of target side is less damaging than shuffling. Parent with both source and target side shuffled has the worst performance. However, it is still a good parent model for transfer learning and improves the performance of the child be more than 2 BLEU points over the baseline. 

Interestingly, the neural network has equal performance whenever the source and the target are shuffled despite the different performance of the parents (12.63 vs. 2.78 BLEU). We checked the performance of child model by various automatic metrics, and in all of them, the shuffled source and target reached similar scores. 

This suggests that the word order of the target language is not the main feature in transfer learning. A better explanation is that the shuffling of either source or target breaks the attention mechanism in the same way, and then it needs to retrain on the child data.
However, further analysis is needed as there can be other factors that can cancel each other out.

% Shuffling of words in source or target parent sentences plays a little role for final child performance.
\observation{There is little difference for the final child performance between shuffled word order of parent's source or target language.}{obs:shuffling_has_same_effect}

The poor performance of parent on \transl{English}{Czech} testset is understandable because the BLEU is computed based on $n$-grams of words that are heavily disrupted by shuffling. When we studied the outputs of the parent models, we noticed that they actually learned to translate and only the shuffle the output on top of that. The ``sorted target'' model translates and even alphabetically sorts the outputs. This documents the high flexibility of Transformer's skills in capturing word relations: between source and target, it happily resorts to lexical relations, and within the target, it easily captures (memorizes) alphabetical sorting. 

We have noticed interesting behavior of ``sort target''. The model generates correct words in alphabetical order. We compute the unigram BLEU score and get 51.1 (compared to an unmodified model that has 53.8 unigram BLEU).
If we evaluate the BLEU on the sorted target, we obtain 14.77 BLEU (result is not in the table). This result shows that the model learns to translate without access to the word order.

\observation{\ac{NMT} models, in general, can learn to some extent in the scenario with source sentences having broken word order. Although the word order is crucial for a good performance.}{obs:wordorder_is_important_for_nmt}

\begin{table}[t]
\caption{Various automatic scores on \transl{English}{Estonian} testset. Scores prefixed ``n''
reported as $(1-\text{score})$ to make higher numbers better. }
\begin{center}
\begin{tabularx}{\textwidth}{@{}lC@{~~}C@{~~}C@{~~}C@{~~}C@{~~}c@{}}
\toprule
Parent               & BLEU       & nPER       & nTER       & nCDER      & chrF3     & nCharacTER \\
\midrule
Baseline      & 17.03      & 47.13      & 32.45      & 36.41      & 48.38     &      33.23 \\
\transl{English}{Russian}      & 20.09      & 50.87      & 36.10      & 39.77      & 52.12     &      39.39 \\
\transl{English}{Czech}     & 20.41      & 51.51      & 36.84      & 40.42      & 52.71     &      40.81 \\
\bottomrule
\end{tabularx}
\end{center}
\label{tab:output_analysis_enet}
\end{table}

We showed that damaging the word order in the parent model leads to a slight drop in performance of the child, still staying high above the baseline. In the next section, we analyze the outputs of the child model and look for potential over-estimations of translation quality that could emerge from the usage of BLEU metrics.

Lastly, ``Shuffled sentences'' cannot learn anything obtaining 0.03 BLEU. The parent model learned to generate one sentence for each input, and the child only changed the sentence into Estonian and generated ``See on meie jaoks väga tähtis.'' (\ac{MT} gloss: ``This is very important to us.'').

\subsection{Output Analysis}
\label{sec:output_analysis}

We rely on automatic evaluation. Thus we need to prevent some potential over-estimations of translation quality due to BLEU.
For this, we took a closer look at the baseline \transl{English}{Estonian} model (BLEU of 17.03 in \cref{tab:warmstart_highresourceparent} (\TFA{})) and two \transl{English}{Estonian} children derived from \transl{English}{Czech} (BLEU of 20.41) and \transl{English}{Russian} parent (BLEU 20.09).

\cref{tab:output_analysis_enet} confirms that the improvements are not an artifact of uncased BLEU. The gains are apparent with several (now cased) automatic scores.

\begin{table}[t]
\caption{Candidate total length, BLEU $n$-gram precisions and brevity penalty (BP). The reference length in the matching tokenization was 36062.}
\begin{center}
\begin{tabularx}{\textwidth}{@{}lCcC@{}}
\toprule
Parent               & Length    & BLEU Components         & BP     \\
\midrule               
Baseline      & 35326    & 48.1/21.3/11.3/6.4      & 0.979   \\
\transl{English}{Russian}      & 35979    & 51.0/24.2/13.5/8.0      & 0.998   \\
\transl{English}{Czech}     & 35921    & 51.7/24.6/13.7/8.1      & 0.996   \\
\bottomrule
\end{tabularx}
\end{center}
\label{tab:output_analysis_bleu_comps}
\end{table}

As documented in \cref{tab:output_analysis_bleu_comps}, the outputs of transferred models are slightly longer in terms of words produced. In the table, we also show individual $n$-gram precisions and \ac{BP} of BLEU. The longer output helps to reduce the incurred \ac{BP}, but the improvements are also apparent in $n$-gram precisions. In other words, the observed gain cannot be attributed solely to producing longer outputs.

\observation{Transfer learning helps the child model to generate slightly longer sentences, and there are also clear improvements in produced n-grams.}{obs:child_generates_longer_sentences}

\begin{table}[t]
\caption{Comparison of child outputs vs. the baseline and reference. Each column shows child trained from different parent either \transl{English}{Russian} or \transl{English}{Czech}.}
\begin{center}
\begin{tabularx}{\textwidth}{@{}lY@{ }XY@{ }X@{}}
\toprule
Appeared in & \multicolumn{2}{c}{\transl{English}{Russian}}    & \multicolumn{2}{c}{\transl{English}{Czech}}      \\
\midrule
Baseline+Reference          & 15902 & (44.2 \%)       & 15924 & (44.3 \%) \\
Baseline only           & 7209 & (20.0 \%)        & 7034 & (19.6 \%) \\
Reference only           & 3233 & (9.0 \%)         & 3478 & (9.7 \%) \\
Neither          & 9635 & (26.8 \%)        & 9485 & (26.4 \%) \\
Total      & 35979 & (100.0 \%)      & 35921 & (100.0 \%) \\
\bottomrule
\end{tabularx}
\end{center}
\label{tab:output_analysis_token_anots}
\end{table}

\cref{tab:output_analysis_token_anots} explains the gains in unigram precisions by checking words in the child outputs that were present also in the baseline and/or confirmed by the reference. 
We see that about 44+20\% of words of child outputs can be seen as ``unchanged" compared to the baseline because they appear already in the baseline output. (The reference confirms the 44\% words.)

The differing words are more interesting: ``Neither'' denotes the cases when the child model produced something different from the baseline and also from the reference. Gains in BLEU are due to ``Reference only'' words, i.e. words only in the child output and the reference but not in the baseline. For both parent setups, there are about 9--9.7 \% of such words. We looked at these 3.2k and 3.5k words, and we have to conclude that these are regular Estonian words; no Czech or Russian leaks to the output and the gains are not due to simple word types common to all the languages (punctuation, numbers or named entities). We see nearly identical BLEU gains even if we remove all such simple words from the child output and references. 

% \XX{tohle by se jeste dalo vylepsit srovnanim s trenovacimi daty; ty tokeny, ktere to driv nedelalo a nove zacalo, jsou iv trenovacich datech? A pak nejake srovnani chodu baseline vs. chodu transferovaneho na vetach, kde k tomuto zlepseni doslo ... proc je pro transferovy model levnejsi udelat hezci nez to bylo pro baseline}

\subsection{Various Lengths of Parent Sentences}
\label{sec:various_length_of_parent_training_corpus}

In \cref{sec:output_analysis}, we showed that the child model generates slightly longer sentences than when trained without transfer learning. In this section, we investigate the effect is visible also when the parents are trained on corpora with sentences limited to certain length ranges.

In this experiment, we take the \pair{Czech}{English} corpus and randomly select sentences of various length creating training corpus for different parents. Each of the parents is trained with the corpus of sentences with lengths in the predefined range. The length of the source and target sentences are different. Therefore we use the sum of both lengths as the criterion.

We use four parent models. The first has sentence pairs with the length of either source or target of at most 10 words (tokenized by spaces). The second parent uses sentences of length 10 to 20 words. The third parent uses sentences of length 20 to 40, and the last parent uses sentences of length 40 to 60 words.
We use \transl{English}{Estonian} as the child model in \TFC{} setup for this experiment.

We want to make the experiment comparable. Therefore each training corpus has exactly 300M words. Thus training set with the shortest sentences has the most sentences altogether.

\begin{table}
\caption{Performance and average number of words generated over the testset. The references have average number of words 15.6 for the Czech testset 15.4 for the Estonian testset.}
\begin{center}
\begin{tabularx}{\textwidth}{@{}lCCCC@{}}
\toprule
 & \multicolumn{2}{c}{Parent}    & \multicolumn{2}{c}{Child}\\
\cmidrule(r){2-3} \cmidrule(l){4-5}
Sentence lengths & BLEU & Avg. words & BLEU & Avg. words  \\
\midrule
1-10 words  &  \p{}8.57 & 10.9 & 16.57 & 15.3\\
10-20 words & 16.21 & 15.4 & 17.48 & 15.3\\
20-40 words & 12.59 & 21.9 & 17.99 & 15.3\\
40-60 words &  \p{}5.76 & 35.5 & 16.80 & 15.5\\
1-60 words & 22.30 & 15.3 & 19.15 & 15.4 \\
\bottomrule
\end{tabularx}
\end{center}
\label{tab:parent_with_various_length}
\end{table}

\cref{tab:parent_with_various_length} shows the results from our experiment. We start by discussing the parent performance. The parent model is unable to generalize from training data of different length for the training set. Although it generates sentences with slightly better length ratios than the training set, however, it cannot generalize the sentence length well. For example, when training on ``1-10 words'' it produces testset sentences with an average length of 10.9, which is more than it saw during the training. Similarly, ``40-60 words'' generates sentences of an average length of 35.5, which is shorter than training sentences.

It is an important finding that shows the need for checking training data in advance if they have similar length distribution as the target domain.

\observation{\ac{NMT} systems are highly influenced by the lengths of training sentences, and they are unable to generalize the sentence lengths.}{obs:nmt_is_highly_influences_by_training_corpora}

When we discuss the performance of the child, we see that the child model generates sentences of valid length as the actual distribution over the testset is 15.4 words per test sentence. The performance is reaching 16.57 to 17.99 BLEU where the best parent model looks like the parent trained on ``20-40 words'' corpus. However, training the model on sentences of all lengths (row 1-60 words) leads to the best performance.

\observation{Child model is not significantly influenced by the lengths of parent training sentences and can learn the length distribution by itself from child training data.}{obs:child_not_influenced_by_parent}

Therefore the finding that child model generates slightly longer sentences than baseline from \cref{sec:output_analysis} is not due to the lengths of sentences in the parent training set. 

In conclusion, it seems that the length of a parent model is not that important factor behind the performance improvements of the child model. 

\subsection{Parent's Performance Influence}
\label{sec:parent_performance_influence}

So far, we discussed features that could be considered linguistic. In the rest of the section, we examine various attributes that are associated with the neural architecture.

Transfer learning is, in fact, a way how to initialize weights for the child model in contrast to random initialization in the baseline. One of the explanations of the success could be attributed solely to the initialization or learning rate as these are only two attributes that change with the parent model. 

In this section, we study the effect of various learning rates, and final parent performances on various models trained for a different amount of time in order to find out if there is a correlation between them and the child performance. Furthermore, we answer a question if it is necessary to train the parent model fully or is only a fragment of the training time enough? 

In transfer learning, we train the parent model to its maximal performance, e.g. until convergence on the development set. In the following experiment, we compare the parent model in various stages of training and investigate the influence on the performance of the child model. 

In \persciteA{kocmi2018trivial} (\TFA{}), we took \transl{English}{Finnish} as the parent model and used the warm-start transfer learning of child's \transl{English}{Estonian} model after 50k, 100k, 200k, 400k, and 800k of parent's steps. In order to simplify notation in this section, we label a child model that started after X parent's steps as a ``X-child'', e.g. 400k-child is a model trained after 400k parent's steps.
% We take a model for \transl{English}{Finnish} and use it in various stages of training for transfer learning to the child \transl{English}{Estonian} \resultfrompaper{kocmi2018trivial}. We start transfer learning after 50k, 100k, 200k, 400k, and 800k parent's steps and examine the effect on the parent. 

\begin{figure}
\caption{Learning curves on development set for \transl{English}{Finnish} parent and \transl{English}{Estonian} child. The child starts training after various number of parent's training steps. Black dots specify a performance of the parent at the moment when the child training started. The colored dots show the best performance of the child model. The grey curve shows, how the learning rate depends on the global step number.}
\begin{center}
\input{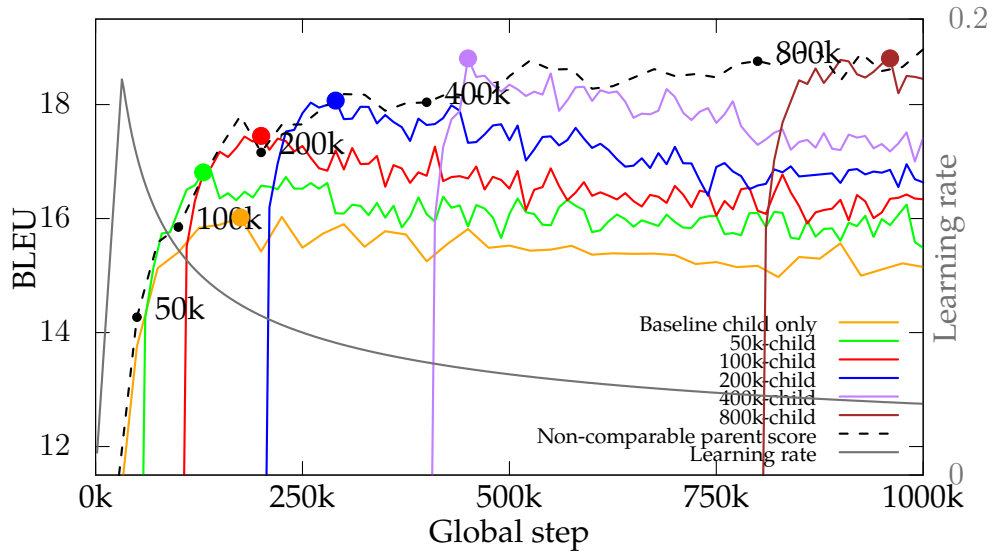}
\end{center}
\label{fig:parent_performance_progress}
\end{figure}

We conclude based on the learning curves in \cref{fig:parent_performance_progress} that the child's performance improvements correlate with the parent's performance. Furthermore, we see that only 50k steps are enough to outperform the baseline.

\observation{Longer-trained parent model (or model trained until convergence) leads to a better performing child model.}{obs:better_parent_is_better_for_child}

However, notice that the performance of the 400k-child is equal to the 800k-child, both having 18.8 BLEU despite the parent's performance improving from 18.0 to 18.8 BLEU.\footnote{We remind that the BLEU scores of parent and child are not comparable as they are calculated on different testsets, and also that the scores fluctuate a lot between iterations.} This could be only an anomaly, or it could suggest that the child's performance also depends on different factors, which we try to examine further.

One such factor is the learning rate, which is the only parameter that depends on the global number step and decreases during the training. As the learning rate, we use a function of the inverse square root of steps with the warm-up stage of 16k steps, known as the ``Noam'' scheme. The steepness of the learning rate throughout the training is visualized in grey color in \cref{fig:parent_performance_progress}.

In order examine the factor of the learning rate, we prepare experiment where we fine-tune the child model on various stages of parent training as in \cref{fig:parent_performance_progress} however we investigate various learning rates for each parent step. 

Initially, we wanted to fix the learning rate to a constant value. However, it would add a new factor to the question since even during the parent model training, the learning rate slowly decreases. Instead, we change the global step value before starting the child training to pretend that the parent was trained to a different stage than it actually was. The child learning rate follows the Noam learning curve. Thus we refer to the learning rate value using the global step index, e.g. 400k-child with 800k learning rate represent a model that fine-tunes on the parent model trained for 400k steps, and its learning rate behaves as if the parent was trained for 800k steps.

We now provide more details in the new experiment with slightly different settings. We decided to use the \transl{English}{Czech} as the parent model and \transl{English}{Estonian} as the child. We artificially downsampled the child training corpus \pair{Estonian}{English} to 100k sentences. We used Czech as a parent language pair in contrast to Finnish in \cref{fig:parent_performance_progress}. We have used only 100k compared to 800k of child's training data in the original experiment. It was motivated to examine the effect of the less resourceful language. The initial learning rate is 0.2. The following experiment uses \TFC{}.

\begin{table}[t]
\caption{Experiment comparing various stages of parent model and learning rate. Each row corresponds to the same parent model as saved at given step (parent). Columns correspond to different learning rate shifts named by the global step at which it starts. The ``0k-child'' row is therefore baseline trained only on the child training set. The column ``Parent'' represent a performance of the parent at the step when the child was spawned.}
\begin{center}
\small
\begin{tabularx}{\textwidth}{@{}lRRRRRRRr@{}}
\toprule
\multicolumn{2}{r}{Learning rate at: \hfill 0k}  & \multicolumn{1}{r}{50k} & \multicolumn{1}{r}{100k} & \multicolumn{1}{r}{200k} & \multicolumn{1}{r}{400k} & \multicolumn{1}{r}{800k} & \multicolumn{1}{r}{1600k}  & Parent\\
\midrule
0k-child    &   8.16 &  7.55 &  7.24 &  6.43 &  6.09 & 5.45 &  4.61 &  0.0\\
25k-child   & 12.02 & 12.47 & 12.76 & 12.95 & 12.97 & 12.91 & 12.98 & 17.7\\
50k-child   & 12.72 & 13.48 & 13.71 & 13.77 & 13.72 & 13.58 & 13.70 & 19.9\\
100k-child  & 13.31 & 14.01 & 14.06 & 14.26 & 14.64 & 14.59 & 14.56 & 21.6\\
200k-child  & 13.70 & 14.97 & 14.92 & 15.22 & 15.15 & 15.15 & 15.67 & 23.0\\
400k-child  & 14.16 & 15.20 & 15.89 & 15.89 & 16.08 & 15.83 & 15.84 & 23.9\\
800k-child  & 13.59 & 15.64 & 15.82 & 16.19 & 16.23 & 16.39 & 16.58 & 24.8\\
1600k-child & 14.21 & 15.68 & 15.86 & 16.29 & 16.86 & 16.61 & 16.72 & 25.2\\
\bottomrule
\end{tabularx}
\end{center}
\label{tab:parent_performance_progress_lr}
\end{table}

\cref{tab:parent_performance_progress_lr} presents the results of the experiment; all results are evaluated on the same testset. Thus they are comparable, and we present them in the form of a heat-map for better visualization. However, the values within columns should be compared with care as they differ in the parent model that has been trained for a various number of steps. Note that the learning rate scheme of the parent model never changed.

From \cref{tab:parent_performance_progress_lr}, we conclude that the learning rate schedule is important for the training. The 0k-child baseline trained without transfer learning (see the first row in \cref{tab:parent_performance_progress_lr}) has the best performance with learning rate starting at 0k. Therefore the initial warm-up steps (16k in total) are needed for proper training. In contrast, resetting learning rate to 0k across all the transfer setups (see the column ``0k'') harms the final child performance.

\observation{Warm-up steps and the peak in learning rate are crucial for good performance of the baseline. However, repeating this peak in child training damages the performance of transfer learning.}{obs:warmup_is_crucial_for_baseline}

Beyond that, there is not a clear pattern of the best learning rate. For all transfer learning results it fluctuates between 200k and 1600k learning rate where the differences in BLEU are mostly not significant, we suppose that it is due to the shape of learning rate that is close to being constant. Therefore we conclude that the learning rate is not the primary source of improvements behind transfer learning.

\observation{The improvements by transfer learning cannot be attributed to better chosen learning rate stage in its warmup-delay scheme.}{obs:it_is_not_better_due_learning_rate}

We see that with a better performing parent, the best performance of the child grows, contrarily to \cref{fig:parent_performance_progress} where the child performance has not changed between 400k-child and 800k-child. Therefore the parent performance plays an important role in transfer learning. The best child performance of 16.86 is obtained with the parent trained for 1600k steps, which is around three weeks of training on one GPU. In comparison, the average training of the child in this experiment was 50k steps. We use stopping criterion from \cref{sec:stopping_criterion} and saw all child models started to overfitting.

The improvements for the child model diminish in relation to the parent's step, for example, it takes only 400k steps to reach the performance of 16.08, but additional 1200k steps to improve by 0.72 BLEU. Interestingly, the 25k-child already outperforms the baseline. The 25k steps were trained for eight hours, thus proving that the improvements in transfer learning are usable after a short period of training the parent model. 

To answer our initial questions, we conclude that child performance depends on the performance of the parent model. We showed that the learning rate is not the main factor of the transfer learning gains, but it can limit the maximum gains if changed.
Moreover, we showed that transfer learning significantly improves child performance over the baseline, even when the parent has not been adequately trained.

\subsection{Same Language Pair in Reverse Direction}
\label{sec:same_lang_in_reverse}

We showed that transfer learning can extract knowledge from a parent model whenever the shared language is on the different translation side as well as improve the performance whenever both languages are different. In both cases, it is the additional data from the parent that probably affect the performance of the model. However, there is yet another explanation of the improvements: transfer learning improvements could be attributed to the better initialization of weights. 

When we train the neural network, we have to decide on the initialization of the whole network. As \perscite{glorot2010understanding, mishkin2015all} and we in \cref{sec:embedding_initialization} showed, \ac{NN}s are sensitive to the variance of random initialization, and bad initialization can have a huge effect on the final model performance.

We can perceive transfer learning as a way of finding a better initialization of weights for the training of the child. Thus it could improve the performance mainly due to the effect of having weights initialized to better values.

In order to investigate this explanation, we prepare an experiment where the parent model does not have any additional training data. We train the parent on reversed training data than the child, in other words, the parent is \transl{XX}{YY} model, and the child is \transl{YY}{XX}. Thus the child model does not have access to any new training data. The experiments for \pair{Estonian}{English} are from our paper \persciteA{kocmi2018trivial} based on \TFA{}, the rest are new in this thesis based on \TFC{}. We selected languages to cover low-resource languages (\pair{Estonian}{English}) as well as high-resource languages (Spanish-French).

\begin{table}[t]
\caption{Results of child following a parent with swapped direction. The \significantmark{} represents significantly better results.}
\begin{center}
\begin{tabularx}{\textwidth}{@{}llCCC@{}}
\toprule
Parent & Child &  Transfer & Child-only  & Difference\\
\midrule
\transl{\EN{}}{Estonian} & \transl{Estonian}{\EN{}} & \bf 22.04\significant{} & 21.74 & +0.30\\
\transl{Estonian}{\EN{}} & \transl{\EN{}}{Estonian} & \bf 17.46 & 17.03 & +0.43\\
\transl{\EN{}}{Finnish} & \transl{Finnish}{\EN{}} & \bf 20.23\significant{} & 19.90 & +0.33\\
\transl{Finnish}{\EN{}} & \transl{\EN{}}{Finnish} & \bf 14.51 & 14.25 & +0.16\\
\transl{\EN{}}{Odia} & \transl{Odia}{\EN{}}       & \bf \p{}7.95\significant{} & \p{}6.97 & +0.98\\
\transl{Odia}{\EN{}} & \transl{\EN{}}{Odia}       & \bf \p{}3.22 & \p{}3.19 & +0.03\\
%\transl{\EN{}}{Kazakh} & \transl{Kazakh}{\EN{}}   &\bf  3.24 & 3.11\\
%\transl{Kazakh}{\EN{}} & \transl{\EN{}}{Kazakh}   & \bf 2.10\significant{} & 1.65\\
\transl{Spanish}{French} & \transl{French}{Spanish}   & \bf 28.54\significant{} & 27.89 & +0.65\\
\transl{French}{Spanish} & \transl{Spanish}{French}   & \bf 27.69\significant{} & 27.21 & +0.48\\
\bottomrule
\end{tabularx}
\end{center}
\label{tab:shared_english_samelanguahe}
\end{table}

\cref{tab:shared_english_samelanguahe} shows a particularly exciting result: the parent does not use any other parallel sentences, but the very same corpus as a child with source and target side swapped and obtained a performance improvement. 
We see gains in both directions, although not always statistically significant. 

\observation{Transfer learning improves performance even in the scenario, where no new data are available, and the child is trained on parent training corpus in reverse direction.}{obs:swapped_direction_improves_without_new_data}

One explanation of the improvements could be that the training corpus is noisy and often contains English sentences on the wrong side, for example, in the Estonian part of the corpus. In order to verify it, we ran an automatic language identification and found out that Estonian part of the corpus contains only 0.1\% English sentences, Finnish contains 3.4\%, and Odia contains 0.0\% of English sentences. The Estonian and Odia cannot be attributed to the noisiness as there is nearly zero English sentences and the 3.4\% for Finnish is low that we do not think it is the main reason behind the improvements.

The low-resource language as \pair{Odia}{English} reached low performance due to insufficient data for model training (see \cref{sed:odia_subword_irregularity}). In contrast, the high-resource language pair of French-Spanish reached significant improvements in both directions.

It is an exciting result. The model did not have access to any new data, yet it could extract new features from the reverse language pair, which it would not learn only from the original direction. 
Similar behavior has been shown in \perscite{niu2018bidirectional}, where they mixed both directions and added an artificial token indicating the target language.

The results from this experiment support our alternative explanation that the main improvements are simply from a better initialization of the model. Although, we showed that other features further improve the final performance of the model, such as shared language between parent and child, language relatedness, or size of parent training data.

\section{Summarizing Transfer Learning Analysis}
\label{sec:discussion_on_analysis}

In this chapter, we analyzed various aspects of transfer learning. We tried to shed some light on the behavior of transfer learning as well as to peek inside the \ac{NMT} black box. Although we made numerous observations and proposed several conclusions, we only scratched the surface, and much more work is needed in order to understand the behavior of the neural network.

At the beginning of the chapter in \cref{sec:negative_transfer}, we talked about the negative transfer and its effects on transfer learning in \ac{NMT}. We found out that in \ac{NMT} transfer learning the negative effects are limited in comparison to other fields where the negative transfer is a problem. We found out that using a parent model with less training data than the child can hurt the performance more than training from random initialization. On the other hand, transfer learning is especially useful for extremely low-resource child languages, and it does not produce relics from the parent target language.

In \cref{sec:shared_decoder_is_easier}, we discussed the influence of the position of a shared language between parent and child. We found out that the shared-target language scenario converges faster and has a higher slope of the learning curve than the shared-source scenario. Both observations suggest that the network can learn more in the shared-target scenario. Furthermore, in the no-shared language scenario, the network does not forget the performance as quickly and can perform the parent translation to some extent, even with the final child model.

What is the relation between language relatedness and the parent training size was discussed in \cref{sec:more_data_is_more_important}. We observed that relatedness does improve the performance of the child. However, the size of parent training data has a more substantial effect on the final performance of the child and even parent trained on a gibberish language can improve the performance of child more than a related language with less parallel sentences. Furthermore, we confirmed the finding by training the parent model on a vast corpus from mixed languages and obtained better results than high-resource parent \pair{Czech}{English}.

In the last section, we discussed if the gains can be mainly attributed to some linguistic features or if it is merely some better initialization of weights. We observed that the child is generating longer sentences, and we found out that word order and parent sentences length play a small role in transfer learning. We confirmed that the gains are not due to a better learning rate staging. However, in the last experiment, we showed that the gains are seen even in the scenario, where the parent does not have any additional training data and is trained in reverse translation direction. We take it as the main hint that the gains are simply by a better initialization of the network.

In the next section, we provide a case study of using transfer learning in backtranslation and show that with our technique, the backtranslation can be used even on low-resource language pairs.

\section{Case Study: Transfer Learning with Backtranslation}
\label{sec:backtranslation}

Let us conclude the analysis with a case study, where we apply transfer learning to the standard backtranslation approach for a low-resource language. The backtranslation approach relies on the performance of the initial model, and whenever the initial model has low performance, the backtranslation is not suitable, which is the case of low-resource languages. In this section, we show that transfer learning can help to overcome this problem. 

This section is based on our paper \persciteA{kocmi2019wmt} (\TFC{}).

\subsection{Backtranslation}

The number of parallel sentences needed for training \ac{MT} models is often very scarce. This is especially true for low-resource languages. However, the monolingual data are often more abundant. Many strategies have been used in \ac{MT} in the past for employing resources from additional languages, see e.g. \perscite{wu2007pivot}, \perscite{nakov2012improving}, \perscite{elkholy2013pivot}, or \perscite{hoang2016pivoting}.

The approaches of using monolingual data to improve \ac{MT} date back to \ac{SMT} where they were used to improve the language model \parcite{brants2007large} or the translation model \parcite{schwenk2008investigations, bertoldi2009domain, bojar2011improving}. A similar technique of improving only the target language model was examined in early \ac{NMT} \parcite{gulcehre2015using}. 

\perscite{sennrich2016backtranslation} took a different approach. They showed that creating synthetic parallel sentences by translation of monolingual text and using this synthetic corpus as a training set leads to significant improvements in performance. This approach quickly gained popularity, and it is considered the current best practice of using monolingual data in \ac{NMT}.

The technique of backtranslation, as the method is named, consists of first training a system in the reverse direction on human-generated ``authentic'' parallel sentences, i.e. target language to source language. This system is then used to translate monolingual data of a target language. The resulting translations and their monolingual counterparts are used as additional training data for the training of the final model in the original translation direction.
Thus we can consider the backtranslation as a way of transferring knowledge learned in one direction to the reverse direction through generated data.

% prvne je smicha dohromady, to dela na nemcine ... pro TED talks pouziva adaptivni metodu
\perscite{sennrich2016backtranslation} propose two regimes of incorporating synthetic training data. Either as a fine-tuning a general model or training the model on a mixed corpus. The former method uses a general model trained on authentic data that is fine-tuned by 1:1 mix of authentic and synthetic data. The latter method by training the model from beginning on the mix of the authentic and synthetic data. The second approach became the standard approach in the \ac{NMT} systems, as can be seen annually in the WMT Translation Shared Tasks by competing systems \parcite{findings2019wmt}.

\perscite{popel2018machine} investigated yet another variation of incorporating synthetic and authentic data. He proposed to switch iteratively between synthetic and authentic data without mixing them. This approach, called concat-regime backtranslation outperformed the training on the mixture.

Furthermore, backtranslation can be used as a domain adaptation when a limited number of in-domain parallel sentences exists \parcite{chinea2017adapting, chu2018comprehensive}, and the in-domain monolingual data are available.

Lastly, \perscite{hoang2018iterative} showed a way of improving the backtranslation by repeating the backtranslation process. They presented a method of training two \ac{NMT} systems in parallel, each in the opposite translation direction of a language pair, which alternately generate synthetic data for the reverse system to improve. \perscite{hoang2018iterative} concluded that the second round of backtranslation improves performance. However, the third round seems not to lead to any significant improvements.

\subsection{Backtranslation with Transfer Learning}

Backtranslation is helpful mainly in scenarios where the parent model, which is used to translate monolingual data, has a reasonable performance \parcite{hoang2018iterative, bawden2019wmt}. However, for low-resource language as it is hard to train any suitable initial model.

We propose to use transfer learning on the low-resource language pair for training the initial model with a reasonable score. We train two models in parallel, one for each translation direction. The models iteratively generate backtranslated data for the other one. We show this approach on two low-resource languages \pair{Gujarati}{English} and \pair{Kazakh}{English}.

As a parent model we use \pair{Czech}{English} for \pair{Gujarati}{English} and \pair{Russian}{English} for \pair{Kazakh}{English}. The training procedure is as follows. 

First, we train two high-resource parent models for each studied language until convergence: \transl{English}{Czech}, \transl{Czech}{English}, \transl{English}{Russian} and \transl{Russian}{English}.

Then we apply transfer learning with the use of an authentic dataset of the corresponding low-resource language pair. 
We preserve the English side: \transl{Czech}{English} serves as the parent to \transl{Gujarati}{English} and \transl{English}{Czech} to \transl{English}{Gujarati}. The same strategy is used for transfer learning from Russian to Kazakh.

After transfer learning, we select one of the translation directions to translate monolingual data (model \circled{1}). 
As the starting system for the backtranslation process, we have selected the \transl{English}{Gujarati} and \transl{Kazakh}{English}. 
The decision for \pair{Kazakh}{English} is motivated by choosing the better performing model, see \cref{tab:main_results}. This is however only a rough estimate because higher BLEU scores across various language pairs do not always need to indicate better performance; the properties of the target language such as its morphological richness affect the absolute value of the score. For the \pair{Gujarati}{English}, we decided to start with the model with English source side in contrast to \transl{Kazakh}{English}.

After the backtranslation, we mix the synthetic data with the authentic parallel corpus and train the first backtranslated model \circled{2}. 
We repeat this process: Use the improved system (\circled{3} and then \circled{4}) to backtranslate the monolingual data, and use this data in order to train the improved system in the reverse direction.
We make two rounds of backtranslation for both directions on \pair{Gujarati}{English} and only one round of backtranslation on \pair{Kazakh}{English}.

\begin{table}[t]
\caption{Testset BLEU scores of our setup. Except for the baseline, each column shows improvements obtained after fine-tuning a single model on different datasets beginning with the score on a trained parent model. The circled names points to the systems in the right side of the table.}
\begin{center}
\begin{tabularx}{\textwidth}{@{}l|rr|rr@{}}
\toprule
Training dataset & \transl{EN}{GU} & \transl{GU}{EN} & \transl{EN}{KK} & \transl{KK}{EN}\\
\midrule
Authentic (baseline)       & 2.0  & 1.8  & 0.5   &  4.2 \\
\midrule
Parent dataset      &  0.7 & 0.1 & 0.7 & 0.6\\
Authentic (transfer learning)        &  \circled{1} 9.1 & 9.2 & 6.2 & \circled{1} 14.4 \\
Synth generated by model \circled{1} &  -- & \circled{2} 14.2 & \circled{2} 8.3 & -- \\
Synth generated by model \circled{2} & \circled{3} 13.4 & --  & -- & 17.3\\
Synth generated by model \circled{3} &    -- & \circled{4} 16.2  & -- & --\\
Synth generated by model \circled{4} & 13.7  & --  & -- & --\\ 
% \hdashline
Averaging + beam 8         & 14.3 & 17.4  & 8.7 & 18.5\\
\bottomrule
\end{tabularx}
\end{center}
\label{tab:main_results}
\end{table}

The baseline models in \cref{tab:main_results} are trained on the authentic data only, and it seems that the number of parallel sentences is not sufficient to train the NMT model for the investigated language pairs (we obtained performance of 0.5 to 4.2 BLEU). 
The remaining rows show incremental improvements as we perform the various training stages. The last stage of model averaging takes the best performing model and averages it with the previous seven checkpoints that are one and a half hours of training time from each other.

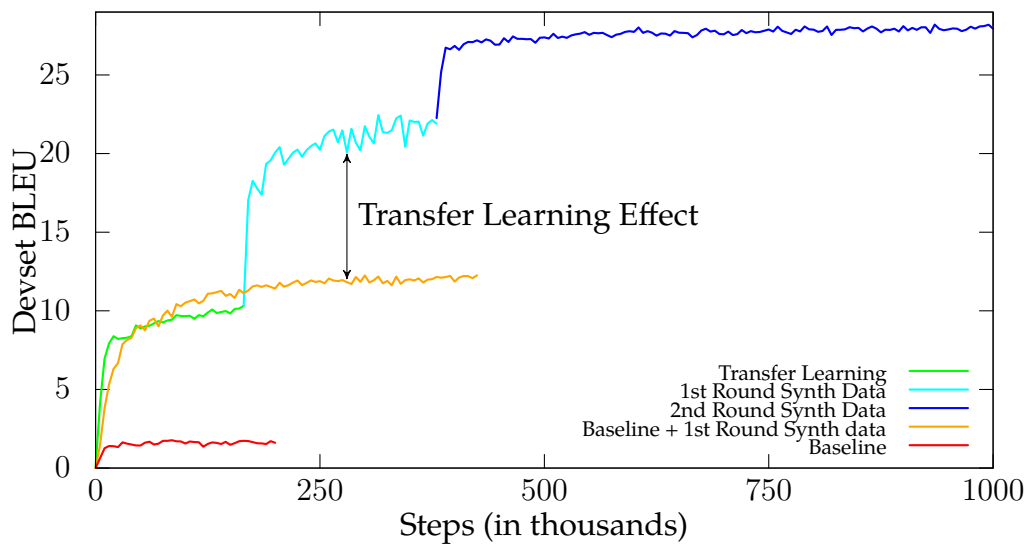
\begin{figure}
\caption{Learning curves of \transl{Gujarati}{English} models. Our approach is combination of four steps. The first is to train the parent model for 2000k steps (not in the figure), then transfer learning (green curve). Then we continue with two rounds of backtranslated data (both blue curves). Baseline without transfer learning and backtraslation is in red. Standard approach of backtranslation without transfer learning is in orange.}
\begin{center}
\begin{tikzpicture}[gnuplot]
%% generated with GNUPLOT 5.0p3 (Lua 5.1; terminal rev. 99, script rev. 100)
%% Sat 28 Sep 2019 10:58:43 AM DST
\path (0.000,0.000) rectangle (4.500,7.400);
\gpcolor{color=gp lt color border}
\gpsetlinetype{gp lt border}
\gpsetdashtype{gp dt solid}
\gpsetlinewidth{1.00}
\draw[gp path] (1.136,0.985)--(1.316,0.985);
\draw[gp path] (12.945,0.985)--(12.765,0.985);
\node[gp node right] at (0.952,0.985) {$0$};
\draw[gp path] (1.136,2.027)--(1.316,2.027);
\draw[gp path] (12.945,2.027)--(12.765,2.027);
\node[gp node right] at (0.952,2.027) {$5$};
\draw[gp path] (1.136,3.070)--(1.316,3.070);
\draw[gp path] (12.945,3.070)--(12.765,3.070);
\node[gp node right] at (0.952,3.070) {$10$};
\draw[gp path] (1.136,4.112)--(1.316,4.112);
\draw[gp path] (12.945,4.112)--(12.765,4.112);
\node[gp node right] at (0.952,4.112) {$15$};
\draw[gp path] (1.136,5.155)--(1.316,5.155);
\draw[gp path] (12.945,5.155)--(12.765,5.155);
\node[gp node right] at (0.952,5.155) {$20$};
\draw[gp path] (1.136,6.197)--(1.316,6.197);
\draw[gp path] (12.945,6.197)--(12.765,6.197);
\node[gp node right] at (0.952,6.197) {$25$};
\draw[gp path] (1.136,0.985)--(1.136,1.165);
\draw[gp path] (1.136,7.031)--(1.136,6.851);
\node[gp node center] at (1.136,0.677) {$0$};
\draw[gp path] (4.088,0.985)--(4.088,1.165);
\draw[gp path] (4.088,7.031)--(4.088,6.851);
\node[gp node center] at (4.088,0.677) {$250$};
\draw[gp path] (7.041,0.985)--(7.041,1.165);
\draw[gp path] (7.041,7.031)--(7.041,6.851);
\node[gp node center] at (7.041,0.677) {$500$};
\draw[gp path] (9.993,0.985)--(9.993,1.165);
\draw[gp path] (9.993,7.031)--(9.993,6.851);
\node[gp node center] at (9.993,0.677) {$750$};
\draw[gp path] (12.945,0.985)--(12.945,1.165);
\draw[gp path] (12.945,7.031)--(12.945,6.851);
\node[gp node center] at (12.945,0.677) {$1000$};
\draw[gp path] (1.136,7.031)--(1.136,0.985)--(12.945,0.985)--(12.945,7.031)--cycle;
\node[gp node left] at (4.443,4.321) {Transfer Learning Effect};
\draw[gp path,<->](4.443,3.487)--(4.443,5.155);
\node[gp node center,rotate=-270] at (0.246,4.008) {Devset BLEU};
\node[gp node center] at (7.040,0.215) {Steps (in thousands)};
\node[gp node right,font={\fontsize{8pt}{9.6pt}\selectfont}] at (11.699,2.272) {Transfer Learning};
\gpcolor{rgb color={0.000,1.000,0.000}}
\gpsetlinewidth{2.00}
\draw[gp path] (11.846,2.272)--(12.614,2.272);
\draw[gp path] (1.136,1.005)--(1.195,1.848)--(1.254,2.437)--(1.313,2.638)--(1.372,2.733)%
  --(1.431,2.697)--(1.490,2.707)--(1.549,2.713)--(1.608,2.732)--(1.667,2.876)--(1.726,2.837)%
  --(1.785,2.863)--(1.845,2.873)--(1.904,2.902)--(1.963,2.934)--(2.022,2.915)--(2.081,2.942)%
  --(2.140,2.952)--(2.199,3.013)--(2.258,3.001)--(2.317,3.000)--(2.376,3.004)--(2.435,2.963)%
  --(2.494,3.013)--(2.553,2.998)--(2.612,3.047)--(2.671,3.087)--(2.730,3.041)--(2.789,3.054)%
  --(2.848,3.067)--(2.907,3.035)--(2.966,3.095)--(3.025,3.100)--(3.084,3.136);
\gpcolor{color=gp lt color border}
\node[gp node right,font={\fontsize{8pt}{9.6pt}\selectfont}] at (11.699,2.026) {1st Round Synth Data};
\gpcolor{rgb color={0.000,1.000,1.000}}
\draw[gp path] (11.846,2.026)--(12.614,2.026);
\draw[gp path] (3.084,3.136)--(3.144,4.537)--(3.203,4.793)--(3.262,4.692)--(3.321,4.610)%
  --(3.380,5.017)--(3.439,5.064)--(3.498,5.169)--(3.557,5.240)--(3.616,5.008)--(3.675,5.088)%
  --(3.734,5.168)--(3.793,5.207)--(3.852,5.113)--(3.911,5.199)--(3.970,5.254)--(4.029,5.290)%
  --(4.088,5.205)--(4.147,5.387)--(4.206,5.447)--(4.265,5.473)--(4.324,5.303)--(4.383,5.462)%
  --(4.443,5.170)--(4.502,5.483)--(4.561,5.299)--(4.620,5.197)--(4.679,5.519)--(4.738,5.377)%
  --(4.797,5.291)--(4.856,5.664)--(4.915,5.439)--(4.974,5.429)--(5.033,5.463)--(5.092,5.622)%
  --(5.151,5.658)--(5.210,5.244)--(5.269,5.594)--(5.328,5.574)--(5.387,5.578)--(5.446,5.392)%
  --(5.505,5.553)--(5.564,5.599)--(5.623,5.551);
\gpcolor{color=gp lt color border}
\node[gp node right,font={\fontsize{8pt}{9.6pt}\selectfont}] at (11.699,1.780) {2nd Round Synth Data};
\gpcolor{rgb color={0.000,0.000,1.000}}
\draw[gp path] (11.846,1.780)--(12.614,1.780);
\draw[gp path] (5.623,5.625)--(5.682,6.234)--(5.742,6.559)--(5.801,6.537)--(5.860,6.582)%
  --(5.919,6.530)--(5.978,6.604)--(6.037,6.635)--(6.096,6.632)--(6.155,6.656)--(6.214,6.628)%
  --(6.273,6.669)--(6.332,6.651)--(6.391,6.598)--(6.450,6.600)--(6.509,6.684)--(6.568,6.634)%
  --(6.627,6.708)--(6.686,6.676)--(6.745,6.676)--(6.804,6.671)--(6.863,6.673)--(6.922,6.624)%
  --(6.981,6.690)--(7.041,6.695)--(7.100,6.680)--(7.159,6.743)--(7.218,6.687)--(7.277,6.705)%
  --(7.336,6.689)--(7.395,6.721)--(7.454,6.736)--(7.513,6.745)--(7.572,6.765)--(7.631,6.729)%
  --(7.690,6.755)--(7.749,6.750)--(7.808,6.755)--(7.867,6.725)--(7.926,6.698)--(7.985,6.731)%
  --(8.044,6.742)--(8.103,6.768)--(8.162,6.777)--(8.221,6.768)--(8.280,6.827)--(8.339,6.755)%
  --(8.399,6.778)--(8.458,6.757)--(8.517,6.718)--(8.576,6.747)--(8.635,6.744)--(8.694,6.719)%
  --(8.753,6.779)--(8.812,6.757)--(8.871,6.755)--(8.930,6.751)--(8.989,6.696)--(9.048,6.726)%
  --(9.107,6.787)--(9.166,6.810)--(9.225,6.772)--(9.284,6.784)--(9.343,6.740)--(9.402,6.736)%
  --(9.461,6.772)--(9.520,6.726)--(9.579,6.738)--(9.638,6.793)--(9.698,6.751)--(9.757,6.778)%
  --(9.816,6.763)--(9.875,6.767)--(9.934,6.783)--(9.993,6.800)--(10.052,6.772)--(10.111,6.837)%
  --(10.170,6.782)--(10.229,6.761)--(10.288,6.740)--(10.347,6.797)--(10.406,6.774)--(10.465,6.840)%
  --(10.524,6.802)--(10.583,6.798)--(10.642,6.769)--(10.701,6.780)--(10.760,6.840)--(10.819,6.727)%
  --(10.878,6.795)--(10.937,6.796)--(10.997,6.802)--(11.056,6.830)--(11.115,6.817)--(11.174,6.821)%
  --(11.233,6.774)--(11.292,6.813)--(11.351,6.840)--(11.410,6.811)--(11.469,6.762)--(11.528,6.797)%
  --(11.587,6.794)--(11.646,6.810)--(11.705,6.773)--(11.764,6.844)--(11.823,6.788)--(11.882,6.821)%
  --(11.941,6.795)--(12.000,6.829)--(12.059,6.829)--(12.118,6.754)--(12.177,6.865)--(12.236,6.808)%
  --(12.296,6.790)--(12.355,6.800)--(12.414,6.814)--(12.473,6.810)--(12.532,6.778)--(12.591,6.797)%
  --(12.650,6.837)--(12.709,6.811)--(12.768,6.840)--(12.827,6.845)--(12.886,6.863)--(12.945,6.810);
\gpcolor{color=gp lt color border}
\node[gp node right,font={\fontsize{8pt}{9.6pt}\selectfont}] at (11.699,1.534) {Baseline + 1st Round Synth data};
\gpcolor{rgb color={1.000,0.647,0.000}}
\draw[gp path] (11.846,1.534)--(12.614,1.534);
\draw[gp path] (1.136,0.985)--(1.195,1.294)--(1.254,1.781)--(1.313,2.093)--(1.372,2.300)%
  --(1.431,2.377)--(1.490,2.628)--(1.549,2.682)--(1.608,2.711)--(1.667,2.839)--(1.726,2.877)%
  --(1.785,2.811)--(1.845,2.938)--(1.904,2.967)--(1.963,2.864)--(2.022,3.008)--(2.081,3.072)%
  --(2.140,2.991)--(2.199,3.159)--(2.258,3.130)--(2.317,3.178)--(2.376,3.198)--(2.435,3.221)%
  --(2.494,3.169)--(2.553,3.202)--(2.612,3.297)--(2.671,3.301)--(2.730,3.318)--(2.789,3.334)%
  --(2.848,3.268)--(2.907,3.293)--(2.966,3.241)--(3.025,3.349)--(3.084,3.303)--(3.144,3.338)%
  --(3.203,3.393)--(3.262,3.407)--(3.321,3.387)--(3.380,3.408)--(3.439,3.387)--(3.498,3.363)%
  --(3.557,3.442)--(3.616,3.389)--(3.675,3.411)--(3.734,3.445)--(3.793,3.472)--(3.852,3.410)%
  --(3.911,3.440)--(3.970,3.473)--(4.029,3.453)--(4.088,3.463)--(4.147,3.435)--(4.206,3.498)%
  --(4.265,3.472)--(4.324,3.464)--(4.383,3.477)--(4.443,3.450)--(4.502,3.425)--(4.561,3.519)%
  --(4.620,3.454)--(4.679,3.539)--(4.738,3.445)--(4.797,3.475)--(4.856,3.525)--(4.915,3.434)%
  --(4.974,3.467)--(5.033,3.408)--(5.092,3.524)--(5.151,3.427)--(5.210,3.477)--(5.269,3.487)%
  --(5.328,3.472)--(5.387,3.500)--(5.446,3.446)--(5.505,3.483)--(5.564,3.490)--(5.623,3.519)%
  --(5.682,3.509)--(5.742,3.518)--(5.801,3.533)--(5.860,3.458)--(5.919,3.511)--(5.978,3.531)%
  --(6.037,3.530)--(6.096,3.503)--(6.155,3.540);
\gpcolor{color=gp lt color border}
\node[gp node right,font={\fontsize{8pt}{9.6pt}\selectfont}] at (11.699,1.288) {Baseline};
\gpcolor{rgb color={1.000,0.000,0.000}}
\draw[gp path] (11.846,1.288)--(12.614,1.288);
\draw[gp path] (1.136,0.985)--(1.195,1.116)--(1.254,1.248)--(1.313,1.277)--(1.372,1.275)%
  --(1.431,1.263)--(1.490,1.326)--(1.549,1.309)--(1.608,1.297)--(1.667,1.285)--(1.726,1.284)%
  --(1.785,1.321)--(1.845,1.334)--(1.904,1.297)--(1.963,1.308)--(2.022,1.346)--(2.081,1.346)%
  --(2.140,1.354)--(2.199,1.340)--(2.258,1.338)--(2.317,1.310)--(2.376,1.337)--(2.435,1.330)%
  --(2.494,1.327)--(2.553,1.268)--(2.612,1.305)--(2.671,1.325)--(2.730,1.318)--(2.789,1.304)%
  --(2.848,1.333)--(2.907,1.293)--(2.966,1.315)--(3.025,1.341)--(3.084,1.346)--(3.144,1.343)%
  --(3.203,1.329)--(3.262,1.314)--(3.321,1.312)--(3.380,1.303)--(3.439,1.342)--(3.498,1.320);
\gpcolor{color=gp lt color border}
\gpsetlinewidth{1.00}
\draw[gp path] (1.136,7.031)--(1.136,0.985)--(12.945,0.985)--(12.945,7.031)--cycle;
%% coordinates of the plot area
\gpdefrectangularnode{gp plot 1}{\pgfpoint{1.136cm}{0.985cm}}{\pgfpoint{12.945cm}{7.031cm}}
\end{tikzpicture}
%% gnuplot variables
\end{center}
\label{fig:backtranslation_progress}
\end{figure}

\cref{fig:backtranslation_progress} above shows the progress of training of \pair{Gujarati}{English} models in both directions. 
We can notice that after each change of parallel corpus, there is a substantial improvement in the performance. 
The learning curve is computed on the development data. The corresponding scores for the testsets are in \cref{tab:main_results}.

We also run a standard approach of backtranslation without transfer learning, where we first trained baseline model to translate monolingual data and then trained a model (in reverse direction) on those synthetic data with concatenation with authentic data. We visualize the training with orange in \cref{fig:backtranslation_progress}. When comparing with a transferred model trained with the first round of backtranslation data (light blue), we can see the clear improvement in performance due to transfer learning.\footnote{Each scenario (orange/light blue) use synthetic data generated by different model from an equal monolingual corpus. The model generating the synthetic data either used transfer learning (light blue) or did not (orange).}

\observation{Transfer learning can be used as an initial step for the low-resource \ac{NMT} training with backtranslation technique to largely improve the performance.}{obs:transfer_helps_backtranslation}

In conclusion, we showed that transfer learning can be used in combination with other techniques. Furthermore, it can be used to generate a reasonably good initial model for backtranslation technique in the low-resource scenario, where it is hard to train the model from random initialization.

\subsection{Ratio between Authentic and Synthetic}

Backtranslation as a way of transferring knowledge through data to the other direction can be used in theory on as much monolingual data as accessible. For example, English has billions of monolingual data. Thus we could create a training corpus for translation from any language to the English with billions of sentences for all languages. However, this corpus would not contain the authentic data in 1:1 ratio as proposed by \perscite{sennrich2016backtranslation}.

\perscite{edunov2018backtranslation} showed that the shortage of authentic data in the synthetic corpus could be supplemented by oversampling the authentic data.

\perscite{poncelas2018investigating} investigated the effect of various ratios between authentic and synthetic training data on \transl{German}{English} scenario. The authors suggest the ratio of 1:2 in favor of synthetic data. The higher ratio of synthetic data does not help but also does not decrease performance significantly. The authors have not experimented with oversampling of authentic data.

To conclude their findings, we can use a synthetic corpus of any size as long as we mix the authentic data into the corpus. In case that we lack a large quantity of authentic data, we can oversample the available ones. Nonetheless, they experimented with the high-resource language only. Hence we evaluated their findings on a low-resource scenario.

We experimented with \pair{Gujarati}{English} language pair, which has only 172k parallel sentences. We used a warm-start transfer learning and two rounds of backtranslation as proposed by \perscite{hoang2018iterative}. During the second round of training on the backtranslated data, we evaluated various ratios between the synthetic and authentic data.

In order to report the ratio between authentic and synthetic, we backtranslated 3.6M English sentences into Gujarati, which is exactly twenty times more monolingual data than the size of authentic data. Then we oversampled the authentic data and mixed them to the synthetic corpus to get final training set. For example, the ratio ``authentic:synthetic 10:1'' means that the authentic has been multiplied ten times, which to match half size of the synthetic sentences.

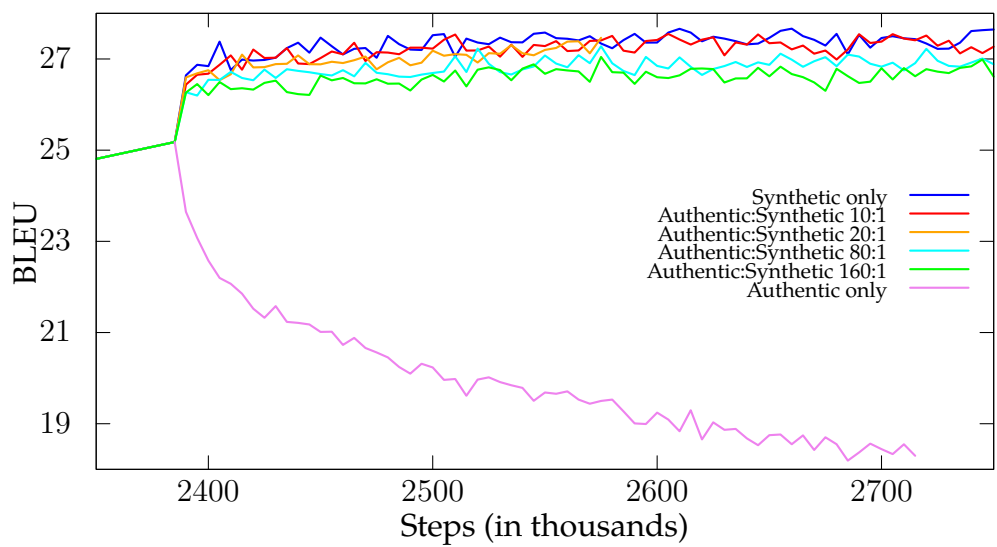
\begin{figure}
\caption{Comparison of various ratio between the synthetic and authentic sentences.}
\begin{center}
\begin{tikzpicture}[gnuplot]
%% generated with GNUPLOT 5.0p3 (Lua 5.1; terminal rev. 99, script rev. 100)
%% Fri 13 Sep 2019 01:27:35 PM DST
\path (0.000,0.000) rectangle (4.500,7.400);
\gpcolor{color=gp lt color border}
\gpsetlinetype{gp lt border}
\gpsetdashtype{gp dt solid}
\gpsetlinewidth{1.00}
\draw[gp path] (1.136,1.590)--(1.316,1.590);
\draw[gp path] (12.945,1.590)--(12.765,1.590);
\node[gp node right] at (0.952,1.590) {$19$};
\draw[gp path] (1.136,2.799)--(1.316,2.799);
\draw[gp path] (12.945,2.799)--(12.765,2.799);
\node[gp node right] at (0.952,2.799) {$21$};
\draw[gp path] (1.136,4.008)--(1.316,4.008);
\draw[gp path] (12.945,4.008)--(12.765,4.008);
\node[gp node right] at (0.952,4.008) {$23$};
\draw[gp path] (1.136,5.217)--(1.316,5.217);
\draw[gp path] (12.945,5.217)--(12.765,5.217);
\node[gp node right] at (0.952,5.217) {$25$};
\draw[gp path] (1.136,6.426)--(1.316,6.426);
\draw[gp path] (12.945,6.426)--(12.765,6.426);
\node[gp node right] at (0.952,6.426) {$27$};
\draw[gp path] (2.612,0.985)--(2.612,1.165);
\draw[gp path] (2.612,7.031)--(2.612,6.851);
\node[gp node center] at (2.612,0.677) {$2400$};
\draw[gp path] (5.564,0.985)--(5.564,1.165);
\draw[gp path] (5.564,7.031)--(5.564,6.851);
\node[gp node center] at (5.564,0.677) {$2500$};
\draw[gp path] (8.517,0.985)--(8.517,1.165);
\draw[gp path] (8.517,7.031)--(8.517,6.851);
\node[gp node center] at (8.517,0.677) {$2600$};
\draw[gp path] (11.469,0.985)--(11.469,1.165);
\draw[gp path] (11.469,7.031)--(11.469,6.851);
\node[gp node center] at (11.469,0.677) {$2700$};
\draw[gp path] (1.136,7.031)--(1.136,0.985)--(12.945,0.985)--(12.945,7.031)--cycle;
\node[gp node center,rotate=-270] at (0.246,4.008) {BLEU};
\node[gp node center] at (7.040,0.215) {Steps (in thousands)};
\node[gp node right,font={\fontsize{8pt}{9.6pt}\selectfont}] at (11.699,4.623) {Synthetic only};
\gpcolor{rgb color={0.000,0.000,1.000}}
\gpsetlinewidth{2.00}
\draw[gp path] (11.846,4.623)--(12.614,4.623);
\draw[gp path] (1.136,5.101)--(2.169,5.325)--(2.317,6.209)--(2.465,6.349)--(2.612,6.331)%
  --(2.760,6.656)--(2.907,6.280)--(3.055,6.418)--(3.203,6.405)--(3.350,6.413)--(3.498,6.443)%
  --(3.645,6.570)--(3.793,6.642)--(3.941,6.510)--(4.088,6.709)--(4.236,6.595)--(4.383,6.486)%
  --(4.531,6.561)--(4.679,6.571)--(4.826,6.442)--(4.974,6.732)--(5.122,6.622)--(5.269,6.551)%
  --(5.417,6.545)--(5.564,6.741)--(5.712,6.756)--(5.860,6.454)--(6.007,6.696)--(6.155,6.644)%
  --(6.302,6.622)--(6.450,6.709)--(6.598,6.648)--(6.745,6.646)--(6.893,6.760)--(7.041,6.776)%
  --(7.188,6.702)--(7.336,6.694)--(7.483,6.672)--(7.631,6.728)--(7.779,6.631)--(7.926,6.568)%
  --(8.074,6.676)--(8.221,6.759)--(8.369,6.643)--(8.517,6.644)--(8.664,6.776)--(8.812,6.826)%
  --(8.959,6.776)--(9.107,6.660)--(9.255,6.721)--(9.402,6.696)--(9.550,6.662)--(9.698,6.620)%
  --(9.845,6.630)--(9.993,6.706)--(10.140,6.799)--(10.288,6.828)--(10.436,6.729)--(10.583,6.680)%
  --(10.731,6.604)--(10.878,6.759)--(11.026,6.488)--(11.174,6.746)--(11.321,6.576)--(11.469,6.704)%
  --(11.616,6.728)--(11.764,6.700)--(11.912,6.689)--(12.059,6.631)--(12.207,6.560)--(12.355,6.563)%
  --(12.502,6.642)--(12.650,6.796)--(12.797,6.808)--(12.945,6.817);
\gpcolor{color=gp lt color border}
\node[gp node right,font={\fontsize{8pt}{9.6pt}\selectfont}] at (11.699,4.377) {Authentic:Synthetic 10:1};
\gpcolor{rgb color={1.000,0.000,0.000}}
\draw[gp path] (11.846,4.377)--(12.614,4.377);
\draw[gp path] (1.136,5.101)--(2.169,5.325)--(2.317,6.086)--(2.465,6.217)--(2.612,6.231)%
  --(2.760,6.349)--(2.907,6.473)--(3.055,6.284)--(3.203,6.551)--(3.350,6.439)--(3.498,6.443)%
  --(3.645,6.570)--(3.793,6.366)--(3.941,6.358)--(4.088,6.438)--(4.236,6.525)--(4.383,6.488)%
  --(4.531,6.641)--(4.679,6.399)--(4.826,6.516)--(4.974,6.511)--(5.122,6.489)--(5.269,6.576)%
  --(5.417,6.576)--(5.564,6.563)--(5.712,6.681)--(5.860,6.750)--(6.007,6.537)--(6.155,6.539)%
  --(6.302,6.591)--(6.450,6.457)--(6.598,6.615)--(6.745,6.454)--(6.893,6.612)--(7.041,6.597)%
  --(7.188,6.663)--(7.336,6.531)--(7.483,6.539)--(7.631,6.665)--(7.779,6.672)--(7.926,6.732)%
  --(8.074,6.536)--(8.221,6.511)--(8.369,6.664)--(8.517,6.678)--(8.664,6.758)--(8.812,6.684)%
  --(8.959,6.617)--(9.107,6.682)--(9.255,6.714)--(9.402,6.474)--(9.550,6.646)--(9.698,6.616)%
  --(9.845,6.751)--(9.993,6.635)--(10.140,6.645)--(10.288,6.553)--(10.436,6.604)--(10.583,6.496)%
  --(10.731,6.538)--(10.878,6.419)--(11.026,6.552)--(11.174,6.754)--(11.321,6.638)--(11.469,6.656)%
  --(11.616,6.755)--(11.764,6.697)--(11.912,6.674)--(12.059,6.737)--(12.207,6.601)--(12.355,6.668)%
  --(12.502,6.500)--(12.650,6.582)--(12.797,6.504)--(12.945,6.589);
\gpcolor{color=gp lt color border}
\node[gp node right,font={\fontsize{8pt}{9.6pt}\selectfont}] at (11.699,4.131) {Authentic:Synthetic 20:1};
\gpcolor{rgb color={1.000,0.647,0.000}}
\draw[gp path] (11.846,4.131)--(12.614,4.131);
\draw[gp path] (1.136,5.101)--(2.169,5.325)--(2.317,6.181)--(2.465,6.236)--(2.612,6.277)%
  --(2.760,6.145)--(2.907,6.223)--(3.055,6.484)--(3.203,6.314)--(3.350,6.318)--(3.498,6.358)%
  --(3.645,6.362)--(3.793,6.473)--(3.941,6.357)--(4.088,6.355)--(4.236,6.389)--(4.383,6.372)%
  --(4.531,6.410)--(4.679,6.458)--(4.826,6.289)--(4.974,6.382)--(5.122,6.441)--(5.269,6.342)%
  --(5.417,6.377)--(5.564,6.536)--(5.712,6.469)--(5.860,6.490)--(6.007,6.481)--(6.155,6.379)%
  --(6.302,6.503)--(6.450,6.495)--(6.598,6.610)--(6.745,6.502)--(6.893,6.475)--(7.041,6.546)%
  --(7.188,6.572)--(7.336,6.656)--(7.483,6.661)--(7.631,6.501)--(7.779,6.707);
\gpcolor{color=gp lt color border}
\node[gp node right,font={\fontsize{8pt}{9.6pt}\selectfont}] at (11.699,3.885) {Authentic:Synthetic 80:1};
\gpcolor{rgb color={0.000,1.000,1.000}}
\draw[gp path] (11.846,3.885)--(12.614,3.885);
\draw[gp path] (1.136,5.101)--(2.169,5.325)--(2.317,5.985)--(2.465,5.941)--(2.612,6.150)%
  --(2.760,6.148)--(2.907,6.255)--(3.055,6.174)--(3.203,6.144)--(3.350,6.285)--(3.498,6.174)%
  --(3.645,6.290)--(3.793,6.268)--(3.941,6.250)--(4.088,6.226)--(4.236,6.207)--(4.383,6.278)%
  --(4.531,6.196)--(4.679,6.374)--(4.826,6.243)--(4.974,6.224)--(5.122,6.192)--(5.269,6.187)%
  --(5.417,6.219)--(5.564,6.240)--(5.712,6.259)--(5.860,6.462)--(6.007,6.253)--(6.155,6.564)%
  --(6.302,6.362)--(6.450,6.252)--(6.598,6.220)--(6.745,6.287)--(6.893,6.327)--(7.041,6.472)%
  --(7.188,6.365)--(7.336,6.316)--(7.483,6.476)--(7.631,6.370)--(7.779,6.594)--(7.926,6.369)%
  --(8.074,6.267)--(8.221,6.212)--(8.369,6.453)--(8.517,6.334)--(8.664,6.300)--(8.812,6.447)%
  --(8.959,6.322)--(9.107,6.214)--(9.255,6.293)--(9.402,6.332)--(9.550,6.386)--(9.698,6.320)%
  --(9.845,6.381)--(9.993,6.348)--(10.140,6.496)--(10.288,6.417)--(10.436,6.322)--(10.583,6.396)%
  --(10.731,6.450)--(10.878,6.330)--(11.026,6.484)--(11.174,6.458)--(11.321,6.363)--(11.469,6.324)%
  --(11.616,6.377)--(11.764,6.274)--(11.912,6.371)--(12.059,6.560)--(12.207,6.402)--(12.355,6.334)%
  --(12.502,6.324)--(12.650,6.378)--(12.797,6.432)--(12.945,6.356);
\gpcolor{color=gp lt color border}
\node[gp node right,font={\fontsize{8pt}{9.6pt}\selectfont}] at (11.699,3.639) {Authentic:Synthetic 160:1};
\gpcolor{rgb color={0.000,1.000,0.000}}
\draw[gp path] (11.846,3.639)--(12.614,3.639);
\draw[gp path] (1.136,5.101)--(2.169,5.325)--(2.317,5.981)--(2.465,6.093)--(2.612,5.947)%
  --(2.760,6.120)--(2.907,6.026)--(3.055,6.038)--(3.203,6.020)--(3.350,6.109)--(3.498,6.139)%
  --(3.645,5.987)--(3.793,5.960)--(3.941,5.949)--(4.088,6.206)--(4.236,6.143)--(4.383,6.174)%
  --(4.531,6.104)--(4.679,6.102)--(4.826,6.157)--(4.974,6.098)--(5.122,6.100)--(5.269,6.008)%
  --(5.417,6.153)--(5.564,6.215)--(5.712,6.127)--(5.860,6.273)--(6.007,6.062)--(6.155,6.284)%
  --(6.302,6.315)--(6.450,6.279)--(6.598,6.144)--(6.745,6.298)--(6.893,6.355)--(7.041,6.229)%
  --(7.188,6.291)--(7.336,6.274)--(7.483,6.261)--(7.631,6.123)--(7.779,6.453)--(7.926,6.251)%
  --(8.074,6.245)--(8.221,6.096)--(8.369,6.257)--(8.517,6.184)--(8.664,6.173)--(8.812,6.209)%
  --(8.959,6.296)--(9.107,6.300)--(9.255,6.292)--(9.402,6.114)--(9.550,6.168)--(9.698,6.169)%
  --(9.845,6.309)--(9.993,6.196)--(10.140,6.324)--(10.288,6.226)--(10.436,6.185)--(10.583,6.115)%
  --(10.731,6.004)--(10.878,6.296)--(11.026,6.198)--(11.174,6.108)--(11.321,6.125)--(11.469,6.303)%
  --(11.616,6.155)--(11.764,6.305)--(11.912,6.197)--(12.059,6.291)--(12.207,6.258)--(12.355,6.239)%
  --(12.502,6.308)--(12.650,6.328)--(12.797,6.419)--(12.945,6.191);
\gpcolor{color=gp lt color border}
\node[gp node right,font={\fontsize{8pt}{9.6pt}\selectfont}] at (11.699,3.393) {Authentic only};
\gpcolor{rgb color={0.933,0.510,0.933}}
\draw[gp path] (11.846,3.393)--(12.614,3.393);
\draw[gp path] (2.169,5.325)--(2.317,4.399)--(2.465,4.053)--(2.612,3.753)--(2.760,3.523)%
  --(2.907,3.445)--(3.055,3.314)--(3.203,3.115)--(3.350,2.995)--(3.498,3.149)--(3.645,2.940)%
  --(3.793,2.927)--(3.941,2.906)--(4.088,2.807)--(4.236,2.810)--(4.383,2.635)--(4.531,2.728)%
  --(4.679,2.594)--(4.826,2.535)--(4.974,2.469)--(5.122,2.342)--(5.269,2.254)--(5.417,2.385)%
  --(5.564,2.336)--(5.712,2.171)--(5.860,2.183)--(6.007,1.961)--(6.155,2.175)--(6.302,2.205)%
  --(6.450,2.142)--(6.598,2.100)--(6.745,2.065)--(6.893,1.894)--(7.041,2.004)--(7.188,1.987)%
  --(7.336,2.019)--(7.483,1.910)--(7.631,1.855)--(7.779,1.892)--(7.926,1.910)--(8.074,1.754)%
  --(8.221,1.594)--(8.369,1.586)--(8.517,1.737)--(8.664,1.645)--(8.812,1.490)--(8.959,1.768)%
  --(9.107,1.382)--(9.255,1.608)--(9.402,1.509)--(9.550,1.521)--(9.698,1.395)--(9.845,1.305)%
  --(9.993,1.438)--(10.140,1.447)--(10.288,1.318)--(10.436,1.435)--(10.583,1.242)--(10.731,1.410)%
  --(10.878,1.318)--(11.026,1.101)--(11.174,1.209)--(11.321,1.326)--(11.469,1.253)--(11.616,1.187)%
  --(11.764,1.317)--(11.912,1.165);
\gpcolor{color=gp lt color border}
\gpsetlinewidth{1.00}
\draw[gp path] (1.136,7.031)--(1.136,0.985)--(12.945,0.985)--(12.945,7.031)--cycle;
%% coordinates of the plot area
\gpdefrectangularnode{gp plot 1}{\pgfpoint{1.136cm}{0.985cm}}{\pgfpoint{12.945cm}{7.031cm}}
\end{tikzpicture}
%% gnuplot variables
\end{center}
\label{fig:backtranslation_ratio_auth}
\end{figure}

In \cref{fig:backtranslation_ratio_auth}, we can see the difference in performance when comparing the ratio between the amount of synthetic and authentic data. The model learning curves start at the 2.4M step as a visualization of their previous transfer learning and the first round of backtranslation (see \cref{fig:backtranslation_progress}).

Using only low-resource authentic parallel sentences drastically damages the performance of an already good model. Thus the previous training on synthetic data is necessary. However, the most surprising fact is that adding more authentic data damages the performance, which is in contrast to claims of \perscite{poncelas2018investigating} and \perscite{edunov2018backtranslation}. 

\observation{Oversampling authentic data to match synthetic data hurts \ac{NMT} performance.}{obs:oversampling_hurts}

To explain this phenomenon, we need to take into account that in order to match the ratio, we oversampled the authentic data multiple times. For example, the worst-performing ratio of ``160:1'' had 160 copies of authentic data within, which could lead to overfitting on them. Thus we believe that the authentic data are useful in synthetic corpus only to some extent and after having much more synthetic data, it is better to use only the synthetic high number of parallel sentences instead of adding more and more copies of authentic data in order to match a ratio between authentic and synthetic data. We would like to understand where the tipping point is and if we can add more monolingual data and obtain a better performance. However, it is over the scope of this thesis, and we leave it as an open question.

This experiment once again supported our intuition that the more data \ac{NMT} has, the better performance it reaches even though they are artificially created, even to the extent that the authentic data become useless.

In conclusion, we showed that transfer learning can be combined with other techniques. It substantially helps when used with backtranslation. Furthermore, we showed that oversampling authentic data \parcite{edunov2018backtranslation} is not useful for low-resource languages and can lead to obtaining lower performance than without oversampling.

\chapter{Conclusion}
\label{chap:conclusion}

As stated in the introduction, this thesis had two main parts. The first was the introduction of various techniques for transfer learning in \ac{NMT}. The second was a broad analysis of transfer learning behavior. 

We presented methods for both the cold-start and warm-start scenario, where both of them outperform a baseline that is trained without transfer learning. We showed that transfer learning helps for low-resource as well as high-resource languages.

In the cold-start scenario, we presented Direct Transfer, a technique that trains a child model without any modifications, and Transformed Vocabulary, a technique that adapts a parent vocabulary for the need of the child by randomly overriding unused embeddings. Furthermore, we showed a proof-of-concept of training a model from the parent that has not been trained by us. This concept opens doors to better replicability of experiments. With this technique, researchers could ``never train their models from scratch ever again''.\footnote{This statement is obviously a bit exaggerated.}

In the warm-start scenario, we proposed two approaches: Merged and Balanced Vocabulary. Both of them prepared vocabulary in advance of the parent training. We showed significant gains even over the cold-start technique.

The second part of this thesis is the analysis of transfer learning behavior. We shed some light on transfer learning in \ac{NMT}. We studied the phenomenon of negative transfer. We showed that transfer learning behaves differently concerning the position of the shared language. Moreover, we studied various aspects of transfer learning, and we concluded that the main gains are due to the size of the parent model. Furthermore, we believe that transfer learning behaves as a better initialization method to some extent.

All our observations can be found in the \emph{List of Observations} on page \pageref{list:observations}. However, our study is only a small step in understanding transfer learning or even neural networks.

\section{Ecological Trace}
\newcommand{\COtwo}{$\text{CO}_2$}

Earlier this year, \perscite{strubell2019energy} published work on the impact of deep learning on \COtwo{} emissions. They estimated that a single Transformer architecture hyper-parameter search produced 284 tonnes of \COtwo{}.\footnote{The paper reports numbers in imperial units, for which it was criticized. We recomputed the numbers for SI units.} Many researchers pointed out that it reports numbers based on the U.S. energy mix. However, Google claims that its platform is 100\% renewable.\footnote{Source: \url{https://cloud.google.com/sustainability/}} Moreover, the most carbon-intensive scenario in the study costs between \$1 million and \$3 million \parcite{strubell2019energy}, which is not an everyday expense.

We believe that it is important to raise awareness of and quantify the potential \COtwo{} impact of deep learning. Thus we try to calculate the impact of this study. We roughly calculate the wall-clock time, power consumption, and \COtwo{} emissions.

We start by calculating the total wall-clock time on GPUs. Our cluster saves information about GPU usage every 10 minutes. We recovered logs from the last 20 months (the time when almost all of our transfer learning experiments have been done) and calculated how many GPU hours we spent on our experiments.

The average power consumption is harder to calculate. Based on discussion with our IT department, the average GeForce takes 200W per hour and Quadro P5000 160W per hour, when entirely in use. We need to take into account also the air-conditioning of the room, which can be roughly calculated by coefficient 1.3 to 1.4, we take the 1.35 as a middle. With this in mind, we use 270W for Tesla and GeForce cards and 216W for Quadro P5000.

The last step is the calculation of \COtwo{} emissions. Based on the report by \perscite{oecd2015report}, the average \COtwo{} emissions during the production of electricity in the Czech Republic is 516g per kWh (in the year 2015). 

\begin{table}[t]
\caption{The total amount of time spent, energy consumed, and \COtwo{} emitted during our transfer learning research. The numbers are only a rough estimate.}
\begin{center}
\begin{tabularx}{\textwidth}{@{}lrrr@{}}
\toprule
GPU type & Time spent & Energy Consumption & \COtwo{} Emissions\\
\midrule
Tesla K40c          &  3658 hours &  1.0 GWh & 0.5 t\\
GeForce GTX 1080    &  2221 hours &  0.6 GWh & 0.3 t\\
GeForce GTX 1080 Ti & 52227 hours & 14.1 GWh & 7.2 t\\
Quadro P5000        & 14996 hours &  3.2 GWh & 1.7 t\\
\midrule
Total               & 73102 hours & 18.9 GWh & 9.8 t\\
\bottomrule
\end{tabularx}
\end{center}
\label{tab:ecological_trace}
\end{table}

\cref{tab:ecological_trace} represent the total number of hours, energy consumption and \COtwo{} emissions for our experiments. An important notice is that energy consumption is  \emph{very roughly} estimated. It is based on GPU in full use, and the effect of air-conditioning is only estimated. Therefore the numbers are more likely the upper bound.

\begin{table}[t]
\caption{Comparison of \COtwo{} emissions from various sources. The numbers are from \parcite{strubell2019energy}.}
\begin{center}
\begin{tabularx}{\textwidth}{@{}lY@{}}
\toprule
GPU type & \COtwo{} Emissions\\
\midrule
Our research                        &  9.8 t\\
Air travel NY $\leftrightarrow$ SF per passenger &  0.9 t \\
Average person production in year                 &  5.0 t \\
Average American person production in year         & 16.4 t \\
Lifetime car consumption             & 57.2 t \\
\bottomrule
\end{tabularx}
\end{center}
\label{tab:ecological_trace_comparison}
\end{table}

Our experiments produced a similar amount of \COtwo{} as eleven round-trip-flights from New York to San Francisco for one passenger or as an average person produces within two years. For more comparisons, see \cref{tab:ecological_trace_comparison}.

Nonetheless, we proposed a way that reduces the total training time. Primarily the cold-start scenario can be used as a technique, which can be used to improve performance and lower the total training time.

Furthermore, machine learning can help to tackle climate change in various ways: predicting the electricity demand; flexibly managing household, commercial or electric vehicle demand; optimizing transportation routes; forecasting extreme climate events; modeling disease spreading; and many more as summarized by \perscite{rolnick2019tackling}.

In conclusion, deep learning is energy-intensive, and further research is needed to find the best ways to minimize the impact. However, restrictions on research are not the answer to the problem. The main issue is the generation of electricity, which for most of \COtwo{} emission in the Czech Republic is in coal power plants. As mentioned earlier, Google claims that its AI cluster is 100\% run by renewable energy; Amazon claims 50\%.\footnote{Source: \url{https://aws.amazon.com/about-aws/sustainability/}}

\section{Final Words}

I am fortunate to have gone through an exciting journey of studying \ac{NMT} since its early days of becoming the standard approach in \ac{MT}. And I hope this dissertation thesis will provide useful background and inspiration for future research in \ac{NMT} transfer learning.

Lastly, it would mean a lot to me if you would give me feedback or a short comment at \url{http://kocmi.tk/thesis/}.

%%% Seznam použité literatury
\bibliography{thesis}
\bibliographystyle{plainnat}

\bibliographyA{A}
\bibliographystyleA{plainnat}
\label{list:publications}

%%% Tabulky v disertační práci, existují-li.
% \chapwithtoc{List of Tables}

%%% Použité zkratky v disertační práci, existují-li, včetně jejich vysvětlení.
\chapwithtoc{List of Abbreviations}

\begin{acronym}
\acro{BiRNN}{Bidirectional Recurrent Neural Network}
\acro{BP}{Brevity Penalty}
\acro{BPE}{Byte Pair Encoding}
\acro{CNN}{Convolutional Neural Network}
\acro{IWSLT}{International Workshop on Spoken Language Translation}
\acro{LanideNN}{Language Identification by Neural Networks}
\acro{LM}{Language Model}
\acro{LSTM}{Long Short-Term Memory cells}
\acro{MT}{Machine Translation}
\acro{NLP}{Natural Language Processing}
\acro{NMT}{Neural Machine Translation}
\acro{NN}{Neural Network}
\acro{PBMT}{Phrase-Based Machine Translation}
\acro{POS}{Part-of-Speech}
\acro{OOV}{Out-of-Vocabulary}
\acro{RNN}{Recurrent Neural Network}
\acro{SMT}{Statistical Machine Translation}
\acro{T2T}{Tensor2tensor}
\acro{WAT}{Workshop on Asian Translation}
\acro{WMT}{Workshop on Statistical Machine Translation}
\end{acronym}
\cleardoublepage
\listofapiadd
\addcontentsline{toc}{chapter}{List of Observations}
\label{list:observations}
\cleardoublepage
\listoffigures
\cleardoublepage
\listoftables

%%% Přílohy k disertační práci, existují-li (různé dodatky jako výpisy programů,
%%% diagramy apod.). Každá příloha musí být alespoň jednou odkazována z vlastního
%%% textu práce. Přílohy se číslují.
% \chapwithtoc{Attachments}

\openright
\end{document}